\PassOptionsToPackage{unicode=true, bookmarksopen={true}, pdffitwindow=true, colorlinks=true, linkcolor=bluecite, citecolor=bluecite, urlcolor=bluecite, hyperfootnotes=true, pdfstartview={FitH}, pdfpagemode= UseNone}{hyperref}
\documentclass[ai,review,accept,pdftex,moreauthors]{Definitions/mdpi} 

\firstpage{1} 
\pubvolume{1}
\issuenum{1}
\articlenumber{0}
\pubyear{2023}
\copyrightyear{2023}
\externaleditor{Academic Editor: Firstname Lastname}
\datereceived{22 June 2023} 
\daterevised{6 August 2023} % Comment out if no revised date
\dateaccepted{15 August 2023} 
\datepublished{ } 
\hreflink{https://doi.org/} % If needed use \linebreak
\usepackage{amsmath,amsfonts}
\usepackage{algorithmic}
\usepackage{algorithm}
\usepackage{array}
\usepackage[caption=false,font=normalsize,labelfont=sf,textfont=sf]{subfig}
\usepackage{textcomp}
\usepackage{stfloats}
\usepackage{url}
\usepackage{verbatim}
\usepackage{graphicx}
\usepackage{rotating}
\usepackage{makecell}
\usepackage{multirow}
\usepackage{graphicx}
\usepackage{xcolor} %write text in a given color
\usepackage{soul}	%to highlight text, in addition to loading color or xcolor, we also need to load the soul package
\usepackage{tabularx}
\usepackage{xurl}
\usepackage{array}
\usepackage{changepage}
\usepackage{makecell}
\usepackage{graphicx}
\usepackage{epstopdf}

\renewcommand{\textsf}[1]{\text{#1}}
\renewcommand{\highlighting}[1]{#1}
\renewcommand{\hl}[1]{#1}
\Title{\highlighting{A} %MDPI: The article type is different from the one submitted online at susy.mdpi.com. Please confirm which one is correct -> the article type has been updated to "review"
 Comprehensive Review and a Taxonomy of Edge Machine Learning: Requirements, Paradigms, and Techniques}

\TitleCitation{A Comprehensive Review and a Taxonomy of Edge Machine Learning: Requirements, Paradigms, and Techniques}

\Author{\highlighting{Wenbin Li} %MDPI: Please carefully check the accuracy of names and affiliations.  -> correct
 *\orcidA{}, Hakim Hacid, Ebtesam Almazrouei and Merouane Debbah}

\AuthorNames{Wenbin Li, Hakim Hacid, Ebtesam Almazrouei and Merouane Debbah}

\AuthorCitation{Li, W.; Hacid, H.; Almazrouei, E.; Debbah, M.}
\address[1]{%
Technology Innovation Institute, P.O. Box 9639, \hl{Abu Dhabi} %MDPI:Please add the postal code (or ZIP code in the U.S.). If the postal code is not available, Post Office Box number can be added instead. -> po box added
, United Arab Emirates; hakim.hacid@tii.ae (H.H.); ebtesam.almazrouei@tii.ae (E.A.); merouane.debbah@tii.ae (M.D.)}

\corres{\hangafter=1 \hangindent=1.05em \hspace{-0.82em}Correspondence: wenbin.li@tii.ae}

\abstract{The union of Edge Computing (EC) and Artificial Intelligence (\textsf{AI}) has brought forward the \textsf{Edge AI} concept to provide intelligent solutions close to the end-user environment, for privacy preservation, low latency to real-time performance, and resource optimization. Machine Learning (\textsf{ML}), as the most advanced branch of \textsf{AI} in the past few years, has shown encouraging results and applications in the edge environment. Nevertheless, edge-powered ML solutions are more complex to realize due to the joint constraints from both edge computing and \textsf{AI} domains, and the corresponding solutions are expected to be efficient and adapted in technologies such as data processing, model compression, distributed inference, and advanced learning paradigms for \textsf{Edge ML} requirements. Despite the fact that a great deal of the attention garnered by Edge ML is gained in both the academic and industrial communities, we noticed the lack of a complete survey on existing Edge ML technologies to provide a common understanding of this concept.%Check meaning retained
To tackle this, this paper aims at providing a comprehensive taxonomy and a systematic review of \textsf{Edge ML} techniques, focusing on the soft computing aspects of existing paradigms and techniques. We start by identifying the \textsf{Edge ML} requirements driven by the joint constraints. We then extensively survey more than twenty paradigms and techniques along with their representative work, covering two main parts: edge inference, and edge learning. In particular, we analyze how each technique fits into \textsf{Edge ML} by meeting a subset of the identified requirements. We also summarize Edge ML frameworks
and open issues to shed light on future directions for Edge~ML.}

\keyword{\textls[15]{edge artificial intelligence; edge machine learning; distributed learning; distributed} \textls[15]{inference; federated learning; split learning; transfer learning; model approximation; model compression}} 

\begin{document}
%%%%%%%%%%%%%%%%%%%%%%%%%%%%%%%%%%%%%%%%%%

\section{Introduction}
The tremendous success of Artificial Intelligence (\textsf{AI}) technologies~\cite{Clark2022} in the past few years has been driving both industrial and societal transformations through domains such as Computer Vision (\textsf{CV}), Natural Language Processing (\textsf{NLP}), Robotics, Industry 4.0, Smart Cities, etc. This success is mainly brought by deep learning, providing the conventional Machine Learning (ML) techniques with the capabilities of processing raw data and discovering intricate structures~\cite{Lecun2015}. Daily human activities are now immersed with \textsf{AI}-enabled applications from content search and service recommendation to automatic identification and knowledge~discovery.

%\subsection{Context}
The existing \textsf{ML} models, especially deep learning models, such as GPT-4~\cite{OpenAI2023}, Segment Anything~\cite{Kirillov2023}, OVSeg~\cite{Liang2022a}, Make-A-Video~\cite{Singer2022}, and Stable Diffusion~\cite{StableDiffision2}, tend to rely on complex model structures and large model size to provide competitive performances. For~instance, the~largest \textsf{WuDao 2.0} model~\cite{Romero} trained on 4.9TB of data has surpassed state-of-the-art levels on nine benchmark tasks with a striking 1.75 trillion parameters. As~a matter of fact, large models have clear advantages on multi-modality, multi-task, and~benchmark performance. However, such models  require a relatively very large training datasets to be built as well as a large amount of computing resources during the training and inference phases. This dependency makes them usually closed to public access, and~unsuitable to be directly deployed for end devices or even the small/medium enterprise level to provide real-time, offline, or~privacy-oriented~services. 

In parallel with \textsf{ML} development, Edge Computing (\textsf{EC}) was firstly proposed in 1990~\cite{Dilley2002}. The~main principle behind \textsf{EC} is to bring the computational resources at locations closer to end-users. This was intended to deliver cached content, such as images and videos, that are usually communication-expensive, and~prevent heavy interactions with the main servers. This idea has later evolved to host applications on edge computing resources~\cite{Davis2004}. The~recent and rapid proliferation of connected devices and intelligent systems has been further pushing \textsf{EC} %\sout{edge computing} 
from the traditional base station level or the gateway level to the end device level. This offers numerous technical advantages such as low latency, mobility, and~location awareness support to delay-sensitive applications~\cite{Khan2019}. This serves 
%\sout{serving} 
as a critical enabler for emerging technologies like 6G, extended reality, and~vehicle-to-vehicle communications, to~mention only a~few.

Edge  \textsf{ML}~\cite{Lee2018}, as~the ML instantiation powered by \textsf{EC} and a union of \textsf{ML} and \textsf{EC}, has brought the processing in \textsf{ML} to the network edge and adapted \textsf{ML} technologies to the edge environment. In~this work, the edge environment refers to the end-user side pervasive environment composed of devices from both the base station level and the end device level.
In classical \textsf{ML} scenarios, users run \textsf{ML} applications on their resource-constrained devices (e.g., mobile phones, and~Internet of Things (IoT) sensors and actuators), while the core service is performed on the cloud server.
%
%\sout{Comparing to the classic \textsf{ML} scenario where users run \textsf{ML} applications on their resource-constrained devices (e.g., mobile phone and PC) while the core service and processing are performed on cloud servers, edge } 
In  \textsf{Edge ML}, either optimized models and services are
%\sout{directly} 
deployed and executed 
%\sout{optimized models and services} 
in the end-user's device, or~the \textsf{ML} models are directly built on the edge  side. This computing paradigm provides ML applications with advantages 
%\sout{so as to provide} 
such as real-time immediacy, low latency, offline capability, enhanced security and privacy, etc. 
%\sout{ to complement the large \textsf{ML} models performance with edge  computing advantages. }
%

\hl{In order} %We removed the color of text. Please confirm this revision. -> ok
 to illustrate the transformative potential and versatility of Edge ML across different sectors, we briefly explore several application sectors where Edge ML has already demonstrated substantial impact. The~applications represent just a fraction of the potential of Edge ML, while the breadth and depth of Edge ML applications are expanding with the technique's evolution and its increased adoption.%Check meaning retained
\begin{itemize}
    \item \textbf{\hl{Healthcare:} %Is the bold necessary? same as follows -> yes, please keep, if possible. 
} In healthcare, Edge ML enables real-time patient monitoring and personalized treatment strategies. Wearable sensors and smart implants equipped with Edge ML can process data locally, providing immediate health feedback~\cite{Amin2021}. This advancement permits the early detection of health irregularities and swift responses to potential emergencies, while also maintaining patient data privacy by avoiding the need for data transmission for analysis. Furthermore, in~telemedicine, Edge ML could be used to interpret diagnostic imaging locally and provide immediate feedback to remote healthcare professionals, improving patient care efficiency and outcomes.

    \item \textbf{\hl{Autonomous Vehicles:}} Edge ML is a key enabler for the advancements in the field of autonomous vehicles, which includes both Unmanned Aerial Vehicles (UAVs) and self-driving cars. These vehicles are packed with a myriad of sensors and cameras that generate enormous amounts of data per second~\cite{Yang2021a}. Processing this data in real-time is crucial for safe and efficient operation, and~sending all the data to a cloud server is impractical due to latency and bandwidth constraints. Edge ML, with~its capability to process data at the edge, can help in reducing the latency, and~enhancing real-time responses.

    \item \textbf{\hl{Smart City:}} Edge ML plays a crucial role in the realization of smart cities~\cite{Lv2021}, where real-time data processing is paramount. Applications such as intelligent traffic light control, waste management, and~urban planning greatly benefit from ML models that can analyze sensor data on-site and respond promptly to changes in the urban environment. Moreover, Edge ML can power public safety applications such as real-time surveillance systems for crime detection and prevention. Here, edge devices like surveillance cameras equipped with ML algorithms can detect unusual activities or behaviours and alert relevant authorities in real time, potentially preventing incidents and enhancing overall city safety.

    \item \textbf{\hl{Industrial IoT (IIoT):}} In the realm of Industrial IoT~\cite{Tang2022}, Edge ML is instrumental in predictive maintenance and resource management. With~ML models running at the edge, real-time anomaly detection can be carried out to anticipate equipment failures and proactively schedule maintenance. Additionally, Edge ML can optimize operational efficiency by monitoring production line performance, tracking resource usage, and~automating quality control processes.
  
\end{itemize}

However, the~Edge \textsf{ML}’s core research challenge remains how to adapt \textsf{ML} technologies to edge  environmental constraints such as limited computation and communication resources, unreliable network connections, data sensitivity, etc., while keeping similar or acceptable performance. Research work 
%\sout{have been done} 
was carried out in the past few years tackling different aspects of this meta-challenge, such as:
%\sout{, just to name a few}
 model compression~\cite{Cheng2017}, 
%\sout{is being studied to reduce the existing trained model size without sacrificing much accuracy in order to more easily fit the model into edge  devices;} 
%dimensionality reduction~\cite{Reddy2020},  
%\sout{has been investigated to reduce the input data dimension that is later processed by \textsf{ML} models so as to reduce the required computing resources during the training and the inference phase; to directly train \textsf{ML} models on the edge  environment,}  
transfer learning~\cite{Zhuang2021}, 
%\sout{ accelerates the training process by preserving previously learned knowledge from existing models and adapted models to other related tasks or different data distributions. Instead of training an \textsf{ML} model from scratch, transfer learning can significantly accelerate the time and cost to build a model by reducing the use of computing resources and data; at last, works on} 
and federated learning~\cite{Abreha2022}. 
%\sout{process sensitive data distributed in edge  devices locally in each device and only aggregate local calculated weights to build global model for the sake of privacy.} 

%\subsection{Methodology and Contribution}
With the above-mentioned promising results in diverse areas, we noticed that very little work has been realized to deliver a systematic view of relevant Edge ML techniques, rather focusing on Edge ML in specific contexts. %Check meaning retained
One example worth reporting is Wang~et~al.~\cite{Wang2020a,Wang2020}, 
%\sout{the work in }
who present a comprehensive survey on the convergence of edge computing and deep learning, which covers aspects of hardware, communication, models, as~well as edge  applications and edge  optimization. The~work is a good reference as an Edge ML technology stack. %check meaning retained
On the other hand, the~analysis of edge ML paradigms are rather brief without a comprehensive analysis of diverse related problems and the matching solutions. %check meaning retained
Abbas~et~al.~\cite{Abbas2022} review the role and impact of the relevant ML techniques in addressing the safety, security, and privacy challenges in the specific context of the IoT systems.  
Mustafa~et~al.~\cite{Mustafa2022} center around two significant themes in edge computing: Wireless Power Transfer (WPT) and Mobile Edge Computing (MEC). Their~work surveys the methodologies of offloading tasks in MEC and WPT to end devices, and~analyzes how the conjunction of WPT and MEC offloading can help overcome limitations in smart device battery lifetime and task execution delay.
Murshed~et~al.~\cite{SarwarMurshed2022a} introduce a machine learning survey at the network edge, for which the training, inference, and deployment aspects are briefly summarized, and~the technique coverage is limited and only focuses on representative techniques such as federated learning and~quantization. 

{Compared to the existing works that briefly review the representative techniques, our paper aims to provide a panoramic view of Edge ML requirements, and~offers a comprehensive technique review for edge machine learning on the soft computing aspects of model training and model inference. Throughout our review process, we strive for comprehensiveness, including more than twenty technique categories and over fifty techniques in this paper. Our work fills a significant gap in the literature by providing a single point of reference that offers extensive coverage of the Edge ML field. The~paper delivers a more complete picture of the landscape of Edge ML, allowing readers to understand the full breadth and depth of the available techniques, their respective advantages and limitations, and~their fit within different Edge ML contexts.}

{In contrast with~\cite{Mustafa2022}, our paper concentrates on the integration of ML techniques in the edge environment, dealing with the aspects of data processing, model compression, distributed inference, and~advanced learning paradigms to explore a broader range of techniques and paradigms. In~comparison with~\cite{SarwarMurshed2022a}, which covers representative works in topics of training, inference, applications, frameworks, software, and hardware, our paper focuses on model training and inference computing aspects, including a far richer list of techniques (e.g., in~the edge learning section, we include thirteen technique categories benefiting Edge ML, compared to the three training techniques presented in~\cite{SarwarMurshed2022a}). We also analyze in detail the Edge ML requirements to provide a broad taxonomy and show how each technique can satisfy different Edge ML requirements as a systematic review.}

\newpage
Specifically, our paper answers the three following questions: 
 
\begin{itemize}
	\item What are the computational and environmental constraints and requirements for \textsf{ML} on the edge?	
       {\item What are the Edge ML techniques to train intelligent models or enable model inference while meeting Edge ML requirements?}
	\item How can existing \textsf{ML} techniques  fit into an edge environment regarding these requirements?
\end{itemize}
		
To answer the three above questions, this review is realized by firstly identifying the Edge ML requirements, and~then individually reviewing existing ML techniques and analyzing if and how each technique can fit into edge by fulfilling a subset of the requirements. Following this methodology, our goal is to be as exhaustive as possible in the work coverage and provide a panoramic view of all relevant Edge ML techniques with a special focus on machine learning for model training and inference at the edge. Other topics, such as Edge ML hardware~\cite{Li2020} and edge communication~\cite{Wang2017b}, are beyond the scope of this paper. As~such, we do not discuss them in this~review.

%\subsection{Paper organization}
The remainder of the paper is organized as follows: Section~\ref{sec:EdgeML:Requirements} introduces the Edge ML motivation driven by the requirements. 
Section~\ref{Technology Overview} provides an overview of all the surveyed edge ML techniques.
In Sections~\ref{sec:edgeinference} and~\ref{sec:edgelearning}, we describe each technique and analyze them, respectively, in relation to Edge ML requirements. Section~\ref{sec:TechniqueReviewSummary} summarizes the technique review part, and~Section~\ref{sec:EdgeMLFramework} briefly introduces the frameworks supporting \textsf{Edge ML} implementation.
Section~\ref{sec:OpenIssues} identifies the open issues and future directions in Edge ML. Section~\ref{sec:Conclusion} concludes our work and sheds light on future~perspectives.
	
\section{Edge  \textsf{Machine~Learning}: Requirements}
\label{sec:EdgeML:Requirements}
In the context of machine learning, be it supervised learning, unsupervised learning, or~a reinforcement learning, an~\textsf{ML} task could be either a training or an inference. 
%
%\sout{is either a training process or inference with a fixed amount of data samples for the case of supervised learning and unsupervised learning (e.g., clustering and density estimation), or the process of reaching a predefined goal in the case of reinforcement learning. }
%
As in every technology, it is critical to understand the underlying requirements that ensure proper expectations.
By definition, the~edge  infrastructure is generally resource-constrained in terms of the following: %\sout{hardware for computing} 
computation power, i.e.,~processor and memory;
%\sout{capacity}, 
storage capacity, i.e.,~auxiliary storage; and~communication capability, i.e.,~network bandwidth.  %\sout{(i.e., network connectivity and bandwidth), while }. 
\textsf{ML} models, on the other hand, are commonly known to be hardware-demanding, with computationally expensive and memory-intensive features. Consequently, the~union of \textsf{EC} and \textsf{ML}  exhibits both constraints from edge  environment and \textsf{ML} models. When designing edge-powered \textsf{ML} solutions, requirements from both the hosting environment and the \textsf{ML} solution itself need to be considered and fulfilled for suitable, effective, and~efficient results. 
%
%\sout{Before introducing the Edge ML requirements, we hereby briefly define the concept of \textsf{ML} task.} 

\begin{figure}[H]
	%\centering
	\includegraphics[width=9cm]{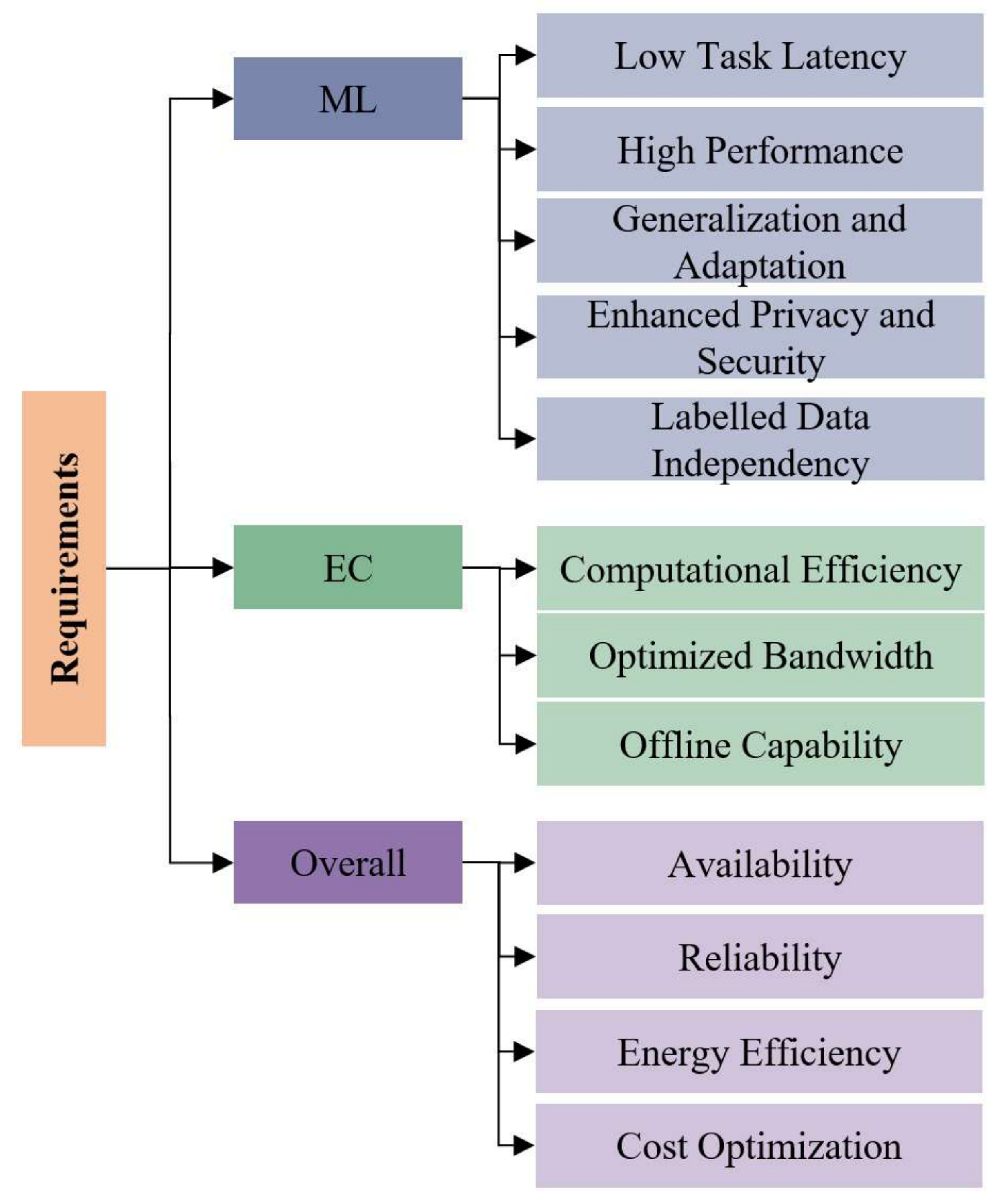}
	\caption{{\hl{Edge} %We removed the color of caption. Please confirm this revision. -> ok
 \textsf{ML} Requirements.}}
	\label{f1requirements}
\end{figure}

We introduce in this section the Edge ML requirements, 
%\sout{are introduced as} 
structured in three categories: 
%\sout{ three relative parts}: 
(i) \textsf{ML} requirements, (ii) EC requirements, and~(iii) overall requirements, which are composite indicators from \textsf{ML} and EC for Edge ML performance. 
% It is worth mentioning that the general quality of service attributes, e.g.,~availability and reliability, are always relevant but not listed here. This is because they are applicable to all services but not directly related to \textsf{Edge ML}. 
The three categories of requirements are summarized in Figure~\ref{f1requirements}.

\subsection{\textsf{ML} Requirements} 
We foresee five main requirements an \textsf{ML} system should consider: (i) Low Task Latency, (ii) High Performance, (iii) Generalization and Adaptation, (iv) Labelled Data Independence, and~(v) Enhanced Privacy and Security. We detail these in the~following.

\begin{itemize}
    \item \textbf{\hl{Low Task Latency:}} Task latency refers to the end-to-end processing time for one \textsf{ML} task, in~seconds (s), and~is  determined by both \textsf{ML} models and the supporting  computation infrastructure. Low task latency is important to achieve fast or real-time \textsf{ML} capabilities, especially for time-critical use-cases such as autonomous driving.  We use the term task latency instead of latency to differentiate this concept from communication latency, which describes the time for sending a request and receiving an answer.%\sout{is introduced later}.

    \item \textbf{\hl{High Performance:}} The performance of an \textsf{ML} task is represented by its results and measured by general performance metrics such as top-n accuracy, %\sout{precision, recall} 
    and f1-score in percentage points (pp),  as~well as use-case-dependent benchmarks such as General Language Understanding Evaluation (GLUE) benchmark for \textsf{NLP}~\cite{Wang2018a} or Behavior Suite for reinforcement learning~\cite{Osband2019}. 
    
    \item \textbf{\hl{Generalization and Adaptation:}} The models are expected to learn the generalized representation of data instead of the task labels, so as to be easily generalized to a domain instead of specific tasks. This brings the models' capability to solve new and unseen tasks and realize a general \textsf{ML} directly or with a brief adaptation process. Furthermore, facing the disparity between learning and prediction environments, \textsf{ML} models can be quickly adapted to specific environments to solve the environmental specific~problems.

    \item \textbf{\hl{Enhanced Privacy and Security:}} The data acquired from edge  carry much private information, such as personal identity, health status, and~messages, preventing these data from being shared in a large extent. In~the meantime, frequent data transmission over a network threatens data security as well. The~enhanced privacy and security requires the corresponding solution to process data locally and minimize the sha\mbox{red~information}.  

    \item \textbf{\hl{Labelled Data Independence:}} The widely applied supervised learning in modern machine learning paradigms requires large amounts of data to train models and generalize knowledge for later inference. However, in~practical scenarios, we cannot assume that all data in the edge  are correctly labeled. The~independence of labelled data indicates the capability of an Edge ML solution to solve one \textsf{ML} task without labelled data or with few labelled data. 
\end{itemize}

\vspace{-6pt}

\unskip

\subsection{EC~Requirements} 
Three main edge  environmental requirements of EC impact the overall \textsf{Edge ML} technology: (i) Computational Efficiency, (ii) Optimized Bandwidth, and~(iii) Offline Capability, summarized~below.

\begin{itemize}
    \item \textbf{\hl{Computational Efficiency:}} Refers to the efficient usage of computational resources to complete an \textsf{ML} task. This includes both processing resources measured by the number of arithmetic operations (OPs), and~the required memory measured in MB. %\sout{, where an arithmetic operation is either an addition or multiplication.}

    \item \textbf{\hl{Optimized Bandwidth:}} Refers to the optimization of the amount of data transferred over network per task, measured by MB/Task. Frequent and large data exchanges over a network can raise communication and task latency. An~optimized bandwidth usage expects Edge ML solutions to balance the data transfer over the network and local data~processing.  

    \item \textbf{\hl{Offline Capability:}} The edge connectivity of edge  devices is often weak and/or unstable, requiring operations to be performed on the edge  directly. The~offline  capability refers to the ability to solve an \textsf{ML} task when network connections are lost or without a network~connection. 

\end{itemize}

\subsection{Overall~Requirements}
 
The global requirements  are composite indicators from \textsf{ML} and environmental requirements for Edge ML performance. We specify four overall requirements in this category: (i)~Availability, (ii) Reliability, (iii) Energy Efficiency, and~(iv) Cost~Optimization.

\begin{itemize}

    \item {\textbf{\hl{Availability:}} Refers to the percentage of time (in percentage points (pp)) that an Edge ML solution is operational and available for processing tasks without failure. For~edge ML applications, availability is paramount because these applications often operate in real-time or near-real-time environments, and~downtime can result in severe operational and productivity loss.}
    
    \item {\textbf{\hl{Reliability:}}  Refers to the ability of a system or component to perform its required functions under stated conditions for a specified period of time. Reliability can be measured using various metrics such as Mean Time Between Failures (MTBF) and Failure Rate.}
    
    \item \textbf{\hl{Energy Efficiency:}} Energy efficiency refers to the number of \textsf{ML} tasks obtained per power unit, in~Task/J. The~energy efficiency is determined by both the computation and communication design of Edge ML solutions and their supporting~hardware. 

    \item \textbf{\hl{Cost optimization:}} Similar to energy consumption, edge  devices are generally low-cost compared to cloud servers. The~cost here refers to the total cost of realizing one \textsf{ML} task in an edge  environment. This is again determined by both the Edge ML software implementation and its supporting infrastructure~usage.

    %\wl{\sout{\textbf{High Reliability and Availability (WL-remove?):} Reliability quantifies the likelihood of Edge ML solutions to be functional as intended without disruptions or downtime. The availability is \sout{the measure of } the percentage of time an Edge ML solution is operable. Both reliability and availability are measured \sout{by} in percentage points (pp).}} 
\end{itemize}

It should be noted that, depending on the nature of Edge ML applications, one Edge ML solution does not necessarily fulfill all the requirements above. The~exact requirements for each specific Edge ML application vary according to each requirement’s critical level to an application. For~example, for~autonomous driving, the~task latency requirement is much more critical than the power consumption and cost optimization~requirements.

\section{Techniques~Overview}	
\label{Technology Overview}

\begin{figure}[H]
	\centering
	\includegraphics[width=\textwidth]{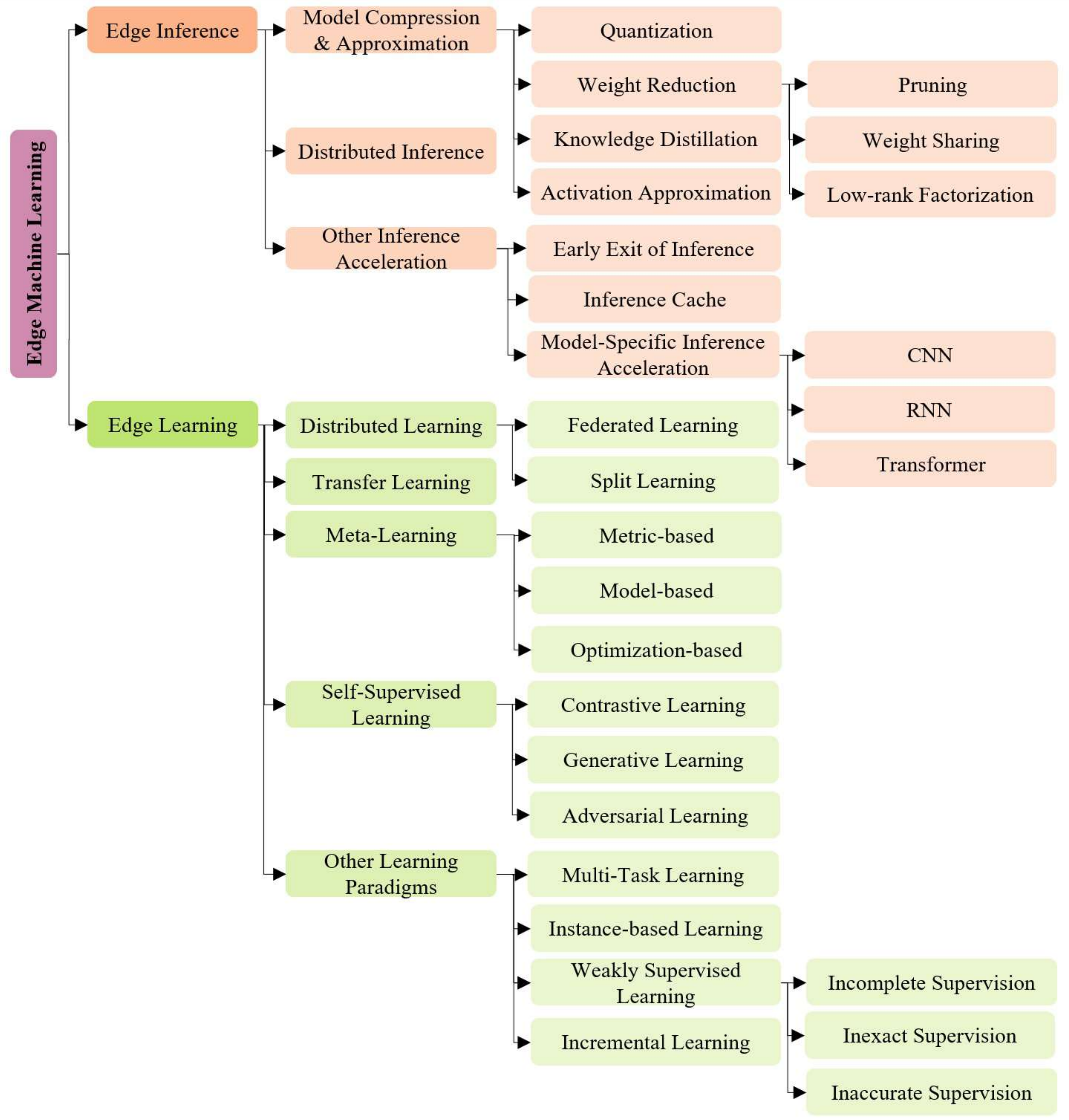}
	\caption{Edge ML Technique~Overview.}
	\label{fig:techno:overview}
\end{figure}

Figure~\ref{fig:techno:overview} shows a global view of edge  Machine Learning techniques reviewed in this paper. We structure the related techniques into: 
%\sout{ consisting of three categories:} 
(i) edge  inference, and~(ii) edge  learning. 
The edge  inference category introduces the technologies to accelerate the task latency of \textsf{ML} model inference. This is performed through, e.g.,~compressing existing models to consume less hardware resources or by dividing existing models into several parts for parallel inference collaboration. The~edge  learning category introduces solutions to directly build \textsf{ML} models on the edge  side by learning locally from edge  data. We detail the categories in the next~sections. 

%====================================================================================
% I put 5 comments signs just in case we want to bring back the below content
%====================================================================================
%%%%\Hed{\textbf{H2W: I suggest we put  this section on comment. We do not need to discuss those concepts as they are expected to be known. However, depending on the next section, we could put here a notation section that introduces the notations that are used throughout the paper. I will advise once I progress further.  }}

Before introducing the details of each reviewed technique, we go through three basic machine learning paradigms, i.e.,~supervised learning, unsupervised learning, and~reinforcement learning, to~lay the theoretical foundation of ML. Briefly, supervised learning involves using an \textsf{ML} model to learn a mapping function between input data and the target variable from the labeled dataset. Unsupervised learning directly describes or extracts relationships in unlabeled data without any guidance from labelled data. Reinforcement learning is the process that an \textsf{ML} agent continuously interacts with its environment, performs actions to obtain awards, and~learns to achieve a goal by the trial-and-error~method. 

Extending the work from~\cite{Huisman2021}, we give below the formal definition of the three basic learning paradigms. {The objective of the definition is not merely to offer a conventional understanding of these methods, but~more importantly, to~create a conceptual bridge between these mainstream learning techniques and their specialized applications and adaptations for edge learning techniques, which helps to understand how these well-established AI techniques transform when applied to the context of Edge computing.} Breakthroughs have been made in all three ML learning paradigms to derive meaningful data insights and bring intelligent capabilities, and the reviewed techniques in this paper all fit into the three general machine learning~paradigms. 
	
%Extra content from Word Version - 3 more references
\subsection{Supervised~Learning}

Supervised learning learns a function $ f_\theta : X \rightarrow Y $ mapping inputs $ x_i \in X $ to the corresponding outputs $ y_i \in Y $ with the help of a labeled dataset $D$ of $m$ samples \mbox{$ D = \{(x_i, y_i)\}_{i=1}^m $}, in~which $\theta$ are \textsf{ML} model parameters (e.g., weights and biases in the case of neural network). The~learning process aims at finding optimal or sub-optimal values for $\theta$ specific to the dataset $D$ that minimizes an empirical loss function $L_D$ through a training process (e.g., backward propagation in the case of neural network) as:
\begin{equation}
	\label{supervised_learning_parameters}
	\theta_{SL} := \underset{\theta}{arg \, min}\ L_D(\theta),
\end{equation}
where $SL$ stands for ``supervised learning''. In~practice, the~labelled dataset $D$ is often divided into training, validation, and testing datasets $D^{tr}, \ D^{val}, \ D^{test} $ to train the model, guide the training process, and evaluate model performance after training, respectively~\cite{Xu2018}. 

Finding globally optimal values of $\theta_{SL}$ is computationally expensive, while in practice the training process is commonly an approximation to find sub-optimal $\theta_{SL}$ values guided by a predefined meta-knowledge $\omega$ including the initial model parameters $\theta$, the~training optimizer, and learning rate in the case of neural network, as:
\begin{equation}
	\label{supervised_learning_knowledge}
	\theta_{SL} \approx g_\omega(D, L_D),
\end{equation}
where $g_\omega$ is an optimization procedure that uses predefined meta-knowledge $\omega$, dataset $D$ and loss function $L_D$ to continuously update the model's %Check meaning retained
parameters $\theta$ and output final $\theta_{SL}$. 

	\subsection{Unsupervised~Learning}

Training an \textsf{ML} model in the unsupervised manner is mostly similar to the supervised learning processing, except~that the learned function  $ f_\theta : X \rightarrow X $ mapping input $ x_i \in X $ to the same input $ x_i $ or other inputs. Unsupervised learning only uses unlabeled dataset $ \bar{D} $ of $n$ sample $ \bar{D} = \{(x_i)\}_{i=1}^n $ to determine $\theta$ values specific to the dataset $\bar{D}$ that minimize an empirical loss function $ L_{\bar{D}}$ through a training process, as:
\begin{equation}
	\label{unsupervised_learning_parameters}
	\theta_{UL} := \underset{\theta}{arg \, min}\ L_{\bar{D}}(\theta),
\end{equation}
where $UL$ stands for ``unsupervised learning''. Furthermore, the~same approximation is applied to unsupervised learning to efficiently fit the $\theta_{UL}$ to $\bar{D}$:
\begin{equation}
	\label{unsupervised_learning_knowledge1}
	\theta_{UL} \approx g_\omega(\bar{D}, L_{\bar{D}})
\end{equation}

In addition to the above unsupervised learning paradigm which is used to train \textsf{ML} models, other unsupervised learning techniques such as clustering~\cite{Golalipour2021} apply predefined algorithms and computing steps to directly generate expected outputs (e.g., data clusters) from $\bar{D}$. In~such context, the~unsupervised learning approximates the values of specific algorithms' hyperparameters $\bar{\theta}_{UL}$ as:
\begin{equation}
	\label{unsupervised_learning_knowledge2}
	\bar{\theta}_{UL} \approx g_\omega(\bar{D}, L_{\bar{D}})
\end{equation}

	\subsection{Reinforcement~Learning}
		
In the classic scenario of reinforcement learning where agents know the state at any given time step, the~reinforcement learning paradigm can be formalized into a Markov Decision Process (MDP) as $M=(S, A, P, r, p_0, \gamma, T)$, where $S$ is the set of states, $A$ the set of actions, P the transition probability distribution defining $P(s_{t+1}|s_t,a_t)$ the transition probability from $s_{t}$ to $s_{t+1}$ via $a_t$, $r : S \times A \rightarrow \mathbb{R}$ the reward function, $p_0$ the probability distribution over initial states, $\gamma \in [0, 1]$ the discount factor prioritizing short- or long-term rewards by respectively decreasing or increasing it, $T$ the maximum number of time steps. At~a time step $t \in T$, a~policy function $\pi_\theta $, usually represented by a model in the case of deep reinforcement learning, is used to determine the action $a_t$ that an agent performs at state $s_{t}: a_t = \pi_\theta(s_t)$, where $\theta$ is the parameters of the policy function; after the action $a_t$, the~agent receives an award  $r_t = r(s_t, \pi_\theta(s_t)), r_t \in \mathbb{R}$ and enters into a new state $s_{t+1}$. The~interaction between agent and environment stops until a criterion is met, such as the rewards are~maximized.

The objective of the reinforcement learning is to make agents learn to act and maximize the received rewards as follows:
\begin{equation}
	\label{reinforcement_learning_parameters}
	\theta_{RL} := \underset{\theta}{arg \, min}\ \mathbb{E}_{traj} \sum_{t=1}^T\gamma^t r(s_t, \pi_\theta(s_t)),
\end{equation}
where $RL$ stands for ``reinforcement learning'', and~$\mathbb{E}_{traj}$ is the expectation over possible trajectories $traj = (s_0, \pi_\theta(s_0),\  ..., \ s_T, \pi_\theta(s_T))$. 

Similar to supervised and unsupervised learning, the sub-optimum of $\theta_{RL}$ are searched via an approximation process:
\begin{equation}
	\label{reinforcement_learning_knowledge}
	\theta_{RL} \approx g_\omega(M, L_M),
\end{equation}
where $g_\omega$ is an optimization procedure that uses predefined meta-knowledge $\omega$, the~given MDP $M$ and loss function $L_M$ to produce final $\theta_{RL}$.

\section{Edge~Inference}
\label{sec:edgeinference}

Edge inference techniques seek to enable large model inference on edge devices and accelerate the inference efficiency. The~techniques can be categorized into three main groups: (i)  model compression and approximation, (ii) distributed inference, and~(iii) other inference acceleration~techniques.

\subsection{Model Compression and~Approximation}
\label{sebsec:ModelCompressionandApproximation}
	
A large amount of redundancy among the \textsf{ML} model parameters (e.g., neural network weights) has been observed~\cite{Denil2013}, showing that a small subset of the weights is sufficient to reconstruct the entire neural network. Model compression and approximation are methods to transform \textsf{ML} models into smaller-size or approximate models with low-complexity computations. This is performed with  the objective to reduce the memory use and the arithmetic operations during the inference, while keeping acceptable performances. Model compression and approximation can be broadly classified into three categories~\cite{Wang2019}: (i)~Quantization, (ii) Weight Reduction, 
%\sout{(including pruning, knowledge distillation, weight sharing, low-rank factorization)} 
and (iii) Activation Function Approximation. We discuss these categories in the following: 

\subsubsection{Quantization}
%\textbf{Theory.} 
Quantization is the process of converting \textsf{ML} model parameters $\theta$ (i.e., weights and bias in neural networks) and activation outputs, represented in Floating Point (FP) format of high precision such as FP64 or FP32, into~a low-precision format and then perform computing tasks such as training or inference. Different formats of quantization can be summarized as: 

\begin{itemize}
	\item \textbf{\hl{Low-Precision Floating-Point Representation}}: A floating-point parameter describes binary numbers in the exponential form with an arbitrary binary point position such as 32-bit floating point (FP32), 16-bit floating point (FP16), 16-bit Brain Floating Point (BFP16)~\cite{Wang2021}.
	
	\item \textbf{\hl{Fixed-Point Representation}}: A fixed-point parameter~\cite{Goyal2021} uses predetermined precision and binary point locations. Compared to a high-precision floating-point representation, the~fixed-point parameter representation can offer faster, cheaper, and~more power-efficient arithmetic~operations.
	
	\item \textbf{\hl{Binarization and Terrorization}}: Binarization~\cite{Yuan2021a} is the quantization of parameters into just two values, typically {\hl{-}%MDPI: Please check if this should be a multiplication sign -> it should be the minus sign 
1, 1} with a scaling factor. The~terrorization~\cite{Li2016a} on the other hand adds the value 0 to the binary value set to express 0 in~models.
	
	\item \textbf{\hl{Logarithmic Quantization}}: In a logarithmic quantization~\cite{Lee2017}, parameters are quantized into powers of two with a scaling factor. Work in~\cite{Lai2017} shows that a weight’s representation range is more important than its precision in preserving network accuracy. Thus, logarithmic representations can cover wide ranges using fewer bits, compared to the other above-mentioned linear quantization formats.
\end{itemize}

In addition to the main exploited data types, AI-specific data formats and several quantization contributions exist in the literature and are introduced in Table~\ref{tab:quantization}. 

\renewcommand{\arraystretch}{1.35}
\begin{table}[H]
\caption{{\hl{AI-specific} %MDPI: please check if the alignment keep and if vertical line can be removed? -> the alignment keeps, vertical lines are removed. 
 data formats and model quantization works.}}
\small %
\begin{adjustwidth}{-4.5cm}{0cm}
\centering
\begin{tabular}{p{4.5cm} p{6.5cm} p{6cm}}
\noalign{\hrule height 1pt}
\textbf{Work} & \textbf{Contribution} & \textbf{Results and Insights} \\
\hline
Posit Representation~\cite{Gustafson_Yonemoto_2017} & The Posit is a data type designed to supersede IEEE Standard 754 floating-point numbers~\cite{4610935}. & Provides larger dynamic range and higher accuracy than traditional floats, along with simpler hardware implementation and exception handling.\\
\hline
Fixed-Posit Representation~\cite{Gohil2021} & Extended the Posit data type by maintaining a fixed count of the bits, unlike conventional posits that have a configurable number of regime and exponent bits. & Enhances the power and area efficiency compared to both Posit and 32-bit IEEE-754 floating-point representations. Improvements of up to 70\% in power, 66\% in area, and~26\% in delay.\\
\hline
Tensor Cores~\cite{tensorcores1} & NVIDIA's Tensor Cores is specialized hardware designed for performing the tensor/matrix computations with mixed-precision computations, enabling FP8 in the Transformer Engine, Tensor Float 32 (TF32)~\cite{tf321}, and~FP16. & Provide an order-of-magnitude higher performance with reduced precisions\\
\hline
8-bit Quantization~\cite{Jacob2018} & An 8-bit quantization schema for MobileNet~\cite{Howard2017} on the ARM NEON-based implementation. & Model size reduction: 4x. Inference task latency reduction: up to 50\%. Accuracy drop: 1.8\% for COCO datasett~\cite{Lin2014}.\\ 
\hline
SLQ~\cite{Lee2019} & A successive logarithmic quantization (SLQ) scheme to quantize the training error again when the quantization error is higher than a certain threshold. & Accuracy drop: less than 1.5\% for AlexNet~\cite{NIPS2012_c399862d}, SqueezeNet~\cite{Iandola2016}, and~VGG-S~\cite{Jin2021} at 4 to 5-bit weight representation.\\
\hline
SLQ Training~\cite{Oh2021} & A specific training method for the Successive Logarithmic Quantization (SLQ) scheme. & Performance degradation of around 1\% at 3-bit weight quantization.\\
\hline
Binary Network~\cite{Qin2022} & An accurate and efficient binary neural network for keyword-spotting applications, along with a binarization-aware training method for optimization. & Impressive 22.3 times speedup of task latency and 15.5 times storage-saving with only less than 3\% accuracy drop on Google Speech Commands V1-12 task~\cite{Warden2018}.\\
\hline
Binary Distilled Transformer~\cite{liu2022bit} & A binarized multi-distilled transformer including a two-set binarization scheme, an~elastic binary activation function with learned parameters, and~a method for successively distilling models. & Accuracy of fully binarized transformer models approaches a full-precision BERT baseline on the GLUE language understanding benchmark within as little as 5.9\%.\\
\hline
FMA~\cite{9823406} & A new class of Fused Multiply--Add (FMA) operators built on BFP16 arithmetic while maintaining accuracy comparable to that of the standard FP32. & Improved performance by 1.28\hl{--}%MDPI: please check if it should be en dash? -> changed.
1.35\hl{$\times$} %MDPI: We revised the letter “x” into a multiplication sign. Please confirm -> ok
 on ResNet compared to FP32.\\
\noalign{\hrule height 1pt}
\end{tabular}
\label{tab:quantization}
\end{adjustwidth}
\end{table}

To produce the corresponding quantized model, post-training quantization and quantization-aware training can be applied. 
Given an existing trained model, post-training quantization directly converts the trained model parameters and/or activation according to the conversion needs, to~reduce model size and improve task latency during the inference phase.
On the other hand, and~instead of quantizing existing models, quantization-aware training is a method that trains an \textsf{ML} model by emulating inference-time quantization, which has proved to be better for model accuracy~\cite{Zhang2022}. During~the training of a neural network, quantization-aware training simulates low-precision behavior in the forward pass, while the backward pass based on backward propagation remains the same. The~training process takes into account both errors from training data labels as well as quantization errors which  accumulate in the total loss of the model, and~hence the optimizer tries to reduce them by adjusting the parameters accordingly. %Check meaning retained
{Similarly, the~work~\cite{9996763} analyzes the significant trade-off between energy efficiency and model accuracy, showing that the application of repair methods (e.g., ReAct-Net~\cite{10.1007/978-3-030-58568-6_9}), could largely offset the accuracy loss after quantization. Zhou~et~al.~\cite{Zhou2016} analyzed various data precision combinations, concluding that accuracy deteriorates rapidly when weights are quantized to fewer than four bits. The~work~\cite{RUOSPO2021104318} investigates the impact of data representation and bit-width reduction on CNN resilience, particularly in the context of safety-critical and resource-constrained systems. The~results indicate that fixed-point data representation offers a superior trade-off between memory footprint reduction and resilience to hardware faults, especially for the LeNet-5 network, achieving a 4x memory footprint reduction at the expense of less than 0.45\% critical faults without requiring network retraining.}

 Overall, moving from high-precision floating-point to lower-precision data representations is especially useful for \textsf{ML} models on edge devices with only low precision operation support such as Application-Specific Integrated Circuit (ASIC) and Field Programmable Gate Arrays (FPGA) to facilitate the trade-off between task accuracy and task latency. Quantization reduces the precision of parameters and/or activation, and~thereby decreases the inference task latency by reducing the consumption of computing resources, while the workload reduction brought by cheaper arithmetic operations leads to energy and cost optimization as well. {Moreover, quantization techniques make it feasible to run models on low-resource edge devices, increasing the availability of the application by allowing it to function with low connectivity. One the other hand, the~techniques can also reduce the application robustness or resilience to hardware faults. Reduced precision leads to a model that is more susceptible to errors due to slight changes in the input or due to hardware faults, and~thus decreases the reliability.}

\subsubsection{Weight~Reduction}
Weight reduction is a class of methods that removes redundant parameters from $\theta$ through pruning and parameter approximation. We reviewed the three following categories of methods in this paper: 

\begin{itemize}
	\item \textbf{\hl{Pruning}}. The~process of removing redundant or non-critical weights and/or nodes from models~\cite{Cheng2017}: weight-based pruning removes connections between nodes (e.g., neurons in neural network) by setting relevant weights to zero to make the \textsf{ML} models sparse, while node-based pruning removes all target nodes from the \textsf{ML} model to make the model~smaller.
	
	\item \textbf{\hl{Weight Sharing}}. The~process of grouping  similar model parameters into buckets and reuse shared weights in different parts of the model to reduce model size or among models~\cite{Chu2021} to facilitate the model structure~design.
	
	\item\textbf{\hl{Low-rank Factorization}}. The~process of  decomposing the weight matrix into several low-rank matrices by uncovering explicit latent structures~\cite{Jaderberg2014}.
\end{itemize}

 A node-based pruning method is introduced in~\cite{Srinivas2015} to remove redundant neurons in trained CNNs. In~this work, similar neurons are grouped together following a similarity evaluation based on squared Euclidean distances and then pruned away. Experiments showed that the pruning method can remove up to 35\% nodes in AlexNet with a 2.2\% accuracy loss on the dataset of ImageNet~\cite{ILSVRC15}. 
A grow-and-prune paradigm is proposed in~\cite{Dai2019} to complement network pruning to learn both weights and compact Deep neural networks(DNNs) architectures during training. The~method iteratively tunes the architecture with gradient-based growth and pruning of neurons and weight. Experimental results showed the compression ratio of 15.7\hl{$\times$} 
 and 30.2\hl{$\times$} for AlexNet and VGG-16 network, respectively. This delivers a significant parameter and arithmetic operation reduction relative to pruning-only methods. %Check meaning retained -> ok.
In practice, pruning is often combined with a post-tuning or a retraining process to improve the model accuracy after pruning~\cite{Yu2017}. A~Dense--Sparse--Dense training method is presented in~\cite{Han2017}, which introduces a post-training step to re-dense and recover the original model symmetric structure to increase the model capacity. This was shown to be efficient as it improves the classification accuracy by 1.1\% to 4.3\% on ResNet-50~\cite{He2016}, ResNet-18~\cite{He2016}, and~VGG-16~\cite{Simonyan2015}. The~pruning method of SparseGPT is proposed in~\cite{Frantar2023}, showing that the large-scale generative pretrained transformer (GPT) family models can be pruned to at least 50\% sparsity in one-shot, without~any retraining, at~minimal loss of accuracy. The~driving idea behind this is an approximate sparse regression solver that runs entirely locally and relies solely on weight updates designed to preserve the input--output relationship for each layer. 
{We also examine the layer-wise weight-pruning method presented in~\cite{s21030880}. The~method relies on a differential evolutionary layer-wise weight pruning operating in two distinct phases---a model-pruning phase, which analyzes each layer's pruning sensitivity and guides the pruning process, and~a model-fine-tuning phase, where removed connections are considered for recovery to improve the model's capacity. Notably, the~approach achieved impressive compression ratios of at least 10\hl{$\times$} across different models, with~a standout 29\hl{$\times$} compression achieved for AlexNet.}
{Another notable development in the field of structural pruning is the Dependency Graph (DepGraph) method~\cite{Fang_2023_CVPR}. This method represents a breakthrough in tackling the complex task of any structural pruning across a broad variety of neural architectures, from~CNNs and RNNs to GNNs and Transformers. DepGraph introduces an automated system that models the dependency between layers to effectively group parameters that can be pruned together. The~method demonstrates a promising performance across a multitude of tasks and architectures, including ResNet, DenseNet, MobileNet, and Vision transformer for images, GAT for graph, DGCNN for 3D point cloud, alongside LSTM for language.}

The aforementioned pruning methods are static, as~they permanently change the original network structure which may lead to a decrease in model capability. On~the other hand, dynamic pruning~\cite{Liang2021} determines at run-time which layers, image channels (for CNN), or~neurons would not participate in further model computing during a task. 
A dynamic channel pruning is proposed in~\cite{Gao2019}. This method dynamically selects which channel to skip or to process using feature boosting and suppression, which is achieved by the use of a side network trained together along the CNN to guide channel amplification and omission. This work achieved a 2\hl{$\times$} acceleration on ResNet-18 with 2.54\% top-1 and 1.46\% top-5 accuracy~loss, respectively. 

A multi-scale weight-sharing method is introduced in~\cite{Aich2020} to share weights among the convolution kernels of the same layer. To~share kernel weights for multiple scales, the~shared tuple of kernels is designed to have the same shape, and~different kernels in the shared tuple are applied to different scales. With~approximately 25\% fewer parameters, the~shared-weight ResNet model provides similar performance compared to the baseline ResNets~\cite{He2016}. 
Instead of looking up tables to locate the shared weight for each connection, HashedNets is proposed in~\cite{Chen2015} to randomly group connection weights into hash buckets via a low-cost hash function. These weights are tuned to adjust to the HashedNets weight sharing architecture with standard back-propagation during the training. Evaluations showed that HashedNets achieved a compression ratio of 64\% with an around 0.7\% accuracy improvement against a five-layer CNN baseline with the MNIST dataset~\cite{LeCun1998c}. 
{Furthermore, the~recent work~\cite{Li2022c} uses a gradient-based method to determine a threshold for attention scores at runtime, thereby effectively pruning inconsequential computations without significantly affecting model accuracy. Their proposed bit-serial architecture, known as LeOPArd, leverages this gradient-based thresholding to enable early termination, resulting in a significant boost in computational speed (1.9\hl{$\times$} on average) and energy efficiency (3.9\hl{$\times$} on average), with~a minor trade-off in accuracy (<0.2\% degradation).} The related work of pruning is summarized in Table~\ref{tab:pruning}.

\begin{table}[H]
\caption{{\hl{Summary} %MDPI: Please cite the table in the text and ensure that the first citation of each table appears in numerical order. -> done
 of pruning works and main results.}}
\small %
%\resizebox{\columnwidth}{!}{
%\begin{adjustwidth}{-1cm}{-1cm}
\centering
\begin{tabular}{p{5cm} p{3cm} p{4.5cm}}
\noalign{\hrule height 1pt}
\textbf{\hl{Methods} %MDPI: please check if the alignment keep and if vertical line can be removed? we add bold to table header, please confirm -> ok with bold in table header, vertical lines are removed. 
} & \textbf{Target Models} & \textbf{Main Results} \\
\hline
Node-based Pruning~\cite{Srinivas2015} & CNNs & 35\% pruning, 2.2\% accuracy loss \\
\hline
Grow-and-Prune~\cite{Dai2019} & DNNs & 15.7$\times$, 30.2$\times$ compression \\
\hline
Dense-Sparse-Dense~\cite{Han2017} & ResNet, VGG-16 & 1.1\% to 4.3\% accuracy increase \\
\hline
SparseGPT~\cite{Frantar2023} & GPT models & 50\% sparsity \\
\hline
Differential Layer-Wise Pruning~\cite{s21030880} & LeNet, AlexNet, VGG16 & 10$\times$ to 29$\times$ compression \\
\hline
DepGraph Structural Pruning~\cite{Fang_2023_CVPR} & Any Model Structure & 8--16$\times$ speedup, $-$6\% to +0.3\% accuracy change\\
\hline
Dynamic Channel Pruning~\cite{Gao2019} & ResNet-18 & 2$\times$ acceleration, 2.54\% top-1, 1.46\% top-5 accuracy loss \\
\hline
Multi-Scale Weight Sharing~\cite{Aich2020} & ResNets & 25\% fewer parameters \\
\hline
HashedNets~\cite{Chen2015} & 5-layer CNN & 64\% compression, 0.7\% accuracy improvement \\
\hline
Gradient-Based Pruning~\cite{Li2022c} & Transformers & 1.9$\times$ speed, 3.9$\times$ energy efficiency, $<$0.2\% accuracy loss \\
\noalign{\hrule height 1pt}
\end{tabular}
%}
%\end{adjustwidth}
\label{tab:pruning}
\end{table}

Structured matrices use repeated patterns within matrices to represent model weights to reduce the number of parameters. The \hl{circulant} %Is the italics necessary? -> removed
 matrix, in~which all row vectors are composed of the same elements and each row vector is shifted one element to the right relative to the preceding row vector, are often used as the structured matrix to provide a good compression and accuracy for RNN-type models~\cite{Wang2017a,Wang2018}. The~Efficient Neural Architecture Search (Efficient NAS) via parameter sharing is proposed in~\cite{Pham2018}, in~which only one shared set of model parameters is trained for several model architectures, also known as child models. The~shared weights are used to compute the validation losses of different architectures. Sharing parameters among child models allows efficient NAS to deliver strong empirical performances for neural network design and use fewer GPU FLOP than automatic model design approaches. The~NAS approach has been successfully applied to design model architectures for different domains~\cite{Liu2021a} including CV and~NLP. 

As for low-rank factorization, to~find the optimal decomposed matrices to substitute the original weight matrix, Denton~et~al.~\cite{Denton2014} analyze three decomposition methods on pre-trained weight matrices: (i) singular-value decomposition, (ii) canonical polyadic decomposition, and~(iii) blustering approximation.
Experimental results on a 15-layer CNN demonstrate that singular-value decomposition achieved the best performance by a compression ratio of 2.4\hl{$\times$} to 13.4\hl{$\times$} on different layers along with a 0.84\% point of top-one accuracy loss in the ImageNet dataset. A~more recent work~\cite{Chen2021} proposes a data-aware low-rank compression method (DRONE) for weight matrices of fully-connected and self-attention layers in large-scale NLP models. As~weight matrices in NLP models, such as BERT~\cite{Devlin2019}, do not show obvious low-rank structures, a~low-rank computation could still exist when the input data distribution lies in a lower intrinsic dimension. The~proposed method considers both the data distribution term and the weight matrices to provide a closed-form solution for the optimal rank-k decomposition. Experimental results show that DRONE can achieve 1.92\hl{$\times$} speedup on the Microsoft Research Paraphrase Corpus (MRPC)~\cite{Dolan2005}  task with only 1.5\% loss in accuracy, and~when DRONE is combined with distillation, it reaches 12.3\hl{$\times$} speedup on natural language inference tasks of MRPC, Corpus of Linguistic Acceptability (CoLA)~\cite{Warstadt2019}, and Semantic Textual Similarity (STS) \cite{STS}.

 Overall, weight reduction directly reduces the \textsf{ML} model size by removing uncritical parameters. When performing tasks after weight reduction, \textsf{ML} models use less memory and require fewer arithmetic operations, which directly reduce the task latency with less workload and improve the computational resource efficiency. {This is critical for time-sensitive applications to improve the perceived availability and responsiveness of the system.} In addition, such improvements contribute to optimized energy consumption and cost. {Similar to quantization, the~weight-reduction techniques potentially make the model less resilient to certain types of hardware faults and such decrease the reliability. For~example, a~fault that affects a critical weight in a pruned network might have a bigger impact on the output than the same fault in an unpruned network, simply because there are fewer weights to 'absorb' the fault. }

\subsubsection{Knowledge~Distillation}

%\textbf{Theory.} Knowledge distillation. 
Knowledge Distillation is a procedure where a neural network is trained on the output of another network along with the original targets in order to transfer knowledge between the \textsf{ML} model architectures~\cite{Borup2021}. In~this process,  a~large and complex network, or~an ensemble model, is trained  with a labelled dataset for a better task performance. Afterwards, a~smaller network is trained with the help of the cumbersome model via a loss function $L$, measuring the output difference of the two models. This small network should be able to produce comparable results, and~in the case of over-fitting, it can even be made capable of replicating the results of the cumbersome~network. 

% Regarding the Knowledge distillation, 
 A knowledge-distillation framework  for a fast objects detection task is proposed in~\cite{Chen2017}. To~address the specific challenges of object detection in the form of regression, region proposals, and~less-voluminous labels, two aspects are considered: (i) a weighted cross-entropy loss, to~address the class imbalance, and~(ii) a teacher-bounded loss, to~handle the regression component and adaptation layers to better learn from intermediate teacher distributions. Evaluations with the datasets of Pattern Analysis, Statistical Modelling and Computational Learning (PASCAL)~\cite{Everingham2010}, Karlsruhe Institute of Technology and Toyota Technological Institute (KITTI)~\cite{Geiger2013}, and~COCO showed accuracy improvements by 3\% to 5\%. 
Wen~et~al.~\cite{Wen2021} argued that overly uncertain supervision of teachers can negatively influence the model's results. This is due to the fact that the knowledge from a teacher is useful but still not exactly right compared with a ground truth. Knowledge adjustment and dynamic temperature distillation are introduced in this work to penalize incorrect supervision and overly uncertain predictions from the teacher, making student models more discriminatory. Experiments on CIFAR-100~\cite{Krizhevsky2009}, CINIC-10~\cite{Darlow2018}, and~Tiny ImageNet~\cite{Le2015} showed nearly state-of-the-art method~accuracy.

MiniVit~\cite{Zhang2022a} proposes to compress vision transformers with weight sharing across layers and weight distillation. A~linear transformation is added on each layer’s shared weights to increase weight diversity. Three types of distillation for transformer blocks are considered in this work: (i) prediction-logit distillation, (ii) self-attention distillation, and~(iii) hidden-state distillation. Experiments showed MiniViT can reduce the size of the pre-trained Swin-B transformer by 48\% while achieving an increase of 1.0\% in Top-1 accuracy on~ImageNet.

Overall, knowledge distillation directly reduces the \textsf{ML} model size by  simplifying model structures. Compared to the source model, the~target model has a more compact and distilled structure with less parameters. Hence the workload of a task is reduced, leading to a better computational efficiency, higher availability, low task latency, and~optimized energy consumption and cost. {The distilled models potentially have higher reliability because they exert less stress on the hardware, reducing the likelihood of hardware faults or overheating. However, depending on the specific setup, the~faults could have a greater impact on the model applications.}

\subsubsection{Activation~Approximation}

%\textbf{Theory.} 
Besides the neural network’s size complexity, i.e.,~in terms of the number of parameters,  and~architecture complexity, i.e.,~in the terms of layers, activation functions also impact the task latency of a neural network. Activation functions approximation replaces non-linear activation functions (e.g., \textit{\hl{sigmoid} %Is the italic necessary? -> yes for mathematic function names.
} and \textit{\hl{tanh}}) in \textsf{ML} models with less computationally expensive functions (e.g., \textit{\hl{ReLU}}) to simplify the calculation or convert the computationally expensive calculation to series of lookup tables.
 In an early work~\cite{Amin1997}, the~Piece-wise Linear Approximation of Non-linear Functions (PLAN) was studied. The~\textit{\hl{sigmoid}} function was approximated by a combination of straight lines, and~the gradient of the lines were chosen such that all the multiplications were replaced by simple shift operations. Compared to \textit{\hl{sigmoid}} and \textit{\hl{tanh}}, Hu~et~al.~\cite{Hu2021} show that \textit{ReLU}, among~other linear functions, is not only less computationally expensive but also proved to be more robust to handle the neural network vanishing gradient problem, in~which the error dramatically decreases along with the back-propagation process in deep neural~networks. 

 Activation approximation improves the computing resource usage by reducing the required number of arithmetic operations in \textsf{ML} models, and~thus decreases the task latency with an acceptable increase in task~error. 

\subsection{Distributed~Inference}	

%\textbf{Theory.} 
Distributed Inference divides \textsf{ML} models into different partitions and carries out a collaborative inference by allocating partitions to be distributed over edge  resources and computing in a distributed manner~\cite{Zhao2018a}. 

The target edge  resources to distribute the inference task can be broadly divided into three levels: (i) local processors in the same edge  device~\cite{Lane2016}, (ii) interconnected edge  devices~\cite{Zhao2018a}, and~(iii) edge  devices and cloud servers~\cite{Li2018d}. Among~the three levels, an~important research challenge is to identify the partition points of \textsf{ML} models by measuring data exchanges between layers to balance the usage of local computational resources and bandwidth among distributed~resources. 

To tackle the tightly coupled structure of CNN, a~model parallelism optimization is proposed in~\cite{Du2021}, where the objective is to distribute the inference on edge  devices via a decoupled CNN structure. The~partitions are optimized based on channel group to partition the convolutional layers and then an input-based method to partition the fully connected layers, further exposing the high degree of parallelism. %Check meaning retained
Experiments show that the decoupled structure can accelerate the inference of large-scale ResNet-50 by 3.21\hl{$\times$} and reduce 65.3\% memory use with 1.29\% accuracy improvement. Another distributed inference framework is also proposed in~\cite{Hemmat2022} to decompose a complex neural network into small neural networks and apply class-aware pruning on each small neural network on the edge  device. The~inference is performed in parallel while considering the available resources on each device. The~evaluation shows that the framework achieves up to 17\hl{$\times$} speed-up when distributing a variant of VGG-16 over 20 edge  devices, with~around a 0.5\% loss in~accuracy.

Distributed inference can improve the end-to-end task latency by increasing the computing parallelism over a distributed architecture. At~a price of bandwidth usage and network dependency, the~overall energy efficiency and cost are optimized. By~distributing the inference task, the~load on individual devices is reduced, allowing more tasks to be processed concurrently, which can increase the availability of the ML application. In~a distributed configuration, if~one node fails, the~task can be reassigned to another node, thereby increasing the overall system~reliability.

\subsection{Other Inference Acceleration~Techniques}	
There exist other ways for accelerating inference in the literature. These have been categorized in a separate category as they are not as popular as the previously discussed techniques. These include: (i) Early Exit of Inference (EEoI),  (ii) Inference Cache, and (iii)~Model-Specific Inference Acceleration. We briefly review them in the~following. 

\subsubsection{Early Exit of Inference (\highlighting{EEoI}%We removed the italics. Please confirm this revision -> ok
)}

%\textbf{Theory.} 
The Early Exit of Inference (EEoI) is powered by a deep network architecture augmented with additional side branch classifiers~\cite{Teerapittayanon2016}. This allows prediction results for a large portion of test samples to exit the network early via these branches when samples can already be inferred with high confidence.

BranchyNet, proposed in~\cite{Teerapittayanon2016}, is based on the observation that features learned at an early layer of a network may often be sufficient for the classification of many data points. By~adding branch structures and exit criteria to neural networks, \hl{BranchyNet} is trained by solving a joint optimization problem on the weighted sum of the loss functions associated with the exit points. During~the inference, BranchyNet uses the entropy of a classification result as a measure of confidence in the prediction at each exit point and allows the input sample to exit early if the model is confident in the prediction. Evaluations have been conducted with LeNet~\cite{LeCun1998c}, AlexNet, and ResNet on MNIST, CIFAR-10 datasets, showing \textit{BranchyNet} can improve accuracy and significantly reduce the inference time of the network by~2\hl{$\times$}--6\hl{$\times$}. 

To improve the modularity of the \textit{EEoI} methods, a~plug-and-play technique named Patience-based Early Exit is proposed in~\cite{Zhou2020} for single-branch models (e.g., ResNet, Transformer). %Check meaning retained
The~work couples an internal classifier with each layer of a pre-trained language model and dynamically stops inference when the intermediate predictions of the internal classifiers remain unchanged for a pre-defined number of steps. Experimental results with the ALBERT model~\cite{Lan2019} show that the technique can reduce the task latency by up to 2.42\hl{$\times$} and slightly improve the model accuracy by preventing it from overthinking and exploiting multiple classifiers for~prediction.

EEoI can statistically improve the latency of inference tasks by reducing the inference workload at the price of a decrease in the accuracy. {By increasing throughput, the~technique leads to better availability}. The~side branch classifiers slightly increase the memory use during inference, while the task computational efficiency is higher as in most of cases where side branch classifiers can stop the inference earlier. {In scenarios where a high level of certainty is needed, an~early exit might introduce a higher probability of error, potentially compromising the reliability of the system.} Generally, a~correctly designed and trained \hl{EEoI} %Is the italics necessary? -> removed
technique is able to improve energy efficiency and optimize~cost. 

\subsubsection{Inference~Cache}

%\textbf{Theory.} 
Inference Cache saves models or models’ inference results to facilitate future inferences of similar interest. This is motivated by the fact that \textsf{ML} tasks requested by nearby users within the coverage of an edge  node may exhibit spatio-temporal locality~\cite{Drolia2017}. For~example, users within the same area might request recognition tasks for the same object of interest, which introduces redundant computation of deep learning~inference.

Besides the \textit{Cachier}~\cite{Drolia2017}, which caches \textsf{ML} models with edge  server for recognition applications and shows 3\hl{$\times$} speedup in task latency, \textit{DeepCache}~\cite{Xu2018c} targets the cache challenge for a continuous vision task. Given input video streams, \textit{DeepCache} firstly discovers the similarity between consecutive frames and identifies reusable image regions. During~inference, \textit{DeepCache} maps the matched reusable regions on feature maps and fills the reusable regions with cached feature map values instead of real Convolutional Neural Network (CNN) execution. Experiments show that \textit{DeepCache} saves up to 47\% inference execution time and reduces system energy consumption by 20\% on average. 
A hybrid approach, semantic memory design (\textsf{SMTM}), is proposed in~\cite{Li2021b}, combining inference cache with \textsf{EEoI}. In~this work, low-dimensional caches are compressed with an encoder from high-dimensional feature maps of hot-spot classes. During~the inference, \textsf{SMTM} extracts the intermediate features per layer and matches them with the cached features in fast memory: once matched, \textsf{SMTM} skips the rest of the layers and directly outputs the results. Experiments with AlexNet, GoogLeNet~\cite{Szegedy2015}, ResNet50, and MobileNet V2~\cite{Sandler2018} show that \textsf{SMTM} can speed up the model inference over standard approaches with  up to 2\hl{$\times$} and prior cache designs with up to 1.5\hl{$\times$} with only 1\% to 3\% accuracy~loss.%check meaning retained

Inference cache methods show their advantages of reducing task latency on continuous inference tasks or task batches. %Check meaning retained
Since the prediction is usually made together with current input and previous caches, the~accuracy can drop slightly. On~the computational efficiency front, the~cache lookup increases the computing workload and memory usage, while the global computational efficiency is improved across tasks, as~the inference computation for each data sample does not start from scratch. Energy consumption and cost are reduced in the context of tasks sharing spatio-temporal~similarity. 

\subsubsection{Model-Specific Inference~Acceleration}

%\textbf{Theory.}  
Besides the above-mentioned edge  inference techniques that can, in~theory, be applied to most of \textsf{ML} model structures, other research efforts aim at accelerating the inference process for specific model structures. 
We briefly review the representative methods of inference acceleration for three mainstream neural network structures: (i) CNN, (ii) Recurrent Neural Network (RNN), and~(iii) Transformers.

For CNN models, \textit{MobileNets}~\cite{Howard2017} constructs small and low-latency models based on depth-wise separable convolution. This factorizes a standard convolution into a depth-wise convolution and a $1\times1$ convolution, as~a  trade off between latency and accuracy during inference. The~latest version of \textit{MobileNets} V3~\cite{Howard2019} adds squeeze and excitation layers~\cite{Hu2020} to the expansion-filtering-compression block in \textit{MobileNets} V2~\cite{Sandler2018}. As~a result, it gives unequal weights to different channels from the input when creating the output feature maps. 
Combined with a later neural architecture search and NetAdapt~\cite{Yang2018}, \textit{MobileNets} V3-Large reaches 75.2\% accuracy and 156ms inference latency on ImageNet classification with a single-threaded core on a Google Pixel 1 phone. %Check meaning retained
\textit{GhostNet}~\cite{Han2020} also uses a depth-wise convolution to reduce the required high parameters and FLOPs induced by normal convolution: given an input image, instead of applying the filters on all the channels to generate one channel of the output, the~input tensors are sliced into individual channels and the convolution is then applied only on one slice. During~inference, $x$\% of the input is processed by standard convolution and the output of this is then passed to the second depth-wise convolution to generate the final output. Experiments demonstrate that \textit{GhostNet} can achieve higher recognition performance, i.e.,~75.7\% better top-1 accuracy than \textit{MobileNets} V3 with similar computational cost on the ImageNet dataset. %Check meaning retained
However, follow-up evaluations show that depth-wise convolution is more suitable for ARM/CPU and not friendly for GPU; thus, it does not provide a significant inference speedup in~practice. 

A real-time RNN acceleration framework is introduced in~\cite{Dong2020} to accelerate RNN inference for automatic speech recognition. The~framework consists of a block-based structured pruning and several specific compiler optimization techniques including matrix reorder, load-redundant elimination, and~a compact data format for pruned model storage. Experiments achieve real-time RNN inference with a Gated Recurrent Unit (GRU) model on an Adreno 640-embedded GPU and show no accuracy degradation when the compression rate is not higher than~10\hl{$\times$}. %Check meaning retained

Motivated by the way  we pay visual attention to different regions of an image or correlate words in one sentence, a~transformer is proposed in~\cite{Vaswani2017} showing encouraging results in various machine learning domains~\cite{Dosovitskiy2020, Wang2022}. On~the downside, transformer models are usually slower than competitive CNN models~\cite{Wang2022a} in terms of task latency due to the massive number of parameters, quadratic-increasing computation complexity with respect to token length, non-foldable normalization layers, and~lack of compiler-level optimizations. Current research efforts, such as~\cite{Graham2021, Roh2021}, mainly focus on simplifying the transformer architecture to fundamentally improve inference latency, among~which the recent \textsf{EfficientFormer}~\cite{Li2022} achieves 79.2\% top-1 accuracy on ImageNet-1K with only 1.6ms inference latency on an iPhone 12. In~this work, a~latency analysis is conducted to identify the inference bottleneck on different layers of vision transformer, and~the \textsf{EfficientFormer} relies on a dimension-consistent structure design paradigm that leverages hardware-friendly 4D MetaBlocks and powerful 3D multi-scale hierarchical framework blocks along with a latency-driven slimming method to deliver real-time inference at MobileNet~speed. 
 
Generally, model-specific inference acceleration techniques lower the workload of an inference task and thus reduce the task latency within the same edge environment, {resulting in higher availability.} Though computational resources usage can vary among techniques, most work reports an acceptable accuracy loss in exchange for a considerable decrease in resources usage. In~the case of model over-fitting, inference acceleration can improve the accelerated model accuracy. The~total energy consumption and cost are therefore reduced. {Under the assumption that the cached data are properly managed, the~ML system provides consistent responses to the same input, which can enhance the reliability of the system.}

\section{Edge~Learning}
\label{sec:edgelearning}

Edge learning techniques directly build \textit{ML} models on native edge devices with local data. Distributed learning, transfer learning, meta-learning, self-supervised learning, and other learning paradigms fitting into \textsf{Edge ML} are reviewed in this section to tackle different aspects of \textsf{Edge ML} requirements.

\subsection{Distributed~Learning}	
Compared to cloud-based learning in which raw or pre-processed data are transmitted to cloud for model training, distributed learning (\textsf{DL}) in the edge divides the model training workload onto the edge nodes, i.e.,~edge servers and/or edge clients, to~jointly train models with a cloud server by taking advantage of individual edge computational resources. 
Modern distributed learning approaches tend to only transmit locally updated model parameters or locally calculated outputs to the aggregation servers, i.e.,~cloud or edge, or~the next edge node: 
in the server--client configuration, the~aggregation server constructs the global model with all shared local updates~\cite{AbhishekVA2022}. On~the other hand, in~the peer-to-peer distributed learning setup, the~model construction is achieved in an incremental manner along with the participating edge nodes together~\cite{Wink2021}.

Distributed learning can be applied to all three basic \textsf{ML} paradigms, namely, supervised learning, unsupervised learning, and~reinforcement learning. Instead of learning from one optimization procedure $g_\omega$, distributed learning constructs a global model by aggregating the optimization results of all participant nodes, as~formalized by Equation~(\ref{fig:distributed_learning}):
\begin{equation}
	\label{fig:distributed_learning}
	\theta \approx \bigsqcup_{i=1}^n g_{\omega^i}(\mathbb{D}^i, L_{\mathbb{D}^i})
\end{equation}

\noindent where $g_{\omega^i}$ is the optimization procedure driven by the meta-knowledge ${\omega^i}$ of the participant node $i, i\!\in\!n$, and~$n$ is the number of distributed learning nodes. $\mathbb{D}$ stands for the data used for learning, which can be for example the labelled dataset $D$ for supervised learning, the~unlabelled dataset $\bar{D}$ for unsupervised learning, or~the MDP $M$ for reinforcement learning. $L_\mathbb{D}^i$ is the corresponding loss on the given data $\mathbb{D}$ and $\bigsqcup$ is the aggregation algorithm (e.g., \textsf{FedAvg} ~\cite{BrendanMcMahan2017} in the case of Federated Learning) to update the model by the use of all participants' optimization results (e.g., model parameters, gradients, outputs,~etc.).

The edge  distributed learning results into two major advantages: 

\begin{itemize}
	\item \textbf{\hl{Enhanced privacy and security}:} Edge data often contain sensitive information related to personal or organizational matters that the data owners are reluctant to share. By~transmitting only updated model parameters instead of the data, the~distributed learning on the edge trains \textsf{ML} models in a privacy-preserving manner. Moreover, the~reduced frequency of data transmission enhances the data security by restraining sensitive data only to the edge~environment.

	\item \textbf{\hl{Communication and bandwidth optimization}:} Uploading data to the cloud leads to a large transmission overhead and is the bottleneck of current learning paradigm~\cite{Kang2017}. A significant amount of communication is reduced by processing data in the edge nodes, and~bandwidth usage optimized via edge distributed learning.
\end{itemize}

From the architectural perspective, there are three main organizational architectures~\cite{Abreha2022, Wang2020a} that exist to achieve distributed learning in the server--client configuration, as~illustrated in Figure~\ref{fig:DistributedArch} and introduced as follows: 

\begin{figure}[H]
	\centering
	\includegraphics[width=\textwidth]{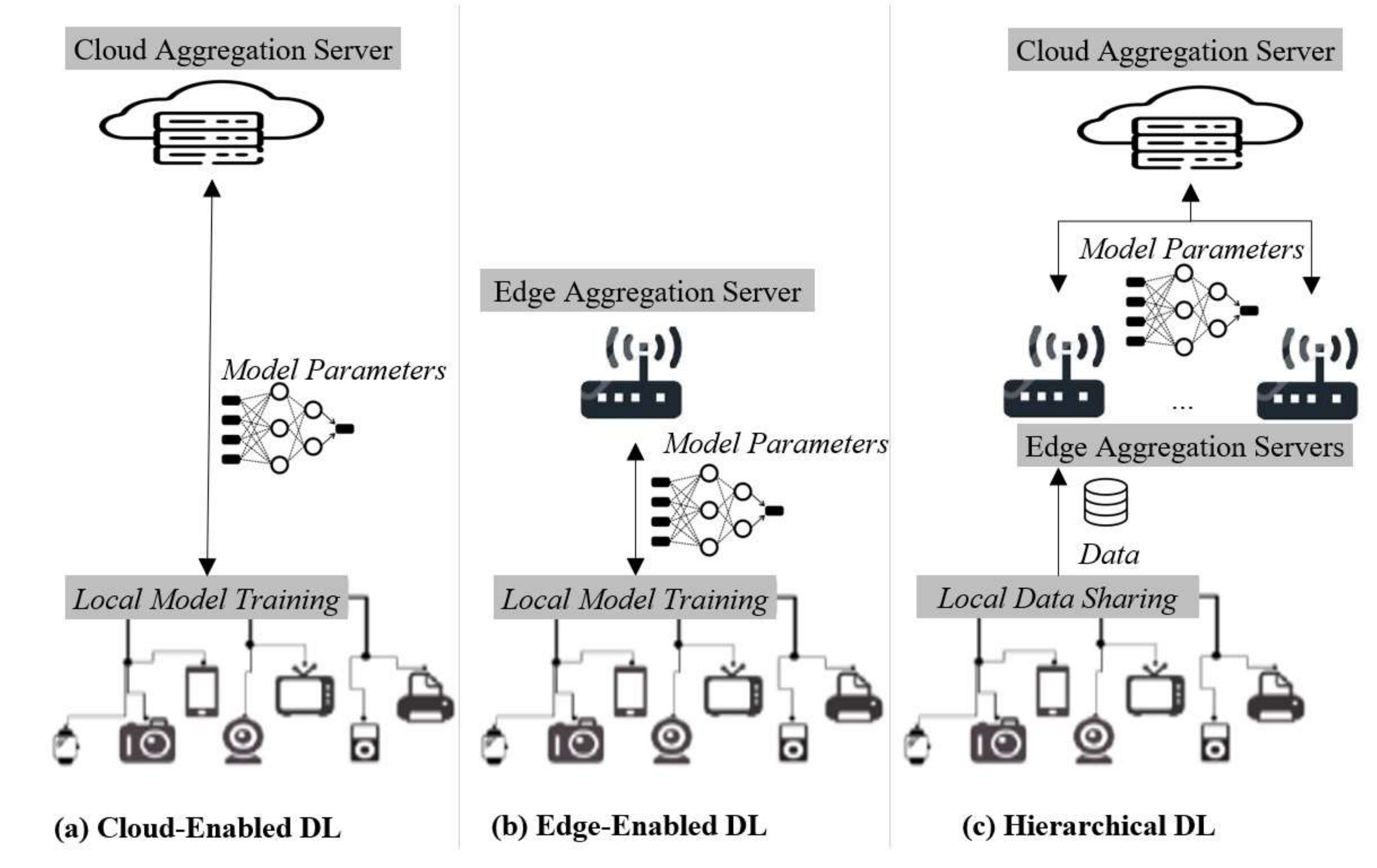}
	\caption{\hl{The} %MDPI: Please add the left bracket in the image, e.g., “A)” should be “(A)”. -> updated. 
 distributed learning architectures available in the~literature.}
	\label{fig:DistributedArch}
\end{figure}

\begin{itemize}
	\item \textbf{\hl{Cloud-enabled \textsf{DL}.}} Given a number of distributed and interconnected edge nodes, cloud-enabled \textsf{DL} (see Figure~\ref{fig:DistributedArch}a) constructs the global model by aggregating in the cloud the local models’ parameters. These parameters are computed directly in each edge device. Periodically, the~cloud server shares the global model parameters to all edge nodes so that the upcoming local model updates are made on the latest global~model. 
	
	\item \textbf{\hl{Edge-enabled \textsf{DL}.}} In contrast to cloud-enabled \textsf{DL}, Edge-enabled \textsf{DL} (see Figure~\ref{fig:DistributedArch}b) uses a local and edge server to aggregate model updates from its managed edge devices. %check meaning retained
	Edge devices, with~the management range of an edge server, contribute to the global model training on the edge aggregation server. Since the edge aggregation server is located near the edge devices, edge-enabled \textsf{DL} does not necessitate communications between the edge and the cloud, which thus reduces the communication latency and brings task offline capability. On~the other hand, edge-enabled \textsf{DL} is often resource-constrained and can only support a limited number of clients. This usually results in a degradation in  task performance over~time.%Check meaning retained
	
	\item \textbf{\hl{Hierarchical \textsf{DL}.}} Hierarchical \textsf{DL} employs both cloud and edge aggregation servers to build the global model. Generally, edge devices within the range of a same-edge server transmit local data to the corresponding edge aggregation server to individually train local models, and~then local models’ parameters are shared with the cloud aggregation server to construct the global model. Periodically, the~cloud server shares the global model parameters to all edge nodes (i.e., servers and devices), so that the upcoming local model updates are made on the latest global model. By~this means, several challenges of distributed learning, such as Non-Identically Distributed Data (Non-IID)~\cite{Zhu2021}, imbalanced class~\cite{Wang2021b}, and the~heterogeneity of edge devices~\cite{Xu2021a} with diverse computation capabilities and network environments, can be targeted in the learning design. In~fact,  as~each edge aggregation server is only responsible for training the local model with the collected data, the~cloud aggregation server does not need to deal with data diversity and device heterogeneity across the edge nodes. 
\end{itemize}

In the following, we review two distributed learning paradigms in the context of \textsf{Edge ML}:  (i) federated learning, and~(ii) split~learning.

\subsubsection{Federated~Learning}
%\textbf{Theory.} 
Federated Learning (FL)~\cite{AbhishekVA2022} enables edge nodes to collaboratively learn a shared model while keeping all the training data on edge nodes, decoupling the ability to conduct machine learning from the need to store the data in the cloud. In~each communication round, the~aggregation server distributes the global model’s parameters to edge training nodes, and~each node trains its local model instance with newly received parameters and local data. The~updated model parameters are then transmitted to the aggregation server to update the global model. 
The aggregation is commonly realized via federated average (\textsf{FedAvg})~\cite{BrendanMcMahan2017} or Quantized Stochastic Gradient Descent (\textsf{QSGD})~\cite{Alistarh2017} for neural networks, involving multiple local Stochastic Gradient Descent (\textsf{SGD}) updates and one aggregation by the server in each communication~round.

\textsf{FL} is being widely studied in the literature. In~particular, the~survey in~\cite{Abreha2022} summarizes and compares more than forty existing surveys on FL and edge computing regarding the covered topics. According to the distribution of training data and features among edge nodes, federated learning can be divided into three categories~\cite{Yang2019}: (i) Horizontal Federated Learning (\textsf{HFL}), (ii) Vertical Federated Learning (\textsf{VFL}), and~(iii) Federated Transfer Learning (\textsf{FTL}). \textsf{HFL} refers to the federating learning paradigm where training data across edge nodes share the feature space but have different ones in samples. %Check meaning retained
\textsf{VFL} federates models trained from data sharing the sample IDs but different feature space across edge nodes. Finally, \textsf{FTL} refers to the paradigm where data across edge nodes are correlated but differ in both samples and feature~space. 

\textsf{HFL} is widely used to handle homogeneous feature spaces across distributed data. In~addition to the initial work of \textsf{FL}~\cite{AbhishekVA2022}, showing considerable latency and throughput when performing the query suggestion task in mobile environments. \textsf{HFL} is highly popular in the healthcare domain~\cite{Xu2021c} where it is, for~instance, used to learn from different electronic health records across medical organizations without violating patients’ privacy and improve the effectiveness of data-hungry analytical approaches. 
To tackle the limitation that \textsf{HFL} does not handle heterogeneous feature spaces, the~continual horizontal federated learning (\textsf{CHFL}) approach~\cite{Mori2022} splits models into two columns corresponding to common features and unique features, respectively, and~jointly trains the first column by using common features through \textsf{HFL} and locally trains the second column by using unique features. Evaluations demonstrate that \textsf{CHFL} can handle uncommon features across edge nodes and outperform the \textsf{HFL} models with are only based on common~features. 

As a more challenging subject than \textsf{HFL}, \textsf{VFL} is studied in~\cite{Nock2018} to answer the entity resolution question, which aims at finding the correspondence between samples of the datasets and learning from the union of all features. Since loss functions are normally not separable over features, a~token-based greedy entity-resolution algorithm is proposed in~\cite{Nock2018} to integrate the constraint of carrying out entity resolution within classes on a logistic regression model. Furthermore, most studies of \textsf{VFL} only support two participants and focus on binary class logistic regression problems.  A~Multi-participant Multi-class Vertical Federated Learning (\textsf{MMVFL}) framework is proposed in~\cite{Feng2020}.  \textsf{MMVFL} enables label sharing from its owner to other \textsf{VFL} participants in a privacy-preserving manner. Experiment results on two benchmark multi-view learning datasets, i.e.,~Handwritten and Caltech7~\cite{Li2015}, show that \textsf{MMVFL} can effectively share label information among multiple \textsf{VFL} participants and match multi-class classification performance of existing~approaches.

As an extension of the federated learning paradigm, \textsf{FTL} deals with the learning problem of correlated data from different sample spaces and feature spaces. FedHealth~\cite{Chen2020a} is a framework for wearable healthcare targeting the \textsf{FTL} as a union of \textsf{FL} and transfer learning. The~framework performs data aggregation through federated learning to preserve data privacy and builds relatively personalized models by transfer learning to provide adapted experiences in edge devices. To~address the data scarcity in \textsf{FL}, an~\textsf{FTL} framework for cross-domain prediction is presented in~\cite{Wang2022b}. The~idea of the framework is to share existing applications’ knowledge via a central server as a base model, and~new models can be constructed by converting a base model to their target-domain models with limited application-specific data using a transfer learning technique. Meanwhile, the~federated learning is implemented within a group to further enhance the accuracy of the application-specific model. The~simulation results on COCO and PETS2009~\cite{Ferryman2009} datasets show that the proposed method outperforms two state-of-the-art machine learning approaches by achieving better training efficiency and prediction~accuracy. 

%Furthermore, a
Besides the privacy-preserving nature of \textsf{FL}~\cite{Li2021a}, and~in addition to the research efforts on \textsf{HFL}, \textsf{VFL}, and~\textsf{FTL}, challenges have been raised in federated learning oriented to security~\cite{Lyu2020}, communication~\cite{Yang2022a}, and~limited computing resources~\cite{Nguyen2021}. This is important as edge devices usually have higher task and communication latency and are in vulnerable environments.
In fact, low-cost IoT and Cyber-Physical System (CPS) devices are generally vulnerable to attacks due to the lack of fortified system security mechanisms. Recent advances in cyber-security for federated learning~\cite{Ghimire2022} reviewed several security attacks targeting \textsf{FL} systems and the distributed security models to protect locally residual data and shared model parameters. With~respect to the parameter aggregation algorithm, the~commonly used \textsf{FedAvg} employs the aggregation server to centralize model parameters;~thus, attacking the central server breaks the \textsf{FL}’s security and privacy. Decentralized \textsf{FedAvg} with momentum (\textsf{DFedAvgM})~\cite{Sun2021} is presented on edge nodes that are connected by an undirected graph. In~\textsf{DFedAvgM}, all clients perform stochastic gradient descent with momentum and communicate with their neighbors only. The~convergence is proved under trivial assumptions, and~evaluations with ResNet-20 on CIFAR-10 dataset demonstrate no significant accuracy loss when local epoch is set to~1. 

From a communication perspective, although~\textsf{FL} evades transmitting training data over a network, the~communication latency and bandwidth usage for weights or gradients shared among edge nodes are inevitably introduced. The~trade-off between communication optimization and the aggregation convergence rate is studied in~\cite{Reisizadeh2019}. A~communication-efficient federated learning method with Periodic Averaging and Quantization (\textsf{FedPAQ}) is introduced. In~\textsf{FedPAQ}, models are updated locally at edge devices and only periodically averaged at the aggregation server. In~each communication round between edge training devices and the aggregation server, only a fraction of devices participate in the parameters aggregation. Finally, a~quantization method is applied to quantize local model parameters before sharing with the server. Experiments demonstrate a  communication--computation trade-off to improve communication bottleneck and \textsf{FL} scalability. Furthermore, knowledge distillation is used in communication-efficient federated learning technique \textsf{FedKD}~\cite{Wu2022}. In~\textsf{FedKD}, a~small mentee model and a large mentor model learn and distill knowledge from each other. It should be noted that only the mentee model is shared by different edge nodes and learns collaboratively to reduce the communication cost. In~such a configuration, different training nodes have different local mentor models, which can better adapt to the characteristics of local datasets to achieve personalized model learning. Experiments with datasets on personalized news recommendations, text detection, and~medical named entity recognition show that \textsf{FedKD} maximally can reduce 94.89\% of communication cost and achieve competitive results with centralized model~learning. 

Federated learning on resource-constrained devices limit both communication and learning efficiency. The~balance between convergence rate and allocated resources in \textsf{FL} is studied in~\cite{DInh2021a}, where an \textsf{FL} algorithm \textsf{FEDL} is introduced to treat the resource allocation as an optimization problem. In~\textsf{FEDL}, each node solves its local training approximately till a local accuracy level is achieved. The~optimization is based on the Pareto efficiency model~\cite{Stiglitz1982} to capture the trade-off between the wall-clock training time and edge nodes energy consumption. Experimental results show that \textsf{FEDL} outperforms the vanilla \textsf{FedAvg} algorithm in terms of convergence rate and test accuracy. Moreover, computing resources can be not only limited but also heterogeneous at edge devices. A~heterogeneity-aware federated learning method, \textsf{Helios}, is proposed in~\cite{Xu2021e} to tackle the computational straggler issue. This  implies that the edge devices with weak computational capacities among heterogeneous devices may significantly delay the synchronous parameter aggregation. \textsf{Helios} identifies each device’s training capability and defines the corresponding neural network model training volumes. For~straggling devices, a~soft-training method is proposed to dynamically compress the original identical training model into the expected volume through a rotating neuron training approach. Thus, the~stragglers can be accelerated while retaining the convergence for local training as well as federated collaboration. Experiments show that \textsf{Helios} can provide up to $2.5\times$ training acceleration and maximum 4.64\% convergence accuracy improvement in various collaboration~settings. 

Table~\ref{tab:t1} summarizes the reviewed works related to FL topics and challenges. Besides~the efforts for security, communication, and resources, a~personalized federated learning paradigm is proposed in~\cite{Hahn2022}, so that each client has their own personalized model as a result of federated learning.  As~the existence of a connected subspace containing diverse low-loss solutions between two or more independent deep networks has been discovered, the~work combines this property with the model mixture-based personalized federated learning method for improved performance of personalization. Experiments on several benchmark datasets demonstrated that the method achieves consistent gains in both personalization performance and robustness to problematic scenarios possible in realistic~services.

\begin{table}[H]    
	\caption{\hl{FL} %MDPI: please check if the alignment keep and if vertical line can be removed? if this table no header? -> vertical lines are removed, bold style has been added to the headers, alignment keeps. 
 related~work.}
        \small %
	%\resizebox{\columnwidth}{!}{
        \begin{tabular}{p{4cm}p{4.5cm}p{4cm}}
		 \noalign{\hrule height 1pt}
		\multicolumn{3}{c}{\textbf{Data and Features for FL}} \\
		\hline
		HFL & VFL & FTL \\
		\hline
~\cite{AbhishekVA2022, BrendanMcMahan2017, Xu2021c, Mori2022} & \cite{Nock2018, Feng2020} & \cite{Chen2020a, Wang2022b}\\
		\hline
		\multicolumn{3}{c}{\textbf{Challenges}} \\
		\hline
		Enhanced Security & Efficient Communication & Optimized Resources \\
		\hline
~\cite{Lyu2020, Ghimire2022, Sun2021} & \cite{Yang2022a, Sun2021, Reisizadeh2019, Wu2022} & \cite{Nguyen2021, DInh2021a, Xu2021e} \\
	 \noalign{\hrule height 1pt}
	\end{tabular}
	%}
	\label{tab:t1}
\end{table}

Overall, \textsf{FL} is designed primarily to protect data privacy during model training. Sharing models and performing distributed training increases the computation parallelism and reduces the communication cost, and~thus reduces both the end-to-end training task latency and the communication latency. Moreover, specific \textsf{FL} design can provide enhanced security, optimized bandwidth usage and efficient computing resource usage. The~edge-enabled \textsf{FL} as an instance of the edge-enabled \textsf{DL} can further bring offline capability to \textsf{ML} models. {Generally, since local devices can continue training on the local data even if the network connection is down, which can improve the availability of the application during network failures or disruptions. Implementation requires the careful management and coordination of updates from multiple devices, handling devices with differing computational capabilities, and~dealing with potential delays in communication. These factors can impact the reliability of applications of Federated Learning.}

\subsubsection{Split~Learning}
%\textbf{Theory.} 
As another distributed collaborative training paradigm of \textsf{ML} models for data privacy, Split Learning (\textsf{SpL})~\cite{Gupta2018b} divides neural networks into multiple sections. Each section is trained on a different node, either a server or a client. During~the training phase, the~forward process firstly computes the input data within each section and transmits the outputs of the last layer of each section to the next section. Once the forward process reaches the last layer of the last section, a~loss is computed on the given input. The~backward propagation shares the gradients reversely within each section and from the first layer of the last section to the previous sections. During~the backward propagation, the~model parameters are updated in the meantime. The~data used during the training process are stored across servers or clients which take part in the collaborative training. However, none of the involved edge nodes can review data from other sections. The~neural network split into sections and trained via \textsf{SpL} is called Split Neural Network (\textsf{SNN}).

The \textsf{SpL} method proposed in~\cite{Gupta2018b} splits the training between high-performance servers and edge clients, and~orchestrates the training over sections into three steps: (i)~training request, (ii) tensor transmission, and~(iii) weights update. Evaluations with VGG and Resnet-50 models on MNIST, CIFAR-10, and ImageNet datasets show a significant reduction in the required computation operations and communication bandwidth by edge clients. This is because only the first few layers of \textsf{SNN} are computed on the client side, and~only the gradients of few layers are transmitted during backward propagation. When a large number of clients are involved, the~validation accuracy and convergence rate of \textsf{SpL} are higher than \textsf{FL}, as~general non-convex optimization averaging models in a parameter space could produce an arbitrarily bad model~\cite{Goodfellow2015}.

\begin{figure}[H]
	\centering
	\includegraphics[width=\textwidth]{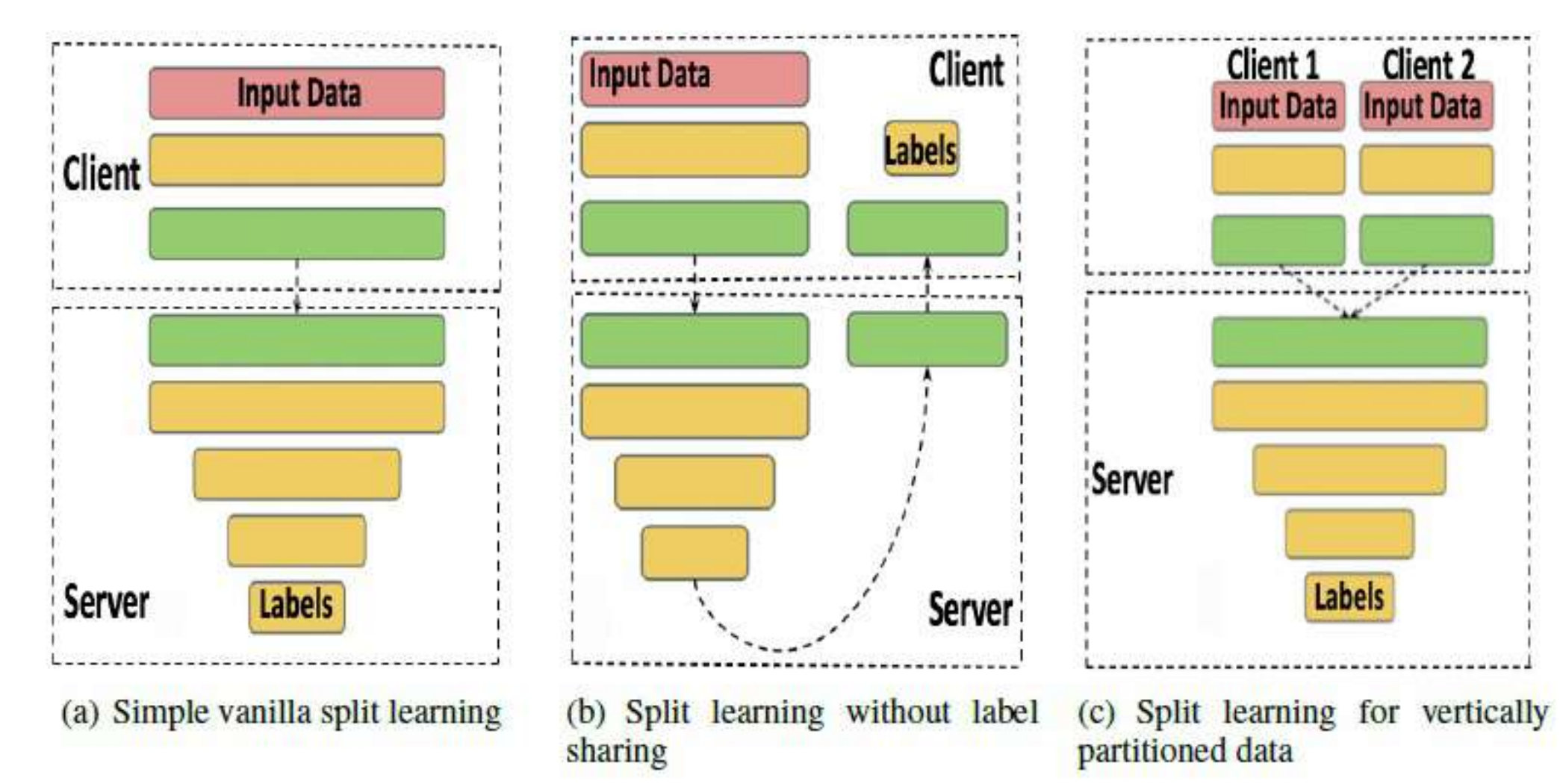}
	\caption{\hl{Split } %MDPI: please check if colors and arrows need explanation? -> not explained in the original paper [157]
Learning configurations~\cite{Vepakomma2018}.}
	\label{fig:3}
\end{figure}

The configuration choice to split a neural network across servers and clients are subject to design requirements and available computational resources. The~work in~\cite{Vepakomma2018} presents several configurations of \textsf{SNN} catering to different data modalities, of~which Figure~\ref{fig:3} illustrates three representative configurations: (i) in vanilla \textsf{SpL}, each client trains a partial deep network up to a specific layer known as the cut layer, and~the outputs at the cut layer are sent to a server which completes the rest of the training. During~the parameters update, the~gradients are back-propagated at the server from its last layer until the cut layer. The~rest of the back propagation is completed by the clients. (ii) In the configuration of \textsf{SpL} without label sharing, the~\textsf{SNN} is wrapped around at the end layers of the servers. The~outputs of the server layers are sent back to clients to obtain the gradients. During~backward propagation, the~gradients are sent from the clients to servers and then back again to clients to update the corresponding sections of the \textsf{SNN}. (iii) \textsf{SpL} for vertically partitioned data allows multiple clients holding different modalities of training data. In~this configuration, each client holding one data modality trains a partial model up to the cut layer, and~the cut layer from all the clients are then concatenated and sent to the server to train the rest of the model. This process is continued back and forth to complete the forward and backward propagation. Although~the configurations show some versatile applications for \textsf{SNN}, other configurations remain to be~explored. \vspace{-6pt}

Compared to \textsf{FL}, the~\textsf{SNN} makes \textsf{SpL} a better option for resource-constrained environments. On~the other hand, \textsf{SpL} performs slower than \textsf{FL} due to the relay-based training across multiple clients. 
To complement both learning paradigms, Split Federated Learning (\textsf{SFL})~\cite{Thapa2020} aims at bringing \textsf{FL} and \textsf{SpL} together for model privacy and robustness. \textsf{SFL} offers model privacy by network splitting and client-side model updates based on \textsf{SpL}, as~well as shorter training latency by performing parallel processing across clients. Experiments demonstrate that \textsf{SFL} provides similar test accuracy and communication efficiency to \textsf{SL}, while significantly decreasing its computation time per global epoch than in \textsf{SpL} for multiple~clients.

% \begin{table}[h]
% \centering
% \caption{Summary of Key Split Learning Works}
% \begin{tabular}{|p{2.5cm}|p{2.5cm}|p{2cm}|p{4cm}|p{2.5cm}|}
% \hline
% \textbf{Reference} & \textbf{Approach} & \textbf{Models} & \textbf{Datasets} & \textbf{Major Findings} \\
% \hline
% Gupta~et~al.~\cite{Gupta2018b} & Split Learning & VGG, Resnet-50 & MNIST, CIFAR-10, ImageNet & Significant reduction in required computation operations and communication bandwidth. Higher validation accuracy and convergence rate than FL. \\
% \hline
% Vepakomma~et~al.~\cite{Vepakomma2018} & Different SNN configurations & Various & Various & Demonstrated versatility of SNNs for different data modalities. More configurations to be explored. \\
% \hline
% Thapa~et~al.~\cite{Thapa2020} & Split Federated Learning & Various & Various & Offers model privacy and robustness. Similar test accuracy and communication efficiency as SL, with significant decrease in computation time per global epoch. \\
% \hline
% \end{tabular}
% \end{table}

Overall, \textsf{SpL} largely improves training task latency by taking advantage of both server-side and edge-side computational resources. Compared to \textsf{FL} where all model gradients or weights are transmitted over a network, \textsf{SpL} only shares gradients of few layers of \textsf{SNN} and thus further optimizes the bandwidth usage. {By reducing the amount of the transmitted data, split learning can help improve the availability of the application, especially in bandwidth-constrained environments.} The \textsf{SNN} model performance is better compared to \textsf{FL} by avoiding \textsf{FedAvg} or \textsf{QSGD} during training. In~addition to data privacy, which is enhanced by all distributed learning paradigms, \textsf{SpL} is excellent at preserving model privacy, as both the data and model structure are opaque across sections. Energy consumption and cost are thus reduced as a result of these \textsf{SpL} advantages. {However, the~implementation of Split Learning relies on proper partitioning handling, device heterogeneity management, and~communication synchronisation, which can impact the reliability of the SpL~applications.}

\subsection{Transfer~Learning}	

%\textbf{Theory.} 
Transfer Learning (\textsf{TL}) is inspired by humans’ ability to transfer knowledge across domains. Instead of training models from scratch, \textsf{TL} aims at creating high-performance models on a target domain by transferring the knowledge from models of a different but correlated source domain~\cite{Panigrahi2021}. The~knowledge transfer in the context of transfer learning can be in the following three levels according to the discrepancy between domains:  

\begin{itemize}
	\item \textbf{\hl{Data Distribution.}} The training data obtained in a specific spatial or temporal point can have different distribution to the testing data in an edge  environment. The~different data distribution, due to different facts such as co-variate shift~\cite{Sugiyama2008}, selection bias~\cite{Huang2006}, and~context feature bias~\cite{Singh2020}, could lead to the degradation of model performance in a testing environment. The~knowledge transfer between two different data distributions is a subtopic of transfer learning, known as Domain Adaptation (\textsf{DA})~\cite{Zhang2019}. %check meaning retained
	
	\item \textbf{\hl{Feature Space.}} Contrary to the homogeneous transfer learning~\cite{Zhuang2021} which assumes that the source domain and the target domain consist of the same feature spaces, heterogeneous transfer learning tackles the (\textsf{TL}) case where the source and target domains have different feature spaces~\cite{Pan2010}. The~heterogeneous transfer learning applies a feature space adaptation process to ease the difficulty to collect data within a target domain and expands the transfer learning to broader~applications. 
	
	\item \textbf{\hl{Learning Task Space.}} %As the commonest application, 
	Transfer learning also transfers knowledge between two specific learning tasks by use of the inductive biases of the source task to help perform the target task~\cite{Weiss2016}. In~this level, the~data of the source and target task can have a same or different distribution and feature space. However, the~specific source and target tasks are supposed to be similarly correlated either in a parallel manner, e.g.,~in the tasks of objects identification and person identification, or~in a downstream manner, e.g.,~from a pretext learning task of image representation to a downstream task of an object detection task. It is worth mentioning that the knowledge generalization in an upstream manner from downstream tasks to out-of-distribution data is Domain Generalization (\textsf{DG})~\cite{Zhou2021}.  
\end{itemize}

As a learning paradigm focusing on the techniques to transfer knowledge between domains,~transfer learning can be applied into all three basic learning categories, i.e.,~supervised learning, unsupervised learning, and~reinforcement learning, for~knowledge transfer between domains~\cite{Weiss2016}. 
Based on the knowledge transfer process, two transfer learning techniques exist to build neural networks for the target domain: (i) Layer Freezing and~(ii) Model Tuning. Layer Freezing is generally applied to transfer knowledge between domains that are correlated in a parallel manner and/or in~situations where a target domain requests low training latency and has few training data. The~process is summarized as~follows.
\newpage

\begin{enumerate}
	\item \hl{Model Collection: %Is the italics necessary? same as follows -> removed
} An existing trained model on the source domain is~acquired.
	
	\item \hl{Layer Freezing:} The first several layers from a source model are frozen to keep the previously learned representation, and~the exact layers to freeze are determined by the source model layers which have learned the source data representation~\cite{Li2018e}, i.e.,~usually the data-encoding part of a~model.
	
	\item \hl{Model Adjustment:} The last few layers of the source model are deleted, and~again the exact layers to delete are determined by the source model structure~\cite{Zhi2017}. New trainable layers are added after the last layer of the modified source model to learn to turn the previous learned representation into outputs on the target~domain.
	
	\item \hl{Model Training}: The updated model is trained with new data from the target domain. 
	\item \hl{Model Tuning:} At last, an~optional step is the tuning process usually based on model fine-tuning~\cite{Chu2016}. During~this step, the~entire newly trained model from the previous step is unfrozen and re-trained on the new data from the target domain with a low learning rate. The~tuning process potentially further improves the model performance by adapting the newly trained representation to the new data.
\end{enumerate}

On the other hand, Model Tuning is generally applied to transfer knowledge among domains that are correlated in a downstream manner and/or in~situations where a target domain has sufficient training data. The~process of tuning based transfer learning can be summarized as~follows.

\begin{enumerate}
	\item \hl{Model Pre-training:} A model is pre-trained on the source domain to learn representations from source domain~data.
	
	\item \hl{Model Adjustment:} As an optional step in tuning process, the~last few layers of the source model are deleted, and~new trainable layers are added after the last layer of the modified source~model.
	
	\item \hl{Model Tuning:} The entire pre-trained model is trained on the new data from the target domain to map the learned representation to the target output. 
\end{enumerate}

During the two transfer learning processes, the~parameters of the original model $\theta$ are updated to the new model parameters $\theta'$ with the dataset $\mathbb{D}'$ from the target domain through an optimization procedure $g_\omega'$:
\begin{equation}
	\label{distributed_learning}
	\theta' := g_{\omega'}(\mathbb{D}', L_\mathbb{D'})
\end{equation}

On the target domain, the~meta-knowledge $\omega'$ and the optimization procedure $g_{\omega'}$ can be derived from the source domain during the transfer process; however, the~focus of transfer learning is the knowledge transfer of model parameters from $\theta$ to $\theta'$.
Transfer learning building models based on previously learned knowledge in a correlated domain brings the following~benefits.  

\begin{itemize}
	\item \hl{Training Efficiency.} The speed of training new models is largely accelerated and uses much less computational resources compared to model training from~scratch. 
	
	\item \hl{Less Training Data.} The model training or tuning process on the target model requires less training data, and~this is especially useful in the case where there is a lot of data available from the source domain and relatively less data for target~domain. 
	
	\item \hl{Model Personalization.} Transfer learning can quickly specialize pre-trained models to a specific environment and improve accuracy when the original pre-trained model cannot generalize well. 
\end{itemize}

Transfer learning techniques are studied and compared in several surveys: an early study~\cite{Pan2010} associates the definition of transfer learning to the reasoning-based categories, and~divides transfer learning into: (i) inductive transfer learning, (ii) transductive learning, and~(iii) unsupervised learning, with respect to the source and target task spaces. 
To handle source and target feature space, homogeneous transfer learning is reviewed in~\cite{Zhuang2021, Weiss2016}, and~heterogeneous transfer learning is analyzed in~\cite{Pan2010, Weiss2016}. Regarding the domain adaptation for different data distributions, the~state-of-the-art methods are summarized based on training loss in~\cite{Wang2018b} for computer vision applications. In~particular, recent research efforts tend to extend the scope of vanilla Domain Adaptation (\textsf{DA}) for different data distributions to different feature spaces or task spaces. The~term ``deep domain adaptation'' is used in~\cite{Wang2018b} to designate the methods that leverage deep neural networks and \textsf{DA} to solve both distribution shift and feature space differences. 
A Universal Domain Adaptation (\textsf{UDA}) method is described in~\cite{You2019} as a more general approach of transfer learning across a task space. \textsf{UDA} targets the supervised model transfer between domains where source and target have overlapped but have different label spaces. Without~prior knowledge on the label sets from both domains, \textsf{UDA} is capable of classifying its samples correctly if it belongs to any class in the source label set or mark it as ``unknown'' otherwise. To~address the unknown label classification, a~Universal Adaptation Network (\textsf{UAN}) is introduced to quantify the transferability of each sample into a sample-level weighting mechanism based on both the domain similarity and the prediction uncertainty of each sample. Empirical results show that \textsf{UAN} works stably across different \textsf{UDA} settings and outperforms the state-of-the-art closed set, partial and open set domain adaptation~methods.

The related works of transfer learning are highlighted in Table \ref{tab: transfer learning} and are introduced in the following. 
%Concerning the transfer process, 
Regarding the layer freezing, one of the most popular application domains is healthcare, as~the training data related to specific diseases can be difficult to obtain due to their rarity and the issue of privacy. %Check meaning retained
Transfer Learning is applied in~\cite{Rios-Urrego2020} to detect Parkinson's disease from speech symptom with layer freezing. In~this work, the~classification of patients with Parkinson's disease is realized with a \textsf{CNN} to analyze Mel-scale spectrograms in three different languages, i.e.,~Spanish, German, and~Czech, via a transfer learning process. During~the knowledge transfer, several consecutive layers are frozen to identify the layer topology characterizing the disease and others in the language. The results indicate that the fine-tuning of the neural network does not provide good performance in all languages, while the fine-tuning of individual layers improves the accuracy by up to 7\%. Moreover, transfer learning among languages improves the accuracy up to 18\%  compared to a model training from scratch. Since fine-tuning and storing all the parameters is prohibitively costly and eventually becomes practically infeasible for pre-trained language models, a~parameter-efficient fine-tuning method is proposed in~\cite{Ding2023} to optimize a small portion of the model parameters while keeping the rest fixed, drastically cutting down computation and storage~costs.

\begin{table}[H]
\centering
\small
\caption{{\hl{Highlighted} %MDPI: please check if the alignment keep and if vertical line can be removed? -> alignment keeps, vertical lines are removed. 
 related work of transfer learning.}}
\begin{tabular}{{p{1.2cm}p{4.4cm}p{6.9cm}}}
 \noalign{\hrule height 1pt}
\textbf{Works} & \textbf{Methods} & \textbf{Insights} \\
\hline
\cite{Rios-Urrego2020} & Layer Freezing & Frozen of consecutive layers to identify characterizing
topology. \\
\hline
\cite{Ding2023} & Parameter-Efficient Fine-tuning & Optimization of a small portion of parameters. \\
\hline
\mbox{\cite{Kara2021, Howard2018}} & Large Model Tuning & Fine-tuning large pre-trained models. \\
\hline
\cite{Houlsby2019} & Adapter Module Based Tuning & Few trainable parameters added per task. \\
\hline
\cite{Lester2021} & Prompt Tuning & Competitive with scale, outperforms GPT-3's few-shots learning. \\
\hline
\cite{Kojima2022} & Prompt Tuning & Increased zero-shot accuracy with ``Let's think step by step''. \\
\hline
\cite{Chung2022} & Instruction Fine-tuning & Large margin outperformance with FLan-PaLM 540B. \\
\hline
\cite{Schratz2019a} & Hyper-parameter Tuning & Finding optimal model hyper-parameters. \\
\hline
\cite{Wang2019b} & Negative Transfer & Adversarial network to filter unrelated source data. \\
 \noalign{\hrule height 1pt}
\end{tabular}
\label{tab: transfer learning}
\end{table}

Concerning the model-tuning, fine-tuning large pre-trained models is an effective transfer mechanism in both \textsf{CV}~\cite{Kara2021} and \textsf{NLP}~\cite{Howard2018} domains. As~the general fine-tuning creates an entirely new model for each downstream task, the~method is not efficient when facing multiple downstream tasks. In~fact, it results in the reproduction of the same-sized model multiple times. An~adapter module-based tuning method is introduced in~\cite{Houlsby2019}, where adapter modules extend the pre-trained models by only adding a few trainable parameters per task. The~parameters of the original network remain fixed, yielding a high degree of parameter sharing. The~experiment transferring \textsf{BERT} transformer to 26 diverse text classification tasks attain near state-of-the-art performance: on \textsf{GLUE} benchmark, the~proposed method shows only 0.4\% degradation compared to fine-tuned results, while adding only 3.6\% parameters per task compared to the 100\% parameter retraining of fine-tuning. 
Moreover, prompt tuning~\cite{Lester2021} is a simple yet effective method to learn prompts to perform specific downstream tasks without modifying models, which is especially useful when handling large language models and vision--language models. The~study in~\cite{Lester2021} shows that prompt tuning becomes more competitive with scale: as models exceed billions of parameters, the~proposed method matches the strong performance of model fine-tuning, and~largely outperforms the few-shots learning of Generative Pre-trained Transformer 3 (\textsf{GPT-3})~\cite{Brown2020}. As~the prompt plays an important role in the model output, an~interesting discovery is made in~\cite{Kojima2022} to perform reasoning tasks with pre-trained Large Language Models (LLMs) by simply adding ``Let’s think step by step'' before each output. The~zero-shot accuracy is increased from 17.7\% to 78.7\% on MultiArith techmark~\cite{Roy2015} and from 10.4\% to 40.7\% on GSM8K benchmark~\cite{Cobbe2021} with an off-the-shelf 175B parameter model. As~explored by the work, the~versatility of this single prompt across very diverse reasoning tasks hints at untapped and understudied fundamental zero-shot capabilities of \textsf{LLMs}. This suggests high-level and multi-task broad cognitive capabilities may be extracted through simple prompting. 
Furthermore, an~instruction fine-tuning method is described in~\cite{Chung2022} focusing on scaling the number of tasks, scaling the model size, and~fine-tuning on chain-of-thought data. The~resulting Flan-PaLM 540B instruction-fine tuned on 1.8K tasks outperforms PALM 540B by a large margin (+9.4\% on average). 
At last, the~tuning process is also applied to find optimal values for model hyper-parameters~\cite{Schratz2019a}, which is however out of the scope of transfer~learning. 

Although transfer learning depends on the correlation between source and target domains to be effective, the~similarities between domains are not always beneficial but can be misleading to the learning. Negative transfer~\cite{Wang2019b} is the transfer process in which the target model is negatively affected by the transferred knowledge. It can be caused by several factors such as the domain relevance and the learner’s capacity to find the transferable and beneficial part of the knowledge across domains. The~work in~\cite{Wang2019b} proposes a method relying on an adversarial network to circumvent negative transfer by filtering out unrelated source data. The~harmful source data are filtered by a discriminator estimating both marginal and joint distributions to reduce the bias between source and target risks. The~experiments involving four benchmarks demonstrate the effectiveness of filtering negative transfer and the improvement of model performance under negative transfer~conditions. 
 
Transfer Learning avoids building models from scratch and largely reduces the workload of training new models, which leads to the low training task latency, {availability improvement}, and~efficient computation. In~parallel, the~required training data in the case of supervised learning is much less than required when training models from scratch. Thus, transfer learning can save expensive data-labeling efforts and drives conventional supervised learning more independent of labelled data. Regarding the edge  requirements of model performance, transfer learning facilitates the construction of personalized models specific to individual edge  environments and are expected to maintain a high model accuracy {and reliability} compared to generalized models. However, in~practice, the~model performance is determined by the quality of the source model, the~training data in a target domain, and~the correlation between the source and the target domains. Thus, the~performance varies according to the specific~configurations. 

\subsection{Meta-Learning}
%\textbf{Theory.} 
Taking the philosophy one step higher, and~focusing on learning the learning process rather than specific tasks, meta-learning~\cite{Vanschoren2018a} is an advanced learning paradigm that observes and ``remembers'' previous learning experiences on multiple learning tasks, and~then quickly learns new tasks from previous meta-data by analyzing the relation between tasks and solutions. The~meta-learning solution for \textsf{ML} tasks is is realized in two levels~\cite{Peng2021}: (i) a base learner for each task, and~(ii) a global meta-learner. The~base learner solves task-specific problems and focuses on a single task, while the meta-learner integrates them using previous learned concepts to quickly learn the associated tasks. For~a new task, meta-learning directly applies or updates the solution of the most similar task. In~the case where no similar task is registered, meta-learning exploits the relation between tasks and solutions to propose an initial reference~solution. 

Meta-learning can also be applied to all three basic machine learning paradigms: supervised learning, unsupervised, and~reinforcement learning. 
Regular supervised learning and unsupervised learning do not assume any given or predefined meta-knowledge. On~the contrary, in~supervised and unsupervised meta-learning, the~goal is not only to realize a specific task but also to find the best meta-knowledge set, enabling the base learner to learn new tasks as quickly as possible.  
Regular reinforcement learning maximizes the expected reward on a single \textsf{MDP}, while meta reinforcement learning's intention is to maximize the expected reward over various \textsf{MDPs} by learning meta-knowledge. 
To summarize, instead of learning separately model parameters $\theta$ for all base learners, meta-learning actually focuses on learning the optimal or sub-optimal meta-knowledge $\omega^\ast$ for the global meta-learner, as~formalized in Equation~(\ref{eq:meta-learning}).
\begin{equation}
	\label{eq:meta-learning}
	\omega^\ast := \underset{\omega}{arg \, min}\ 
	\bigsqcup_{t=1}^n 
	L_{\mathbb{D}^t} (g_{\omega^t}(\mathbb{D}^t, L_{\mathbb{D}^t}))
\end{equation}

\noindent where $g_{\omega^t}$ is the optimization procedure driven by the meta-knowledge ${\omega^t}$ of the task $i, i\!\in\!n$, and~$n$ is the number of the considered base learner tasks. $\mathbb{D}^t$ is the data used for learning the base task $t$, $L_{\mathbb{D}^t}$ is the corresponding loss on the given data ${\mathbb{D}^t}$, $\bigsqcup$ is the aggregation algorithm (e.g., Model-
Agnostic Meta-Learning (\textsf{MAML})~\cite{Finn2017}) that finds the optimal meta-knowledge $\omega^\ast$ by minimizing the losses across different base~learners. 

Depending on the representation of the meta-knowledge, meta-learning techniques can be divided into three categories~\cite{Huisman2021}: (i) metric-based meta-learning, (ii) model-based meta-learning, and~(iii) optimization-based~meta-learning. 

\begin{enumerate}
	\item \hl{Metric-based meta-learning} learns the meta-knowledge in the form of feature space from previous tasks by associating the feature space with model parameters. New tasks are achieved by comparing new inputs, usually with unseen labels (also known as the query set), to~example inputs (a.k.a. the support set) in the learned feature space. The~new input will be associated to the label of the example input with which it shares the highest similarity. 
	The idea behind metric-based meta-learning is similar to distance-based clustering algorithms, e.g.,~K-Nearest Neighbors (\textsf{KNN})~\cite{Zhang2017} or \textsf{K-means}~\cite{NIPS2003_23483314}, but~with a learned model containing the meta-knowledge. Being simple in computation and fast at test-time with small tasks, metric-based meta-learning is inefficient when the tasks contain a large number of labels to compare, while the fact of relying on labelled examples makes the metric-based meta-learning both specialized at and limited by the supervised learning~paradigm.
	
	\item \hl{Model-based meta-learning} \textls[15]{relies on an internal or external memory component} (i.e.,~a~model) to save previous inputs and to empower the models to maintain a stateful representation of a task as the meta-knowledge. Specifically designed for fast training, the~memory component can update its parameters in a few training steps with new data, either by the designed internal architecture or controlled by another meta-learner model~\cite{10.5555/3045390.3045585}. When given new data on a specific task, the~model-based meta-learning firstly processes the new data to train and alter the internal state of the model. Since the internal state captures relevant task-specific information, outputs can be generated for unseen labels of the same task or new tasks. Compared to the metric-based meta-learning, model-based meta-learning has a broader applicability to the three basic machine learning paradigms and brings flexibility and dynamics to the meta-learning technique via quick and dynamic model adjustment to new tasks and~data.
	
	\item \hl{Optimization-based meta-learning} revises the gradient-based learning optimization algorithm so that the model is specialized at fast learning with a few examples, as~the gradient-based optimization is considered to be slow to converge and inefficient with few learning data. Optimization-based meta-learning is generally achieved by a two-level optimization process~\cite{Finn2017}: base-learners are trained in a task-specific manner, while the meta-learner performs cross-task optimization in such a way that all base learners can quickly learn individual model parameters set for different tasks. Optimization-based meta-learning works better on wider task distributions and enables faster learning compared to the two previous meta-learning techniques. On~the other hand, the~global optimization procedure leads to an expensive computation workload, as each task’s base-learner is considered~\cite{Hospedales2021}. 
\end{enumerate}

In all the three meta-learning representations, one important characteristic of meta-learning is that during the testing phase, the~resulting models are generalized and able to deal with the data labels, inputs, and the tasks on which models were not explicitly trained during the previous learning phase. Thus, data and task generalization as well as fast learning are the two main advantages of~meta-learning. 

% \begin{table}[]
% \centering
% \begin{tabular}{|p{3cm}|p{4cm}|p{3cm}|p{3cm}|}
% \hline
% \textbf{Study} & \textbf{Method} & \textbf{Highlights} & \textbf{Use Case} \\ \hline
% \cite{Ravi2017} & LSTM-based meta-learner & Fast convergence & Few-shot learning \\ \hline
% \cite{10.5555/3045118.3045347} & Zero-shot learning & New class prediction & AwA, SUN, aPY \\ \hline
% \cite{Verma2017} & Probabilistic representation & Leverage additional unlabeled data & Superior results compared to~\cite{10.5555/3045118.3045347} \\ \hline
% \cite{Brown2020} & Large Language Models (LLMs) & Few-shot and zero-shot learning & --- \\ \hline
% \cite{Radford2021} & Contrastive Language-Image Pretraining (CLIP) & Learns from raw text and images & ImageNet zero-shot \\ \hline
% \cite{10.5555/3045390.3045585} & Memory-Augmented Neural Network (MANN) & External memory component & Omniglot, Gaussian process \\ \hline
% \cite{Belkhale2021} & Adaptive drone flight & Inference of unknown payload & Drone control with unknown payloads \\ \hline
% \cite{Finn2017} & Model-Agnostic Meta-Learning (MAML) & Gradient descent compatible & General \\ \hline
% \cite{Rajeswaran2019} & Implicit MAML (iMAML) & Approximates higher order derivatives & Robust for larger optimization paths \\ \hline
% \cite{Finn2019} & Online MAML & Continuous learning & Long term learning scenarios \\ \hline
% \end{tabular}
% \caption{Summary of different works related to Transfer Learning}
% \label{tab:transfer_learning_works}
% \end{table}

Meta-learning widens the applicability of machine learning techniques and hence is applied into various domains such as few-shot learning in image classification~\cite{Sun2021a}, zero-shot learning for natural language processing~\cite{Kojima2022}, robot control~\cite{Gupta2018a}, and~reasoning~\cite{Griffiths2019}. Several surveys study the existing meta-learning techniques and works. In~addition to~\cite{Huisman2021}, meta-learning in neural networks is studied in~\cite{Hospedales2021}. This~work proposes a taxonomy and organizes the paper according to the representation of meta-knowledge, the~meta-level optimizer, and~the global objective of the meta-learning. Based on the type of the leveraging meta-data during the learning process, Vanschoren~et~al.~\cite{Vanschoren2018a} categorizes meta-learning techniques into: (i)  learning from model evaluations, (ii) learning from task properties, and~(iii) learning from prior models. 
Wang~et~al.~\cite{Wang2020b} review the metric-based few-shot learning methods targeting the problem of data-intensive applications with little training data. Methods are grouped into three perspectives: data, model, and~algorithm. The~pros and cons of each perspective is analyzed in the work. The~specific related work on Meta-Learning is summarized in Table~\ref{tab:metalearning} and described as~follows.

\begin{figure}[H]
	\centering
	\includegraphics[width=\textwidth]{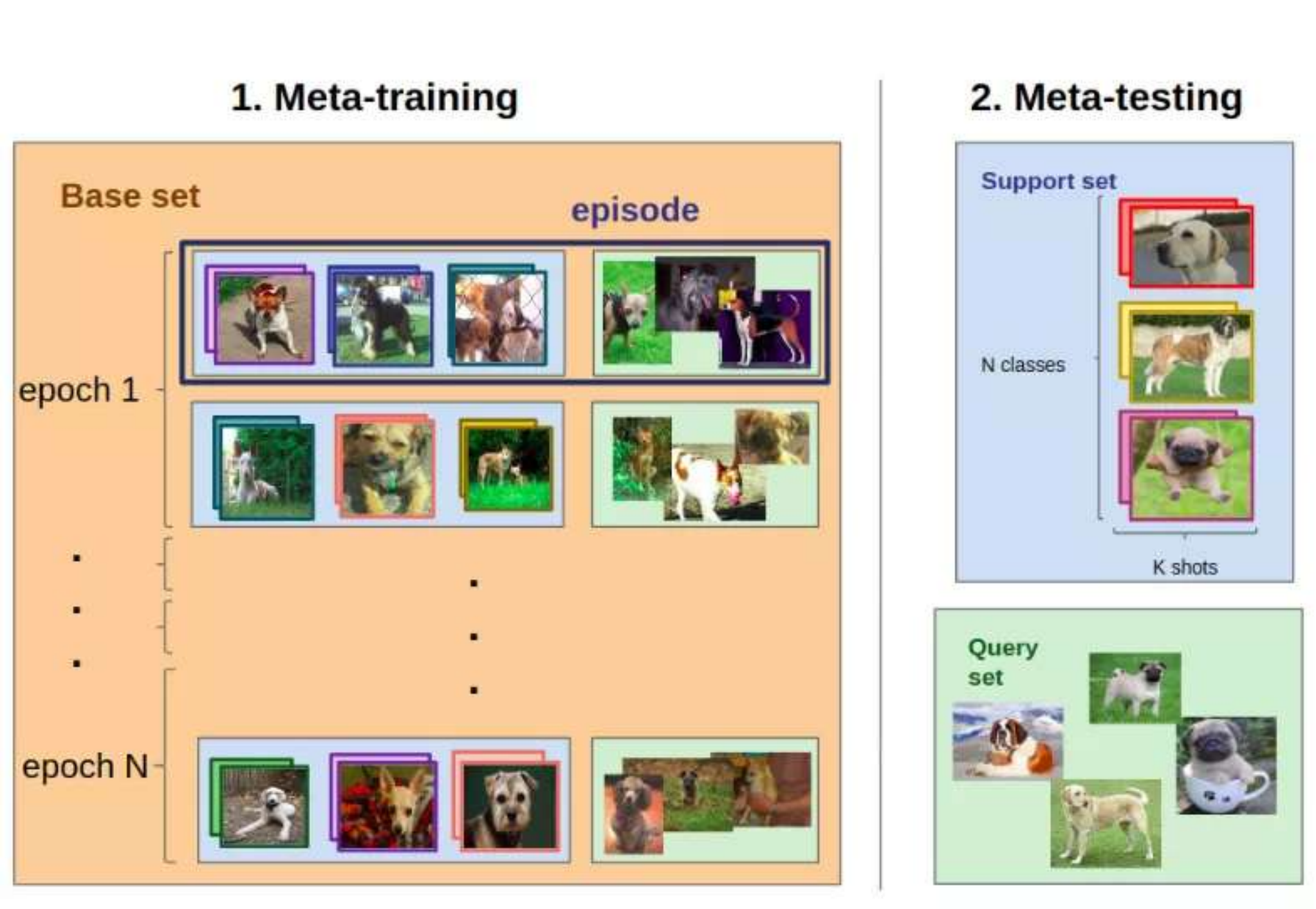}
	\caption{N-way K-shot learning setup~\cite{Bennequin2019}.}
	\label{fig:4}
\end{figure}

The main challenge in meta-learning is to learn from prior experiences in a systematic and data-driven way~\cite{Vanschoren2018a}. For~the metric-based meta-learning, a~typical configuration of few-shot learning is 
%illustrated in Figure~\ref{fig:4} as 
\textsf{N-way} \textsf{K-shot} learning~\cite{Chen2019, Bennequin2019}. \textsf{N-way} refers to the number of classes $N$ existing in the support set of the meta-testing phase. \textsf{K-shot} refers to the number of data samples $K$ in each class in the support set. 
The few-shot learning tackles the supervised learning problem where models need to quickly generalize after training on few examples from each class. During~the meta-training phase, the~training dataset is divided into support set and query set, and~the data embeddings are extracted from all training data, i.e.,~images. Each image from the query set is classified according to its embedding similarity with images from the support set. The~model parameters are then updated by back-propagating the loss from the classification error of the query set. After~training, the~meta-testing phase classifies unseen labels from the meta-training phase 
(i.e., in~Figure~\ref{fig:4}, images of unseen dog breeds are given during meta-testing) 
by use of the new support~set. 

\begin{table}[H]
\caption{{\hl{Summary} %MDPI: please check if the alignment keep and if vertical line can be removed? -> alignment keeps, vertical lines are removed. 
 of related work on Meta-Learning.}}
\begin{adjustwidth}{-4.5cm}{0cm}
\centering
\small
\begin{tabular}{p{4cm} p{5cm} p{8cm}}
 \noalign{\hrule height 1pt}
\textbf{Works} & \textbf{Methods} & \textbf{Results and Insights}\\
\hline
Metric-based: Few-shot learning~\cite{Chen2019, Bennequin2019} & N-way K-shot learning. & Efficient for small tasks, but~inefficient for large label sets.\\
\hline
Metric-based: LSTM-based meta-learner~\cite{Ravi2017} & Captures both short-term and long-term knowledge. & Rapidly converges a base learner to a locally optimal solution and in the meantime learns a task-common initialization as the base learner.\\
\hline
Metric-based: Zero-shot learning~\cite{10.5555/3045118.3045347} & Two linear layers network modeling relationships among features, attributes, and~classes. & Outperforms state-of-the-art approaches on several datasets by up to 17.\%\\
\hline
Metric-based: Class distribution learning~\cite{Verma2017} & Represents each class as a probability distribution, defined as functions of the respective observed class attributes. & Leverages additional unlabeled data from unseen classes and improves estimates of their class-conditional distributions; Superior results in the same datasets compared to~\cite{10.5555/3045118.3045347}.\\
\hline
Metric-based: CLIP~\cite{Radford2021} & Pre-training of vision models from raw-text-describing images. %Check meaning retained
& Benchmarks on over 30 CV datasets produce competitive results with fully supervised baselines\\
\hline
Model-based: MANN~\cite{10.5555/3045390.3045585} & Model-based controller with an external memory component and Least Recently Used Access (LRUA) method. & Superior to LSTM in two meta-learning tasks.\\
\hline
Model-based: Drone adaptation~\cite{Belkhale2021} & Dynamics model with shared dynamics parameters and adaptation parameters. & Drone control with unknown payloads; autonomous determination of payload parameters and adjustment of flight control; performance improvement over non-adaptive methods on several suspended payload transportation tasks.\\
\hline
Optimization-based: MAML~\cite{Finn2017} & Gradient descent with model-specific updates. & General optimization tasks; simple and general, but~higher-order derivatives potentially decrease performance.\\
\hline
Optimization-based: iMAML~\cite{Rajeswaran2019} & Approximation of higher-order derivatives. & General optimization tasks; more robust for larger optimization paths with same computational costs.\\
\hline
Optimization-based: online MAML~\cite{Finn2019} & Extension of MAML for online learning scenarios. & Continuous learning from newly generated data; strong in model specialization, but~computation cost grows over time.\\
 \noalign{\hrule height 1pt}
\end{tabular}
\end{adjustwidth}
\label{tab:metalearning}
\end{table}

The work~\cite{Ravi2017} proposes a Long Short-Term Memory (\textsf{LSTM})-based meta-learner model in the few-shot regime. This is done to learn the exact optimization algorithm used to train another neural network classifier as the base learner: the meta-learner is trained to capture both short-term knowledge within a task and long-term common knowledge among all the tasks. This way, the~meta-learner is able to rapidly converge a base learner to a locally optimal solution on each task and in the meantime learn a task-common initialization as the base learner. 
As a step further, zero-shot learning~\cite{Wang2019a} does not require any example data as support set to perform new tasks or classify new classes which the model has not observed during the training phase. A~simple zero-shot learning approach is introduced in~\cite{10.5555/3045118.3045347} to model the relationships among features, attributes, and~classes as a two linear layers network, where the weights of the second layer are not learned but are given by the environment. During~the inference phase with new classes, the~second layer is directly given so that the model can make predictions on the new labels. Despite  being simple, the~experiment results outperformed the state-of-the-art approaches on the datasets of Animals with Attributes (AwA)~\cite{5206594}, 
SUN attributes (SUN)~\cite{Patterson2012a}, and~aPascal/aYahoo objects (aPY)~\cite{5206772} by up to 17\% at the publication~time. %Check meaning retained

\textls[-15]{Unlike~\cite{10.5555/3045118.3045347} representing classes as fixed embeddings in a feature space, Verma~et~al.~\cite{Verma2017}} represent each class as a probability distribution. The~parameters of the distribution of each seen and unseen class are defined as functions of the respective observed class attributes. This allows the leveraging of additional unlabeled data from unseen classes and the improvement of the estimates of their class-conditional distributions via transductive or semi-supervised learning. Evaluations demonstrate superior results in the same datasets compared to~\cite{10.5555/3045118.3045347}. 
In parallel to \textsf{CV}, the~pre-trained large language models (\textsf{LLMs}) have proven to be excellent for few-shot learner~\cite{Brown2020} and zero-shot learner~\cite{Kojima2022}. %Check meaning retained
Furthermore, Contrastive Language-Image Pre-training (\textsf{CLIP})~\cite{Radford2021} learns computer vision models directly from raw-text-describing images, which leverages a much boarder source of supervision instead of specific data labels. %Check meaning retained
The~pre-training of predicting ``which caption goes with which image?'' is realized on a dataset of 400 million image and text pairs from the Internet. After~pre-training, natural language is used to reference learned visual concepts and describe new ones, enabling zero-shot transfer of the model to downstream tasks. The~work matches the accuracy of the ResNet-50 model on ImageNet zero-shot without dataset-specific training, and~benchmarks on over 30 CV datasets produce competitive results with fully supervised~baselines. 

\vspace{-9pt}

\textls[-15]{As for model-based meta-learning, Memory-Augmented Neural Network (\textsf{MANN})~\cite{10.5555/3045390.3045585}} contains a model-based controller, either feed-forward network or \textsf{LSTM}, to~interact with an external memory component for memory retrieval and update. During~training, the~model learns to bind data representations to their labels regardless of the actual content of the data representation or label, and~then the model maps these bound representations to appropriate classes for prediction. The~memory writing and reading are powered by the proposed Least Recently Used Access (\textsf{LRUA}) method, and~the \textsf{MANN} displays a performance superior to an  \textsf{LSTM} in two meta-learning tasks on the Omniglot classification dataset~\cite{Lake2019} and sampled functions from a Gaussian process for~regression. 

A more concrete use case is illustrated in~\cite{Belkhale2021} to adapt drones to flight with unknown payloads, in~which drones are expected to autonomously determine the payload parameters and adjust the flight control accordingly. During~the training, a~dynamics model with shared dynamics parameters and adaptation parameters are trained over $K$ different payloads. During~the testing, the~robot infers the optimal latent variable representing the unknown payload by use of the learned dynamics parameters and the new sensed data. A~model-predictive controller (\textsf{MPC}) then uses the trained dynamic model to plan and execute drone actions that follow the specified flight path. Experiments demonstrate the performance improvement of the proposed method compared to non-adaptive methods on several suspended payload transportation~tasks.

With respect to optimization-based meta-learning, \textsf{MAML}~\cite{Finn2017} is a general optimization algorithm, compatible with any model that learns through gradient descent. In~\textsf{MAML}, model-specific updates are made by one or more gradient descent steps. Instead of using second derivatives for meta-optimization of models, the~meta-optimization proposes the First-Order \textsf{MAML} (\textsf{FOMAML}) to ignore the second derivative during \textsf{MAML} gradient computation to be less computationally expensive. \textsf{MAML} has obtained much attention due to its simplicity and general applicability. In~the meantime, ignoring higher-order derivatives potentially decreases the model performance, and~thus the \textsf{iMAML}~\cite{Rajeswaran2019} approximates these derivatives in a way that is less memory-consuming. While the \textsf{iMAML} is more robust for larger optimization paths, the~computational costs roughly stay the same compared to \textsf{MAML}. Furthermore, online \textsf{MAML}~\cite{Finn2019} extends the \textsf{MAML} to online learning scenarios where models continuously learn in a potentially infinite time horizon from newly generated data and adapt to environmental changes. Being strong in model specialization, the~computation cost, however, keeps growing over~time.

Overall, meta-learning reduces supervised learning’s dependency on labelled data by enabling models to learn new concepts quickly, which makes meta-learning particularly suitable for the edge side in the sense that it accelerates the training task. Another major advantage of meta-learning is the generalization capability that it brings to models to solve diverse tasks and the potential to realize general \textsf{ML}. Computational resource efficiency is higher for multiple model training, which leads to optimized energy consumption and cost optimization. Nevertheless, the~global optimization procedure of optimization-based meta-learning may yet lead to expensive computation workload according to the number of base learners. Additional computation on the support dataset for metric-based meta-learning introduces extra workload during inference according to the dataset size (in such a case, the~use of metric-based meta-learning is usually avoided). {The generalization capability makes this application versatile and potentially more reliable, on~the one hand; while on the other hand, its~resource-intensive nature may impact its availability in resource-constrained environments.}

\subsection{Self-Supervised~Learning}

In contrast to supervised learning or reinforcement learning, human beings’ learning paradigm is barely supervised and rarely reinforced. Self-Supervised Learning (\textsf{SSL}) is an unsupervised learning paradigm that uses self-supervision from original data and extracts higher-level generalizable features through unsupervised pre-training or optimization of contrastive loss objectives~\cite{Peng2021}. These learned feature representations are generalized and transferable, and~thus can be tuned later to realize downstream tasks, and~the pre-trained models are used as initial models to avoid training from scratch. During~self-supervised learning, data augmentation techniques~\cite{Shorten2019, Shorten2021} are widely applied for contrast or generation purposes, and~data labels are not required since pseudo labels can be estimated from trained models on similar~tasks. 

\begin{figure}[H]
	\centering
	\includegraphics[width=\textwidth]{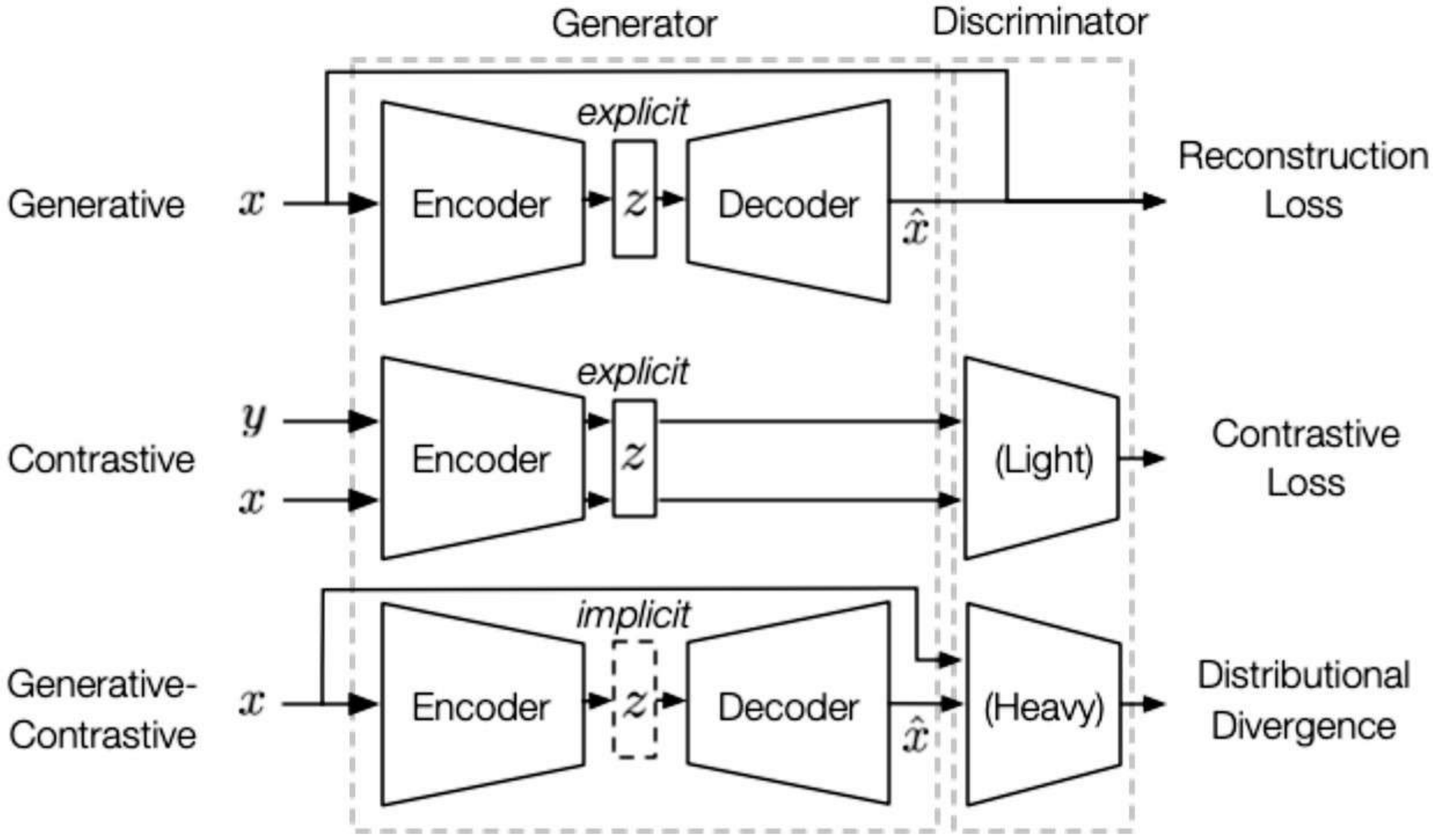}
	\caption{Self-Supervised Learning Architecture~\cite{Liu2021}.}
	\label{fig:ssl}
\end{figure}

According to the loss objectives driving the training process, self-supervised learning can be summarized into three categories~\cite{Liu2021}: (i) generative learning, (ii) contrastive learning, and~(iii) adversarial learning, as~a combination of generative and contrastive learning. {The architectures of the three categories are illustrated in Figure~\ref{fig:ssl} to show the transformation from the traditional supervised learning process to a self-supervised learning approach, which can be particularly useful in Edge scenarios where label data are~scarce.} 

\begin{itemize}
	\item \hl{Generative Learning:} Generative learning trains an encoder to encode the input into an explicit vector and a decoder to reconstruct the input from the explicit vector. The~training simulates pseudo labels for unlabeled data and is guided by the reconstruction loss between the real input and the reconstructed~input. 
 
	\item \hl{Contrastive learning:} Contrastive learning trains an encoder to respectively encode inputs into explicit vectors and measure similarity among inputs. The~contrastive similarity metric is employed as the contrastive loss for model training. During~the training, the~contrastive learning calibrates label-free data against themselves to learn high-level generalizable~representations. 
 
	\item \hl{Adversarial Learning:} Adversarial learning trains an encoder--decoder to generate fake samples and a discriminator to distinguish them from real samples in an adversarial manner. In~other words, it learns to reconstruct the original data distribution rather than the samples themselves, and~the distributional divergence between original and reconstructed divergence is the loss function to minimize during the training phase. The~point-wise (e.g., word in texts) objective of the generative \textsf{SSL} is sensitive to rare examples and contrary to the high-level objective (e.g., texts) in classification tasks, which may result in inherent results without distribution data. %Check meaning retained
	Adversarial \textsf{SSL} abandons the point-wise objective and uses the distributional matching objectives for high-level abstraction learning. In~the meantime,  adversarial learning preserves the decoder component abandoned by the contrastive \textsf{SSL} to stabilize the convergence with more expressiveness.  
\end{itemize}

As an emerging field, self-supervised learning has received significant research attention. A~comprehensive survey of the three above-mentioned \textsf{SSL} categories is presented in~\cite{Liu2021} including existing methods and representative works.  Research works across several modalities of image, text, speech, and~graphs are reviewed and compared in~\cite{Ericsson2022}.
Digging in specific application domains, \textsf{SSL} works for visual feature learning and \textsf{NLP} representation learning are respectively analyzed in~\cite{Jing2021,Kalyan2021}; since graph-structured data are widely used and available over network, efforts on \textsf{SSL} of graph representation are compared in~\cite{Xie2022} to facilitate downstream tasks based on graph neural networks. The~related work on SSL is summarized in Table~\ref{table:ssl} and introduced as~follows. 

\begin{table}[H]
\caption{{\hl{Summary} %MDPI: please check if the alignment keep and if vertical line can be removed?-> alignment keeps, vertical lines are removed. 
 of related work on Self-Supervised Learning.}}
\centering
\small
\begin{adjustwidth}{-4.5cm}{0cm}
\begin{tabular}{p{2.9cm}p{8cm}p{6.1cm}}
 \noalign{\hrule height 1pt}
\textbf{Works} & \textbf{Methods} & \textbf{Results and Insights} \\
\hline
Masked Prediction Method~\mbox{\cite{Baevski2022, Brown2020, Pathak2018, Hu2020a}} & Generative SSL model that trains by filling in intentionally removed and missing data. & Effective to build pre-trained models in different areas like language, speech recognition, image regions, and~graph edges. \\
\hline
Dat2Vec~\cite{Baevski2022} & A general framework for SSL in speech, NLP and CV data. Predicts latent representations of the full input data based on a masked view of the input. & Performs competitively on major benchmarks in speech recognition, image classification, and~natural language understanding. \\
\hline
MoCo~\cite{He2020} & Contrastive SSL with two encoders. Encodes two augmented versions of the same input images into queries and keys. & Outperforms its supervised pre-training counterpart in several CV tasks. \\
\hline
SimCLR~\cite{Chen2020} & A framework for contrastive learning of visual representations. Employs various image augmentation operations and contrastive cross-entropy loss. & Achieves a relative improvement of 7\% over previous state-of-the-art, matching the performance of a supervised ResNet-50. \\
\hline
BiGANs~\cite{Donahue2017} & Projects data back into the latent space to boost auxiliary supervised discrimination tasks. & Captures the difference at the semantic level without making any assumptions about the data. \\
\hline
BigBiGAN~\cite{Donahue2019} & Extends the BiGAN model on representation learning by adding an encoder and correspondingly updating the discriminator. & Achieves the state-of-the-art in both unsupervised representation learning on ImageNet, and~unconditional image generation. \\
\hline
Image Completion~\cite{Iizuka2017} & Uses adversarial SSL for image completion by training a global and local context discriminator networks. & Can complete images of arbitrary resolutions by filling in missing regions of any shape. \\
\hline
Image Inpainting~\cite{Tran2020} & Uses adversarial SSL for image completion in masked chest X-ray images. & Facilitates abnormality detection in the healthcare domain. \\
\hline
SSL with Federated Learning~\cite{Zhuang2022} & Empirical study of federated SSL for privacy preserving and representation learning with unlabeled data. & Tackles the non-IID data challenge of FL. \\
\hline
SSL with Meta Learning~\cite{Peng2021} & Reviews the intersection between SSL and meta-learning. & Shows how SSL can improve the generalization capability of models. \\
\hline
SSL with Transfer Learning~\cite{Mao2020} &  The SSL applications within the transfer learning framework & Introduces methods for designing pre-training tasks across different
domains. \\
 \noalign{\hrule height 1pt}
\end{tabular}
\label{table:ssl}
\end{adjustwidth}
\end{table}

Generative \textsf{SSL} often applies the masked prediction method~\cite{Baevski2022} to train the model to fill in the intentionally removed and missing data. For~instance, in the work~\cite{Brown2020}, generative learning generates words in sentences in \textsf{NLP} by masking the words to generate in each step and updates the model parameters by minimizing the distance between the generated word and the masked word in the text. The~same masking methods have proven to be effective to build pre-trained models by hiding speech time slices~\cite{Baevski2020}, image regions~\cite{Pathak2018}, and~graph edges~\cite{Hu2020a} in speech~recognition. 

In a multi-modal setting context, a~more general framework is introduced in~\cite{Baevski2022} as \textsf{dat2vec} for speech, \textsf{NLP}, and \textsf{CV} data. The~idea is to predict latent representations of the full input data based on a masked view of the input in a self-distillation setup using a standard Transformer architecture.  
Instead of predicting modality-specific targets such as words, visual tokens, or~units of human speech, \textsf{data2vec} predicts contextualized and multi-modal latent representations. Experiments on the major benchmarks of speech recognition Librispeech~\cite{Panayotov2015}, image classification ImageNet-1K, and~natural language understanding GLUE demonstrate a competitive performance to predominant approaches. Generative \textsf{SSL} is the mainstream method in \textsf{NLP} to train \textsf{LLMs} with texts from the Internet, while on the other hand \textsf{SSL} reveals less competitive results than contrastive \textsf{SSL} in \textsf{CV} domains of which the classification is the main~objective. 
 
Contrastive \textsf{SSL} creates multiple views of inputs~\cite{Tian2020} and compares them in the representation space to solve discrimination problems. During~the learning, the~distance between multi-views of the same data sample is minimized and the distance between different data samples is maximized. Negative sampling is a common for contrastive learning, but~this process is often biased and time-consuming. Momentum Contrast (\textsf{MoCo})~\cite{He2020} uses two encoders, an~encoder and a momentum encoder, to~encode two augmented versions of the same input images into queries and keys,  respectively.  During~the training, positive pairs are constructed from queries of keys of current mini-batch, while negative pairs are constructed from queries of current mini-batch and keys from previous mini-batches to minimize the contrastive loss function \textsf{InfoNCE}~\cite{Oord2018}. In~the experiments, \textsf{MoCo} outperforms its supervised pre-training counterpart in seven \textsf{CV} tasks on datasets including PASCAL and~COCO. 

To avoid explicitly using negative examples and prevent feature collapse, several data augmentation operations for images (e.g., original, crop, resize, color distort, Gaussian noise and blur, etc.) are introduced in~\cite{Chen2020} as a simple framework for contrastive learning (\textsf{SimCLR}) of visual representations. The~learning with regularization and contrastive cross entropy loss benefits from a larger batch size and a longer training compared to the supervised counterpart: \textsf{SimCLR} achieves 76.5\% top-1 accuracy, which is a 7\% relative improvement over the previous state-of-the-art, matching the performance of a supervised ResNet-50. Contrastive learning is found to be useful for almost all visual classification tasks due to the class-invariance modeling between different image instances but does not present a convincing result in the \textsf{NLP} benchmarks. The~theory and applications of contrastive \textsf{SSL} to the domains such as \textsf{NLP} and graph learning where data are discrete and abstract is still~challenging. 
 
Inspired by the Generative Adversarial Networks (\textsf{GAN})~\cite{Radford2016}, adversarial \textsf{SSL} either focuses on generating with the learned complete representation of data or reconstructing the whole inputs with partial ones. Instead of learning from the latent distribution of task-related data distributions, Bidirectional Generative Adversarial Networks (\textsf{BiGANs})~\cite{Donahue2017} projects data back into the latent space to boost auxiliary supervised discrimination tasks. The~learned distribution does not make any assumption about the data  and thus captures the difference in the semantic level. \textsf{BigBiGAN}~\cite{Donahue2019} discovers that a \textsf{GAN} with deeper and larger structures produces better results on downstream task and extends the \textsf{BigGAN} model on representation learning by adding an encoder and correspondingly updating the discriminator. Evaluations of the representation learning and generation capabilities of the \textsf{BigBiGAN} models achieve the state-of-the-art in both unsupervised representation learning on ImageNet, and~unconditional image~generation. 

Adversarial \textsf{SSL} proves to be successful in image generation and processing, while still limited in \textsf{NLP}~\cite{Clark2020} and graph learning~\cite{Dai2018}. Alternatively, in-painting is a common use case for Adversarial \textsf{SSL} to reconstruct the entire inputs by filling in target regions with a relevant content, which allows the model to learn representations of different regions as well in order to process specific objects in images, detect anomalies in regions or reconstruct 3D images from 2D. A~method of image completion is presented in~\cite{Iizuka2017} to complete images of arbitrary resolutions by filling in missing regions of any shape. A~global discriminator and a local context discriminator are trained to distinguish real images from completed ones. The~global discriminator assesses if the image is coherent as a whole, while the local discriminator ensures the local consistency of the generated patches at the completed region. The~image completion network is then trained to fool both context discriminator networks. A~similar work is reported in~\cite{Tran2020} to generate regions in masked chest X-ray images to facilitate the abnormality detection in the healthcare~domain.
  
As the key method to alleviate the data labelling and annotation dependency, \textsf{SSL} demonstrates the boosting capability to power other learning paradigms, and~the resulting solutions absorb merits from \textsf{SLL} and its incorporating learning paradigms. Federated \textsf{SSL} is empirically studied in~\cite{Zhuang2022} for both privacy preserving and representation learning with unlabeled data. A~framework is also introduced to tackle the \textsf{non-IID} data challenge of \textsf{FL}. The~intersection between \textsf{SSL} and meta-learning is reviewed in~\cite{Peng2021} showing models can best contribute to the improvement of model generalization capability. The~models trained by \textsf{SSL} for pretext tasks with unlabeled data can be used by transfer learning to build state-of-the-art results. The~self-supervised learning methods and their applications within the transfer learning framework are reviewed and summarized in~\cite{Mao2020}.

Overall, the~essential advantage of \textsf{SSL} is its capability to leverage the tremendous amount of unlabeled data to learn latent representations;~thus, the~labelled data dependency is largely alleviated during the learning process. The~learned data representation via pretext task is in high-level generalization and can be easily used by downstream tasks to provide higher performance in various benchmarks, {and thus SSL can improve reliability by learning a more comprehensive representation of the data.} Although the arithmetic operations required by the training and task latency rises in certain learning setups with larger batch and more epochs, the~testing performance is boosted as well. The~final cost of a training task with \textsf{SSL} is much less compared to the same task requiring the manual labelling of data. {While SSL can improve availability by reducing the need for labeled data, it often requires more computational resources for training as the model learns to understand the data structure and make predictions, which could cause an availability challenge on resource-constrained edge devices. }

\subsection{Other Learning~Paradigms}	

Besides the four major learning techniques fitting to  \textsf{Edge ML}, introduced in previously, in~this section we briefly review relevant \textsf{ML} paradigms that potentially improve  \textsf{Edge ML} solutions by satisfying a subset of its~requirements. 

\subsubsection{Multi-Task~Learning}		

Instead of building $n$ models for $n$ tasks, Multi-Task Learning (\textsf{MTL}) aims at using one \textsf{ML} model to realize multiple correlated tasks at the same time~\cite{Zhang2021c}. This is commonly achieved by training an entire model for all tasks, consisting of a commonly shared part among all tasks and a task-independent part. The~commonly shared part of the model learns the common representation and task relations from all tasks’ inputs, while the task-independent part computes and generates the final output for each task individually. During~the multi-task learning, the~model is trained in a way that data are mutualized among tasks to discover implicit task correlations. The~learning process helps the model better find relevant features for each task and reduces the risk of over-fitting, so that all tasks' performance is improved via relevant features and tasks correlation~\cite{LIW2022}. Among~the multiple tasks, each task can be a general learning task such as supervised tasks (e.g., classification or regression problems), unsupervised tasks (e.g., clustering problems), or~reinforcement~learning. 

\begin{figure}[H]
	\centering
	\includegraphics[width=\textwidth]{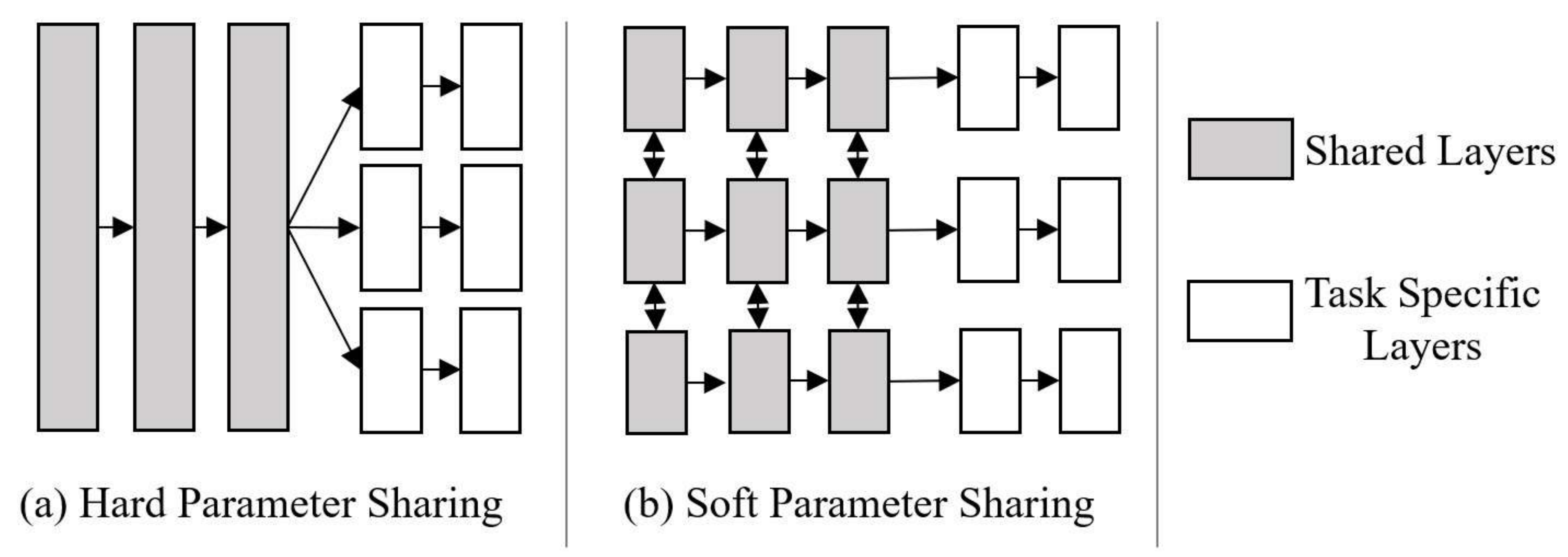}
	\caption{\textsf{\hl{MTL} %MDPI: Please add the left bracket in the image, e.g., “a)” should be “(a)”. -> updated.
} modelling~structure.}
	\label{fig:f6}
\end{figure}

From the modelling perspective, \textsf{MTL} can be divided into: (i) hard parameter sharing and (ii) soft parameter sharing~\cite{Ruder2017}. The~hard parameter sharing generally shares the hidden layers among all tasks, while keeping several task-specific output layers. On~the other hand, soft parameter sharing creates a set of parameters for each task of a similar structure, and~the distance among the task parameters is then regularized during training~\cite{Yang2019a} in order to encourage the parameters to be similar. The~modelling structure is illustrated in Figure~\ref{fig:f6}. The~choice of the two modelling depends on the similarity among input data and task~relation.

A number of works of \textsf{MTL} are surveyed and compared in~\cite{Zhang2021c, Ruder2017, liu2022bit}, illustrating the overview of the literature and recent advances. One important research challenge of \textsf{MTL} lies in the multi-task modelling to take into account task and data relations for parameter structure sharing. An~\textsf{MTL} model directly at the edge of the network is introduced in~\cite{Rago2020} for traffic classification and prediction. Based on autoencoders as the key building blocks for learning common features, the~model anticipates information on the type of traffic to be served and the resource allocation pattern requested by each service during its execution. Simulation results produce higher accuracy and lower prediction loss compared to a single-task schema. The~on-edge multi-task transfer learning is studied in~\cite{Chen2020b}, tackling data scarcity and resource constraints for task allocation. Instead of treating individual tasks equally, the~work proposes to measure the impact of tasks on the overall decision performance improvement and quantify task importance with a Data-driven Cooperative Task Allocation (\textsf{DCTA}) approach. Experiments show that  \textsf{DCTA} reduces  task latency by $3.24\times$, and~saves 48.4\% energy consumption compared with the state-of-the-art when solving the task allocation with task importance for \textsf{MTL}. %Check meaning retained

Via common layers sharing among tasks, model parameters in \textsf{MTL} are largely decreased compared to multiple individual task models, and~thus the computational workload is lower for the multiple task model. This leads to an improvement in task latency and computation efficiency. Via the learning of more relevant features and task correlations, the~performance for correlated tasks is boosted, {while the availability and the reliability are improved.} Overall, in~the context where multiple correlated tasks need to be performed, the~\textsf{MTL} brings an efficient method for energy and cost optimization, making it suitable for the~edge.

\subsubsection{Instance-Based~Learning}	
Instance-based Learning (\textsf{IBL})~\cite{Aha1991}, also called memory-based learning or lazy learning, compares new instances with already-seen instances to perform supervised learning tasks. Instead of learning an explicit representation mapping between features and instance labels, and~predicting based on the learned representation, the~key idea of \textsf{IBL} is to uniquely rely on seen instances to predict new instances. Commonly applied techniques of \textsf{IBL} are \textsf{kNN}, Radial Basis Function (\textsf{RBF})~\cite{Ghosh2001}, and~Case-Based Reasoning (\textsf{CBR})~\cite{Watson1994}. Among~these techniques, \textsf{kNN} is widely used as a non-parametric model which simply retains all of the training instances and uses all of them to predict new instances based on a similarity distance between instances. In~contrast to the metric-based meta-learning which generalizes the learned representation to unseen classes or tasks, \textsf{IBL} is suitable for rapidly realizing supervised learning tasks without generalization when the number of labels and retrained instances are small. Moreover, the~technique can be easily extended to predict previously unseen instances by simply adding unseen instances in the prediction process. On~the other hand, the~computational complexity of \textsf{IBL} grows exponentially with the number of retained instances and the number of available labels, making the learning paradigm not suitable for performing large supervised~tasks. 

 A Distributed storage and computation \textsf{kNN} algorithm (\textsf{D-kNN}) is introduced in~\cite{Zhang2020}. It is based on cloud-edge computing for cyber--physical--social systems. The~main contribution of the work lies in the optimization of distributed computation and storage of \textsf{kNN} and the efficient searching at distributed nodes to reduce the complexity of the algorithm. A~\textsf{CBR} approach is described in~\cite{Gonzalez-Briones2018} to optimize energy consumption in smart buildings. The~approach is based on a multi-agent architecture deployed in a cloud environment with a wireless sensor network, where the agents learn human behaviors through \textsf{CBR}-enabled neural networks and manage device usage. Experiments in office buildings achieve an average energy saving of 41\%.

\textsf{IBL} alleviates the labelled data dependency by reducing the amount of required labelled data to perform supervised learning tasks. Since the computational complexity of \textsf{IBL} scales with the problem complexity, the~task latency, computation efficiency, {availability and reliability}, cost and energy consumption vary according to the specific task setup. The~final performance of a model depends on the representativeness and the distribution of the instances as~well.

\subsubsection{Weakly Supervised~Learning}	

Weakly Supervised Learning (\textsf{WSL}) comprises a family of learning techniques that train models to perform supervised tasks with noisy, limited, or~imprecise labelled data from limited data labelling capacity~\cite{AlexRatnerParomaVarmaBradenHancock2019}. Although~the thorough labelling of edge data is not realistic to achieve by edge users in a continuous basis, the~assumption can be made that users or edge applications can casually provide data labelling assistance under consensus. The~casual data labelling in such a context may produce noisy, imprecise, or~an insufficient amount of labelled data for supervised learning, and~correspondingly requires specific learning paradigms to tackle the weak supervision~problem.  

According to the weakness of the labelled data quality, the~problem of \textsf{WSL} can be divided into three categories~\cite{Zhou2018}: (i) incomplete supervision, (ii) inexact supervision, and~(iii)  inaccurate~supervision. 

\begin{itemize}
	\item \hl{Incomplete supervision} refers to the problem that a predictive model needs to be trained from the ensemble of labelled and unlabeled data, where only a small amount of data is labelled, while other available data remain~unlabeled.
 
	\item \hl{Inexact supervision} refers to the problem that a predictive model needs to be trained from data with only coarse-grained label information. The~multi-instance learning~\cite{Wei2017} is a typical learning problem of incomplete supervision where training data are arranged in sets, and~a label is provided for the entire set instead of the data~themselves.
 
	\item \hl{Inaccurate supervision} concerns the problem that a predictive model needs to be trained from data that are not always labelled with ground-truth. A~typical problem of inaccurate supervision is label noise~\cite{Frenay2014}, where mislabeled data are expected to be corrected or removed before model training. 
\end{itemize}

Aiming the three problems of labelled data, weakly supervised learning brings techniques able to train models from data with low-quality labels and perform supervised~tasks.

\begin{figure}[H]
	\centering
	\includegraphics[width=\textwidth]{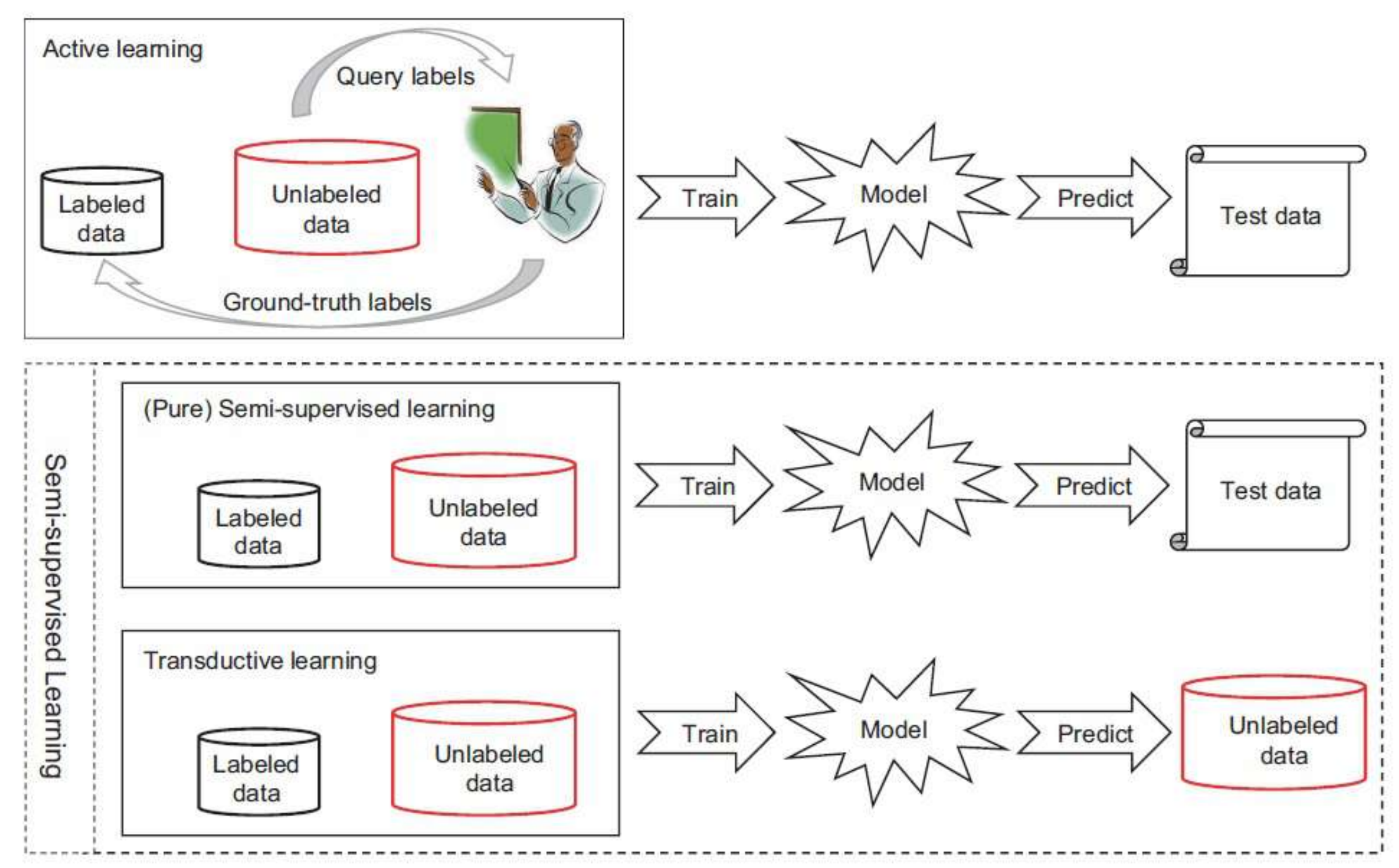}
	\caption{Incomplete Supervised Learning process~\cite{Zhou2018}.}
	\label{fig:f7}
\end{figure}

Existing work on \textsf{WSL} is introduced and summarized in~\cite{Zhou2018} and then further developed in~\cite{Nodet2021} by leveraging the data quantity and adaptability. In~what relates to  the incomplete supervision problems, active learning~\cite{MohriRostamizadehTalwalkar18}, inductive semi-supervised learning~\cite{VanEngelen2020}, and~transductive learning~\cite{Rahman_2019_ICCV} are three typical solutions for supplementing data labelling. 
{The process of the three learning paradigms for incomplete supervision is illustrated in Figure~\ref{fig:f7} to provide a visual understanding of how these methods evolve from the conventional supervised learning paradigm and operate in scenarios of partial data labelling.}
Active learning is a technique where the learner interactively collects training data, typically by querying an oracle to request labels for new data in order to resolve ambiguity during the learning process~\cite{MohriRostamizadehTalwalkar18}. Instead of querying all collected data points, the~active learning goal is to only query the most representative data and use them for model training. The~number of data used to train a model this way is often much smaller than the number required in conventional supervised learning, while the key idea behind it is that a learning paradigm can achieve higher accuracy with fewer training labels, if~it is allowed to choose the data from which it learns~\cite{Settles2010}.

Without queries, inductive semi-supervised learning labels the data with the help of the available labelled data and then trains the model~\cite{VanEngelen2020}. The~general process of semi-supervised learning is to firstly train a small model with the available labelled data to classify the unlabeled data, and~then trains the final model with all data. Such an idea is driven by the assumption that similar data produce similar outputs in supervision tasks, and~unlabeled data can be helpful to disclose which data are similar. 
Instead of training a small model to predict the unlabeled data, transductive learning~\cite{Rahman_2019_ICCV} derives the values of the unknown data with unsupervised learning algorithms and labels the unlabeled data according to the clusters to which they belong. Then a model is trained by use of both the previously available and the newly labeled data. Compared to inductive semi-supervised learning, transductive learning considers all data when performing the data labeling that potentially improve the data labeling results. On~the other hand, due to the fact no model is built for labelling, an~update in the dataset will result in the repetition of the whole learning process. Active learning, inductive semi-supervised learning, and~transductive learning are efficient in the situation where the acquisition of unlabeled data is relatively cheap while labeling is~expensive. 

Regarding the inexact supervision, multi-instance learning has been successfully applied to various tasks such as image classification~\cite{pmlr-v143-sharma21a}, relation extraction~\cite{Eberts2021}, localization~\cite{Luo2020}, and~healthcare~\cite{Raju2020}. The~main idea behind is to adapt single instance supervised learning algorithms for instance discrimination to the multi-instance representation for set discrimination. For~the label noise problem, label smoothing~\cite{Muller2019} is a regularization technique that introduces noise for the labels and can improve both predictive performance and model calibration. The~effect of label smoothing on model training with label noise is studied in~\cite{Gao2016, pmlr-v119-lukasik20a}, showing that the label-smoothing approach incorporating labeled instance centroid and its covariance reduces the influence of noisy labels during training~\cite{Gao2016}. Label smoothing is also competitive with loss-correction under label noise~\cite{pmlr-v119-lukasik20a}. Moreover, loss correction is studied in~\cite{pmlr-v97-arazo19a} using a two-component mixture model as an unsupervised generative model of sample loss values during training to allow an online estimation of the probability that a sample is mislabelled, and~the loss is corrected, relying on the network~prediction.

Overall, targeting the learning problems where labelled data are scarce or imperfect, \textsf{WSL} mitigates the labelled data dependency. Focusing on the data labelling part, the~task latency, cost, and energy consumption are optimized compared to manual labelling process. {This enhance the availability of the model as fewer resources are required for data labeling. The~quality of labels in WSL is often lower than in fully supervised learning, which may impact the reliability of the model's predictions.}

\subsubsection{Incremental~Learning}	

Incremental learning~\cite{Losing2018}, also called continual learning, is a machine learning paradigm that regularly processes periodically collected data and continuously integrates newly learned information to models in order to keep models up to date to the evolving data representation or task. Contrary to conventional offline learning, where all training data are available at the beginning of the learning process, and~models are firstly built by learning all data batches or samples through epochs for prediction, incremental learning is suitable for learning problems where data are collected over time. In~this case, the~data distribution, the~feature space, or~even the task evolve over time. Thus, the~trained model is expected to be periodically updated in order to capture and adapt to the new evaluations. 
Incremental learning takes advantage of a higher quality of data, closeness to the testing environment, and~continuously personalizes the pre-trained model with new classes. This learning paradigm can maintain and improve task accuracy when an original pre-trained model cannot generalize well. Moreover, the~incremental learning updates model locally and thus preserves the privacy in the case of local~deployment. 

With respect to the incremental learning setup, online learning, as~an instantiation of incremental learning in an online scenario~\cite{He2020a}, continuously learns from data provided in sequence from a data stream and produces a series of versions of the same model for prediction. This is performed without having the complete training dataset available at the beginning. The~model is deployed online to continuously realize intervened updates and predictions. In~particular, as~new data are usually generated very fast from the data stream such as in the case of Twitter data~\cite{Lin2012}, online learning typically uses data samples for only one epoch training and then switches for newer samples. Furthermore, lifelong learning~\cite{Ling2021} is another incremental learning branch that is characterized by the time span of the learning process and refers to the incremental learning in an infinite time span, to~accumulate the learned knowledge for future learning and problem~solving.

One major challenge of incremental learning is the continuous model adaptation and efficient paradigm design of learning from new data. One typical cause is the concept drift~\cite{Lu2019} which occurs over time, leading to a change in the functional relationship between the model inputs and outputs. Furthermore, learning data of new classes, the~model can forget previously learned knowledge. This refers to another cause: the catastrophic forgetting~\cite{Kirkpatrick2017}. An~early work~\cite{Maloof2004} incorporates  incremental learning with the partial instance memory of data samples from the boundaries of the induced concepts. The~model updates are based on both previous and new samples. Online learning~\cite{He2020a} employs a cross-distillation loss together with a two-step learning technique respectively for the new class data learning and the exemplar data learning to tackle catastrophic forgetting. Furthermore, it counts on the feature-based exemplary set update to mitigate the concept drift. This method outperforms the results of current state-of-the-art offline incremental learning methods on the CIFAR-100 and ImageNet-1000 datasets in online scenarios. To~perform lifelong learning on edge devices with limited computation resources, a~dynamically growing neural network architecture is introduced in~\cite{Piyasena2020} based on self-organization neural network (\textsf{SONN})~\cite{Marsland2002}. In~the architecture, a~\textsf{CNN} backbone is used as the encoder and the \textsf{SONN} is applied after as the classifier with the capability to grow the network when required to performance lifelong object detection on \textsf{FPGA}.

Incremental learning is excellent at autonomously adapting models to continuously changing environments of data, features, and~task spaces, and~thus improves the reliability. By~learning from data closer to the prediction environment, the~model performance on real environments is improved as well. In~particular, the~incremental learning fits well to the edge environment with limited computing resources, as~data can be fetched for learning in a piecemeal manner and then discarded right after the training, {which improves the availability}. In~an online setting, incremental learning consumes more network bandwidth and computation resources in exchange for a higher model performance and adaptation capability. The~cost and energy consumption are~increased. 

\section{Technique Review~Summary}	
\label{sec:TechniqueReviewSummary}

In this section, we summarize in Tables~\ref{tab:t03} and ~\ref{tab:t04} all reviewed techniques with regard to the  \textsf{Edge ML} requirements. The~three left columns illustrate the individual techniques, or~technique groups, while the top two rows list the \textsf{Edge ML} requirements. The~following notations are used to facilitate the relationship descriptions between techniques and~requirements. 

\begin{itemize}
	\item ``\textbf{+}'': {The reviewed technique improves the corresponding \textsf{Edge ML} solution regarding the specific \textsf{Edge ML} requirement. For~instance, quantization techniques reduce the inference task latency by simplifying the computation complexity.}
	
	\item ``\textbf{\hl{-}%MDPI:Please check if this should be minus sign? -> yes, should be the minus sign, updated.
}'': {The reviewed technique negatively impacts the corresponding \textsf{Edge ML} solution regarding the specific \textsf{Edge ML} requirement. For~instance, quantization techniques lead to accuracy loss during inference due to the low precision representation of data.}
	
	\item ``\textbf{*}'': {The impact of the reviewed technique on the corresponding \textsf{Edge ML} solution varies according to the specific configurations or setup. For~instance, transfer learning techniques improve the target model performance under the conditions that the source task and the target task are correlated, and~the data quantity and quality on the target domain are sufficient. The~weakness in data quantity or quality on the target domain can result in unsatisfactory model performance.}
	
	\item ``\textbf{/}: {The reviewed technique does not directly impact the corresponding \textsf{Edge ML} solution regarding the specific \textsf{Edge ML} requirement. For~instance, federated learning techniques do not directly improve or worsen the labelled data independence for a supervised learning process.}
\end{itemize}

Moreover, the~two following assumptions have been made to assure an objective evaluation of each \textsf{Edge ML} technique regarding the requirements:

\begin{itemize}
	\item Appropriate modelling and learning: all models for \textsf{ML} tasks are designed and trained following the state-of-the-art solution. No serious over-fitting or under-fitting has occurred, so that the models’ performance can be compared before and after applying the \textsf{Edge ML} techniques. 
	
	\item Statistic scenario: When performing a task, statistic scenarios instead of the best or the worst scenario are considered for techniques evaluation, as~certain techniques, e.g.,~Early Exit of Inference, can produce worse results compared to the corresponding conventional technique in extreme cases where all the side branch classifiers in a model do not produce high enough confidence and thus it fails to stop the inference earlier. However, statistically, the \textsf{EEoI} technique is able to improve energy efficiency and optimize cost when performing a number of running tasks. 
\end{itemize}

%\begin{sidewaystable*}

	\startlandscape
\begin{table}    %include the tabular into table floating environment so that the table can be recognized by ToC
~~~~\vspace{6pt}
		\caption{\hl{Edge ML:} %MDPI: please check if the alignment keep and if vertical line can be removed? -> alignment keeps, please keep the vertical line for readability.
 Techniques meet Requirements---Part~I.}
            \small
		\resizebox{\linewidth}{!}{
			\begin{tabular}{|c|c|c|c|c|c|c|c|c|c|c|}
				 \noalign{\hrule height 1pt}
				\multicolumn{3}{|c}{\multirow{5}{*}{\textbf{\hl{Edge ML Techniques} %MDPI: we add bold to table header, please confirm -> ok
}}} & \multicolumn{8}{|c|}{\textbf{Edge ML Requirements}}  \\	
				\cline{4-11}	
				\multicolumn{3}{|c}{} & \multicolumn{5}{|c}{\textsf{\textbf{Machine Learning}}} & \multicolumn{3}{|c|}{\textbf{Edge Computing}} \\
				\cline{4-11}
				\multicolumn{3}{|c|}{} & \makecell{\textbf{Low} \\ \textbf{Task} \\ \textbf{Latency}} 	& \makecell{\textbf{High} \\ \textbf{Performance}}	& \makecell{\textbf{Generalization} \\ \textbf{and} \textbf{Adaptation}} & \makecell{\textbf{Enhanced} \\ \textbf{Privacy} \\ \textbf{and} \\ \textbf{Security}}	& \makecell{\textbf{Labelled Data} \\ \textbf{Independency}}	& \makecell{\textbf{Computational} \\ \textbf{Efficiency}}	& \makecell{\textbf{Optimized} \\ \textbf{Bandwidth}}	& \makecell{\textbf{Offline} \\ \textbf{Capability}} \\
				\cline{1-11}	
				\multirow{8}{*}{\rotatebox[origin=c]{90}{Edge  Inference}} & \multirow{4}{*}{\makecell{Model \\ Compression and \\  Approximation}} & Quantization & + & {*} & /	& /	& /	& +	& /	& /	 \\ 
				\cline{3-11}
				&  & Weight Reduction & + & {*} & /	& /	& /	& +	& /	& /	 \\ 		
				\cline{3-11}
				&  & Knowledge Distillation & + & \hl{-} %MDPI: Please check if this should be minus sign? -> yes, should be the minus sign in the table. 
 & /	& /	& /	& +	& /	& /	 \\ 	
				\cline{3-11}
				&  & \makecell{Activation Function \\ Approximation} & + & \hl{-} & /	& /	& /	& +	& /	& /	  \\ 	
				\cline{2-11}		
				&  {\makecell{Distributed \\ Inference}}	&  & +	& /	& /	& /	& /	& \hl{-}	& \hl{-}	& \hl{-}	\\
				\cline{2-11}
				%\multirow{3}{*}{\rotatebox[origin=c]{90}{Low Task Latency}} 
				& \multirow{3}{*}{\makecell{Other Inference \\ Acceleration}} & Early Exit & +	& -	& /	& /	& /	& +	& /	& /	 \\
				\cline{3-11}
				& \multirow{3}{*}{} & Inference Cache & +	& \hl{-}	& /	& /	& /	& +	& /	& /		\\
				\cline{3-11}
				& \multirow{3}{*}{} & \makecell{Model-Specific \\ Inference Acceleration} & +	& \hl{-}	& /	& /	& /	& +	& /	& /	 \\
				\cline{1-11}		 
				\multirow{9}{*}{\rotatebox[origin=c]{90}{Edge  Learning}} & \multirow{2}{*}{\makecell{Distributed \\ Learning}} & Federated Learning & + &	\hl{-}	& /	& +	& /	& \hl{-}	& +	& -	 \\
				\cline{3-11}		 
				& & Split Learning & + &	/	& /	& +	& /	& \hl{-}	& +	& \hl{-}	 \\	
				\cline{2-11}		 
				& \makecell{Transfer \\ Learning} & & + &	*	& /	& /	& +	& +	& /	& /	 \\ 
				\cline{2-11}			 
				& Meta-Learning 		& 			& + &	/	& +	& /	& +	& *	& /	& /	 \\ 	
				\cline{2-11}			 	
				& \makecell{Self-Supervised \\ Learning} &  		& + &	+	& +	& /	& +	& +	& /	& /	 \\  
				\cline{2-11}			 
				& \multirow{4}{*}{\makecell{Other Learning \\ Paradigms}} & Multi-Task Learning & +	& +	& /	& /	& /	& +	& /	& /	 \\ 
				\cline{3-11}		 
				&  & Instance-based Learning & *	& *	& /	& /	& +	& *	& /	& /	 \\
				\cline{3-11}		 	
				&  & Weakly Supervised Learning & +	& /	& /	& /	& +	& /	& /	& /	 \\	
				\cline{3-11}		 
				&  & Incremental Learning & \hl{-}	& +	& +	& /	& /	& \hl{-}	& \hl{-}	& /	 \\	
			 \noalign{\hrule height 1pt}
				% \multirow{5}{*}{\rotatebox[origin=c]{90}{Data Processing}} & \multirow{4}{*}{\makecell{Dimensionality \\ Reduction}} & Random Projection & +	& *	& /	& /	& /	& +	& /	& /	 \\
				% \cline{3-11}	
				% & & Encoding \& Embedding & +	& *	& /	& /	& /	& +	& /	& /	 \\
				% \cline{3-11}
				% &  & Feature Selection & +	& *	& /	& /	& /	& +	& /	& /	 \\
				% \cline{3-11}	
				% &  & Feature Extraction   & +	& *	& /	& /	& /	& +	& /	& /	 \\
				% \cline{2-11}						 	
				% & \makecell{Input-Dependent \\ Computation} & & +	& -	& /	& /	& /	& +	& /	& /	 \\	
				% \cline{1-11}	  	 		 					
		\end{tabular}}
		\label{tab:t03}
	\end{table}
\finishlandscape

%\end{sidewaystable*}

	\begin{table}[H]    %include the tabular into table floating environment so that the table can be recognized by ToC
	~~\vspace{6pt}
		\caption{\hl{Edge ML:} %MDPI: please check if the alignment keep and if vertical line can be removed?  -> alignment keeps, please keep the vertical line for readability.
 Techniques meet Requirements---Part~II.}
            \small
		\resizebox{\linewidth}{!}{
			\begin{tabular}{|c|c|c|c|c|c|c|}
			 \noalign{\hrule height 1pt}
				\multicolumn{3}{|c}{\multirow{4}{*}{\textbf{\hl{Edge ML Techniques} %MDPI: we add bold to table header, please confirm -> ok
}}} & \multicolumn{4}{|c|}{\textbf{Edge ML Requirements}}  \\	
				\cline{4-7}	
				\multicolumn{3}{|c}{} & \multicolumn{4}{|c|}{\textsf{\textbf{Overall}}} \\
				\cline{4-7}
				\multicolumn{3}{|c|}{} & \makecell{\textbf{Availability}} 	& \makecell{\textbf{Reliability}}	& \makecell{\textbf{Energy} \\ \textbf{Efficiency}}	& \makecell{\textbf{Cost }\\\textbf{Optimization}} \\
				\cline{1-7}	
				\multirow{8}{*}{\rotatebox[origin=c]{90}{Edge  Inference}} & \multirow{4}{*}{\makecell{Model \\ Compression and \\  Approximation}} & Quantization & + & {*} & +	& +		 \\ 
				\cline{3-7}
				&  & Weight Reduction & + & * & +	& +		 \\ 		
				\cline{3-7}
				&  & Knowledge Distillation & + & * & +	& +		 \\ 	
				\cline{3-7}
				&  & \makecell{Activation Function \\ Approximation} & + & * & + & +	  \\ 	
				\cline{2-7}		
				&  {\makecell{Distributed \\ Inference}}	&  & +	& +	& \hl{-} %MDPI: Please check if this should be minus sign? -> yes should be the minus sign in this table. 
	& \hl{-}	\\
				\cline{2-7}
				%\multirow{3}{*}{\rotatebox[origin=c]{90}{Low Task Latency}} 
				& \multirow{3}{*}{\makecell{Other Inference \\ Acceleration}} & Early Exit & +	& -	& +	& +		 \\
				\cline{3-7}
				& \multirow{3}{*}{} & Inference Cache & +	& +	& +	& +			\\
				\cline{3-7}
				& \multirow{3}{*}{} & \makecell{Model-Specific \\ Inference Acceleration} & +	& *	& +	& +		 \\
				\cline{1-7}		 
				\multirow{9}{*}{\rotatebox[origin=c]{90}{Edge  Learning}} & \multirow{2}{*}{\makecell{Distributed \\ Learning}} & Federated Learning & + &	*	& +	& +		 \\
				\cline{3-7}		 
				& & Split Learning & + &	*	& +	& +		 \\	
				\cline{2-7}		 
				& \makecell{Transfer \\ Learning} & & + &	*	& +	& +		 \\ 
				\cline{2-7}			 
				& Meta-Learning 		& 			& * &	+	& *	& *		 \\ 	
				\cline{2-7}			 	
				& \makecell{Self-Supervised \\ Learning} &  		& * &	+	& +	& +		 \\  
				\cline{2-7}			 
				& \multirow{4}{*}{\makecell{Other Learning \\ Paradigms}} & Multi-Task Learning & +	& +	& +	& +		 \\ 
				\cline{3-7}		 
				&  & Instance-based Learning & *	& *	& *	& *		 \\
				\cline{3-7}		 	
				&  & Weakly Supervised Learning & +	& \hl{-}	& +	& +		 \\	
				\cline{3-7}		 
				&  & Incremental Learning & +	& +	& +	& +		 \\	
			 \noalign{\hrule height 1pt}
				% \multirow{5}{*}{\rotatebox[origin=c]{90}{Data Processing}} & \multirow{4}{*}{\makecell{Dimensionality \\ Reduction}} & Random Projection & +	& *	& /	& /	& /	& +	& /	& /	 \\
				% \cline{3-11}	
				% & & Encoding \& Embedding & +	& *	& /	& /	& /	& +	& /	& /	 \\
				% \cline{3-11}
				% &  & Feature Selection & +	& *	& /	& /	& /	& +	& /	& /	 \\
				% \cline{3-11}	
				% &  & Feature Extraction   & +	& *	& /	& /	& /	& +	& /	& /	 \\
				% \cline{2-11}						 	
				% & \makecell{Input-Dependent \\ Computation} & & +	& -	& /	& /	& /	& +	& /	& /	 \\	
				% \cline{1-11}	  	 		 					
		\end{tabular}}
		\label{tab:t04}
	\end{table}

{From Tables~\ref{tab:t03} and ~\ref{tab:t04}, one can see that most of edge inference techniques focus on reducing inference workload to improve computational efficiency, task latency, and availability. Concerning the reliability, the~compressed models put less stress on the hardware, potentially reducing the risk of hardware faults or failures, while in the meantime they are less resilient to hardware faults or slight changes in the input themselves. Distributed inference makes the inference execution of large models possible on the edge side by introducing a greater computational and communication workload for coordination and synchronization among edge clients. 
Regarding the distributed learning, split learning is able to offer a more competitive performance and privacy compared to federated learning, when the cloud server is available to cooperate on the training process. Distributed learning can potentially increase reliability by introducing additional nodes; however, on the other hand, the~additional complexity to manage and synchronize different resources leads to more points of failure. The~overall impact on reliability will depend on the detailed configurations of the distributed learning technique implementation. Transfer learning mainly focuses on accelerating the training task latency by facilitating  knowledge sharing across domains, whilst meta-learning and self-supervised learning respectively provide an efficient and a consolidated way to learn the data representation instead of specific tasks from labeled and unlabeled data to facilitate the learning of new tasks. 
Moreover, other learning paradigms, i.e.,~instance-based learning and weakly supervised learning, provide alternative solutions to directly learn from instances or partially labelled data. Multi-task learning is efficient to reduce model size and discover task correlations for better performance when multiple correlated tasks need to be realized simultaneously.
At last, incremental learning improves the model performance by continuously adapting models to the real environment by learning from ever-evolving data. The~overall requirements of energy efficiency and cost optimization are met by most  \textsf{Edge ML} techniques from different aspects of \textsf{ML} and~EC. }

\section{Edge ML~Frameworks}	
\label{sec:EdgeMLFramework}

To facilitate the implementation of \textsf{ML} solutions on the \textsf{Edge}, specific frameworks have been developed targeting various devices and platforms. In~this section, we briefly review the representative development frameworks supporting the \textsf{Edge ML} implementation. The~future direction related to \textsf{Edge ML} frameworks is summarized together with other open issues in the next~section. 

For edge inference frameworks, the~general process to deploy a model on the edge starts by choosing an existing trained model, and~converting the model to a specific format supported by the framework. In~parallel with the model conversion, model compression methods, such as quantization, are generally offered by frameworks and can be applied to reduce the model size. Finally, the~model is deployed to the edge devices for inference tasks. Frameworks for edge inference can be grouped into two categories: (i) cross-platform frameworks and (ii) platform-specific~frameworks.  

\textbf{\hl{Cross-Platform Inference Framework.}} Several popular native \textsf{Edge ML} frameworks exist to provide a general development solution across different devices and Operating Systems (\textsf{OS}). TensorFlow Lite~\cite{Singh2020a} is a light version of TensorFlow~\cite{Pang2020a} to deploy models on mobile and embedded devices, enabling quantization and on-device inference. 
The framework supports devices running with Android, iOS, or~embedded Linux OS, as~well as devices based on Micro-controllers (\textsf{MCUs}) and Digital Signal Processors (\textsf{DSPs}) without any \textsf{OS}. The~main advantage of TensorFlow Lite is that it does not require \textsf{OS} support, any standard C or C++ libraries, or~dynamic memory allocation. Alternatively, PyTorch Mobile~\cite{PyTorch2022}, as~the edge version of PyTorch~\cite{NEURIPS2019_bdbca288}, supports model inference on devices with an \textsf{iOS}, Android, or Linux system. It also integrates 8-bit kernels for quantization. The~Embedded Learning Library (\textsf{ELL}) from Microsoft is a similar framework allowing the deployment of \textsf{ML} models onto resource-constrained platforms and small single-board computers such as Raspberry Pi, Arduino, and~micro:bit running Windows, Linux, or~macOS. 
To support both edge inference and edge training, \textsf{CoreML}~\cite{Marques2020a} is a machine learning framework across all of Apple’s OSs including macOS, iOS, tvOS, and watchOS. Both training and inference are enabled, with~pre-built features such as memory footprint reduction for performance optimization. However, since models are encapsulated inside an application in \textsf{CoreML}, additional computational complexity is~added. 
    
\textbf{\hl{Platform-Specific Inference Framework.}} Besides the cross-platform frameworks, numerous platform-specific frameworks have been developed to support the inference task on specific processors including (i) General Purpose Processors (\textsf{GPP}s) such as \textsf{CPU}s and \textsf{GPU}s; (ii) Application-Specific Integrated Circuits (\textsf{ASIC}s) such as Google \textsf{TPU}~\cite{Cass2019a} and Intel Movidius \textsf{VPU}~\cite{Ionic2015b}; and~(iii) \textsf{MCU}s such as \textsf{STM32}~\cite{STMicroelectronics2014a}. Most of the existing mobile processors are based on Arm architecture. ARM Compute Library~\cite{Sun2017} offers a collection of low-level machine learning functions optimized for Cortex-A CPU and Mali GPUs architectures. Specifically, it provides the basic \textsf{CNN} building blocks for model inference in \textsf{CV} domains. 
Nvidia \textsf{GPU}s are widely used in PCs, and~TensorRT~\cite{Jeong2022}, as~the corresponding \textsf{ML} framework for edge inference across all Nvidia GPUs, supports distributed inference and numerous model structures extensively optimized for performance including optimizer and run-time that delivers low latency and high throughput for edge~inference.

Besides \textsf{GPP}s, a~large number of \textsf{ASIC}s specific to \textsf{Edge ML} applications are manufactured~\cite{Li2020}. To~facilitate the direct use, \textsf{ASIC}s are commonly integrated into System-on-a-Chip (\textsf{SoC}), System-on-Module (\textsf{SoM}) or Single-Board Computer (\textsf{SBC}), and~the corresponding frameworks specific to \textsf{ASIC} vendors are developed to offer a complete \textsf{ML} computing solution. Aiming at the NVIDIA Jetson family of hardware, NVIDIA EGX Platform~\cite{NVIDIACorporation2021}, a~combination of Nvidia \textsf{GPU}s and VMware vSphere with NVIDIA Certified Systems,  provides a full stack software infrastructure to deploy \textsf{Edge ML} applications for~inference. 

The Qualcomm Neural Processing SDK for ML~\cite{QualcommTechnologies2021} is an \textsf{Edge ML} framework to optimize and deploy trained neural networks on devices with Snapdragon SoC family from Qualcomm. The~Qualcomm Neural Processing SDK supports convolutional neural networks and custom layers. Concerning MCUs, Neural Network on Micro-controller (\textsf{NNoM})~\cite{git-nnom} is a high-level inference neural network library specifically for  micro-controllers, which helps to convert a model trained with Keras a to native~\textsf{NNoM} model for deployment. \textsf{NNoM} supports several convolutional structures including Inception~\cite{Szegedy2015}, ResNet, and~DenseNet~\cite{Huang2017} and recurrent layers such as simple \textsf{RNN}, \textsf{GRU}, and~\textsf{LSTM}. 
X-CUBE-ML~\cite{STMicroelectronics2022} is an \textsf{ML} expansion framework for \textsf{STM32 MCU} with the capability to convert and deploy pre-trained neural networks and classical machine learning models, and~performance measurement functions for \textsf{STM32} are directly integrated as~well. 
    
Again, in~order to match the nature of the target \textsf{ASIC}s and the embedding platforms, most of the frameworks must go through a process of conversion, modification, and, in~some cases, complete retraining. 
Since most of native edge frameworks only support edge inference, we hereafter illustrate general distributed learning frameworks for learning purposes, which can be used in edge devices with relatively high computation capability such as PC and mobile phones. The~frameworks are mainly developed to facilitate the parallel training (i.e., data-parallelism, model-parallelism, and~3D parallelism, as~summarized in~\cite{Narayanan2021}) and federated~learning. 

\textbf{\hl{Distributed Learning Framework}.} PyTorch~\cite{NEURIPS2019_bdbca288} is one of the most popular deep learning frameworks, offering both data-parallelism and model-parallelism functions for model training. Built upon PyTorch, DeepSpeed~\cite{Rasley2020} extends the PyTorch with additional 3D parallelism support as well as memory optimization for exascale model training. As~an alternative ecosystem of PyTorch, Distributed API in Tensorflow provides data-parallel techniques for model training, while Mesh TensorFlow~\cite{git-mesh} extends the Distributed TensorFlow with model-parallelism and enables large model training on Google \textsf{TPU}s. Mindspore~\cite{Chen2021b} is another general \textsf{ML} computing framework from Huawei with end-to-end ML capabilities for model development, execution, and~deployment in various scenarios including distributed training and cloud-edge deployment. 
Regarding the federated learning frameworks, TensorFlow Federated~\cite{GoogleInc.2022} is the federated learning framework from the TensorFlow ecosystem on decentralized data. The~framework enables developers to simulate the included federated learning algorithms on their models and data, as~well as to experiment with novel algorithms. For~deployment in real environments, Open Federated Learning~\cite{Intel} from Intel provides deployment scripts in bash and leverages certificates for securing communication. Finally, NVIDIA CLARA~\cite{NvidiaClara2020} includes full-stack GPU-accelerated libraries covering training schemes, communications, and~hardware configurations for server-client federated learning with a central aggregation~server.

\section{Open Issues {and Future~Directions}}	
\label{sec:OpenIssues}

Despite the diverse methods and paradigms of \textsf{Edge ML} and the initial success of their powered edge solutions, challenges and open issues are not rare in the \textsf{Edge ML} field, slowing down the technological progress. In~this section, we summarize some open issues of \textsf{Edge ML} to shed light on its future~directions. 

\textbf{\hl{Learning Generalization and Adaptation.}} Currently \textsf{ML} techniques are going through a transition from the learning of specific labels to the learning of data representations. Meta-learning and self-supervised learning provide intuitive manners to progress in this direction. Nevertheless, meta-learning usually relies on a support dataset to perform any task-specific adaptation, and~self-supervised learning requires tuning as well for specific tasks. The~generalization from representation learning brings general cognitive abilities to models, while automatic adaptation techniques to specific tasks such as zero-shot learning in \textsf{NLP} need to be further studied and explored so that specific tasks can be solved directly without performing any adaptation process. {In the future, machine learning techniques will continue to advance from merely learning task-specific labels to more generalized data representations. This progression is poised to enhance machine learning models' cognitive abilities, pushing the achievement boundaries. This is particularly important to \textsf{Edge ML} as human intervention or guidance are not guaranteed compared to the cloud-based solutions.}  

\textbf{\hl{Theoretical Foundation.}} With the rapid emergence of \textsf{Edge ML} techniques, the~theoretical foundation related to the emerging techniques for optimal design and empirical validation are not up to date. For~example, most model compression and approximation methods do not have mathematical proofs for the optimal compression ratio. %Check meaning retained
Federated learning also may not converge in the training process, if~the data distribution varies largely from clients. Finally,  self-supervised learning continuously seeks optimal contrastive objective functions to optimize learning efficiency. Theoretical foundations are crucial to validate empirical conclusions from emerging fields and provide guidelines for the optimal design of \textsf{Edge ML} solutions. {In the foreseeable future, significant advancements in the theoretical foundation of Edge ML techniques are expected. These advancements will primarily focus on the development of mathematical proofs and models for such Edge ML methods.} 

\textbf{\hl{Architectures for Heterogeneity and Scalability.}} An \textsf{Edge ML} environment is known to be heterogeneous in distribution of entities such as data, device resources, network infrastructures, and~even \textsf{ML} tasks and models. And~with a large number of participant edge devices, bottlenecks have been identified affecting \textsf{Edge ML} performance. Such bottlenecks include the communication bottleneck in federated learning for gradient communications and the computational bottleneck in meta-learning when the support set is large. Furthermore, all edge devices are not often activated at the same time, and~the temporal disparity feature makes it more challenging for the organizational architecture to manage the \textsf{Edge ML} solution. Adding local edge servers can alleviate the problem of the local perimeter, and~to reach the global heterogeneity management with a large number of edge devices. Advanced distributed architectures for \textsf{ML} tasks are expected to synchronize and coordinate entities among all heterogeneity levels and deliver robust and scalable solutions for dynamic and adaptive aggregation in distributed setup. {Given the inherent diversity in Edge ML environments, including variations in data, device resources, network infrastructures, and~ML tasks and models, there is a pressing need for more flexible and adaptive architectures. As~edge devices continue to proliferate, future work will need to address performance bottlenecks, such as communication constraints in federated learning and computational limitations in meta-learning.}

\textbf{\hl{Fortified Privacy.}} Privacy preservation is the primary objective in distributed learning and inference paradigms, as~no data are shared outside of the local client. However, sensitive information can still be leaked via methods such as the reverse deduction of models. Although~security- and privacy-oriented methods can improve the situation, a~significant computation complexity is introduced in the edge devices in the meantime, increasing task latency and energy consumption. Novel and lightweight computing paradigms are expected to protect data and model leakage during information exchange and go from enhanced privacy to fortified privacy. {The future direction is expected to focus on developing novel, lightweight computing paradigms that not only protect data from breaches but also prevent model leakage during information exchange. This future trend towards fortified privacy should bring forward the development of new methods and architectures that are efficient in terms of computational resources and energy usage. These advancements are expected to increase privacy without sacrificing performance, leading to a more secure and trustable Edge ML environment.}
 
\textbf{\hl{Hybrid Approach.}} With the reviewed techniques tackling different aspects of \textsf{Edge ML} requirements, hybrid strategies with more than one technique is now commonly adopted when designing \textsf{Edge ML} solution. Hybrid \textsf{ML} benefits from several techniques and can achieve better performance than the use of any single method. The~integration of two or three techniques are popular in the reviewed literature, while with a given set of design requirements, complete hybrid approaches covering all \textsf{Edge ML} phases, including data preprocessing, learning, and~inference, are missing. The~hybrid approach with a thorough technical design for each phase can best contribute to the improvement of model capability, and~thus is a direction worth exploring. {We envision the future of Edge ML will entail designing holistic hybrid solutions that address every phase with careful technical consideration. This approach not only enhances model capability but also ensures robust performance in varying edge scenarios.}

\textbf{\hl{Data Quality Assurance.}} Nowadays, a~huge amount of data is created on the edge devices at every second, but~most of it cannot be directly used by \textsf{ML} without the labeling and preprocessing processes. As~a step forward, self-supervised learning proves to be good at learning structured and unlabeled data. However, the~data quality such as noisy data, non-IID data, imbalanced distribution, or~data corruptions and errors, still impacts the learning results and tends to alter the model performance. Although~a number of methods are introduced, the~selection of suitable methods is determinant to the results and highly relies on expertise. Regular interaction with humans for labelling and selection of quality data are not realistic, especially for edge users, and~thus embedded learning paradigms integrating native data selection for quality control and preprocessing of different input qualities is the future of \textsf{Edge ML}. {In the future direction, assuring data quality will become increasingly important within the context of Edge ML. We envision the future of Edge ML to be geared towards the development of embedded learning paradigms, which integrate native data selection mechanisms for quality control and preprocessing. These embedded systems would automatically handle different input qualities, reducing the need for human intervention, and~leading to more reliable, autonomous Edge ML systems. }

\textbf{\hl{Framework Extension.}} The number of frameworks keeps increasing for \textsf{Edge ML}. However, due to the resource-constrained nature of the edge environment, existing frameworks tend to be lightweight for resource efficiency and thus limited in their support of \textsf{ML} features and functions: most of the native \textsf{Edge ML} frameworks are only designed for edge inference, and~involve additional steps and computation for model conversion. Device-specific frameworks often support a subset of neural network layers and activation functions, which requires model re-design and re-training before deployment as well. With~the rapid development of computing capability on edge devices, the~trade-off between resource efficiency and functionality can be further studied to extend the supporting edge features and functions. {Looking towards the future, there is a pressing need for the extension and expansion of the available frameworks for Edge ML. With~the rapid advancements in the computing capabilities of edge devices, there is an opportunity to reconsider the trade-off between resource efficiency and functionality. In~the future, we anticipate a shift towards frameworks that strike a better balance between these two aspects, providing more comprehensive support for Edge ML features and functions without drastically increasing resource requirements. This progression towards more functionally rich and resource-efficient frameworks will significantly impact the design and implementation of Edge ML solutions and remains a promising avenue for exploration and development in the field.} 

\textbf{\hl{Standardization.}} There are widespread standardization organizations (\textsf{SDO}s) on \textsf{ML} (e.g., ISO/IEC JTC 1/SC 42 Artificial Intelligence~\cite{StandardsbyISO/IECJTC2022e}, ITU-T Focus Groups~\cite{InternationalTelecommunicationUnionITU2021, ITU-TFG-ML5G2018}, IEEE SA Artificial Intelligence Systems, only to name a few) contributing to the community development and reference solutions. %Check meaning retained
However, there~is clearly very few ongoing activities within initiatives and SDOs (e.g., ETSI ISG EMC~\cite{Dahmen-Lhuissier2020}) focused on defining native specifications for \textsf{Edge ML} solutions. Along with the uprising development of \textsf{Edge ML} technologies, \textsf{Edge ML} standards and specifications covering \textsf{MLOps} life cycle in the edge environment are expected to fill the gap in the \textsf{Edge ML} ecosystem and optimize \textsf{ML} at the edge for reference and guidance. {Moving forward, it is expected that the standardization ecosystem will change, with~a more dedicated focus on creating standards and specifications that cover the MLOps lifecycle in the edge environment. The~development of such standards will likely address the current gaps in the Edge ML ecosystem, providing both guidance and reference solutions for implementing ML at the edge. Such standardization will not only streamline the development and deployment processes but also enhance system interoperability and reliability, making Edge ML more accessible and effective. }

\section{Conclusions}
\label{sec:Conclusion}

Due to the specific features of privacy preservation, low-latency experiences, and~low energy consumption, edge-powered machine learning solutions have been rapidly emerging in end-user devices for services and applications in the domains of \textsf{CV}, \textsf{NLP}, healthcare, \textsf{UAVs}, etc. In~this paper, we provide a comprehensive review of \textsf{Edge ML} techniques focusing on the two parts of \textsf{ML} solutions: (i) edge inference and~(ii) edge learning.
%, and (iii) data preprocessing. 
The review offers a panoramic view of the technique's perimeter through a thorough taxonomy. %Check meaning retained
Recent and representative works are presented for each technique with its targeting \textsf{Edge ML} requirements. Edge ML frameworks are introduced, while 
open issues are identified for future research directions. To~the best of our knowledge, this is the first review covering the entire and detailed technique perimeter of Edge ML learning and~inference. 

This paper can serve as a reference to select adaptive \textsf{ML} paradigms and build corresponding solutions in edge environments. Due to the large perimeter to cover, we adapt the review strategy to prioritize the technique width rather than the technique depth, and~thus further work will focus on surveying more detailed research challenges and methods for targets and specific techniques' branches. %Check meaning retained
In~the meantime, we are also investigating scalable architectures for \textsf{Edge ML} solutions over heterogeneity infrastructural resources, data and~tasks.

\vspace{6pt} 

\authorcontributions{Conceptualization, methodology, investigation, writing—original draft preparation and revision, W.L.; writing—review and editing, H.H.;  supervision, funding acquisition, E.A.; supervision, funding acquisition, M.D. All authors have read and agreed to the published version of the~manuscript.}

\funding{This research received no external~funding.}

\institutionalreview{Not applicable.}

\informedconsent{Not applicable.}

% Written informed consent for publication must be obtained from participating patients who can be identified (including by the patients themselves). Please state ``Written informed consent has been obtained from the patient(s) to publish this paper'' if applicable.}

\dataavailability{Not applicable.} 

% \acknowledgments{In this section you can acknowledge any support given which is not covered by the author contribution or funding sections. This may include administrative and technical support, or donations in kind (e.g., materials used for experiments).}

\conflictsofinterest{The authors declare no conflict of~interest.} 

%\bibliographystyle{ieeetr}

%\setcitestyle{nosort}

% \bibliographystyle{ACM-Reference-Format}
% \bibliographystyle{unsrt}
% \bibliography{library.bib}
%%%%%%%%%%%%%%%%%%%%%%%%%%%%%%%%%%%%%%%%%%
\begin{adjustwidth}{-\extralength}{0cm}
\reftitle{References}
\PublishersNote{}
\end{adjustwidth}
\end{document}